\documentclass[10pt,twocolumn,letterpaper]{article}

\usepackage[pagenumbers]{cvpr} 

\usepackage[dvipsnames]{xcolor}

\usepackage[accsupp]{axessibility}

\usepackage{xcolor}
\usepackage{array}
\usepackage{subcaption}
\usepackage{dsfont}
\usepackage{tabu}
\usepackage{longtable}
\usepackage{listings}
\usepackage{multirow}
\usepackage{soul}
\usepackage{tcolorbox}
\usepackage{float}
\usepackage[accsupp]{axessibility}
\usepackage{tabularray}
\usepackage{afterpage}
\UseTblrLibrary{booktabs}
\UseTblrLibrary{siunitx}
\NewColumnType{J}[1][]{Q[co=1,1.0,si={table-format=#1,table-number-alignment=center},c]}

\usepackage{siunitx}
\usepackage[misc]{ifsym}
\usepackage{placeins}
\usepackage[ruled,vlined]{algorithm2e}

\newcommand{\std}[1]{{\tiny$\pm$#1}}

\DeclareMathOperator*{\argmax}{arg\,max}

\usepackage{tcolorbox}
\newtcbox{\revisedsec}[1][yellow]{on line, arc=0pt,colback=#1!100!white,colframe=#1!100!white,
  before upper={\rule[-3pt]{0pt}{10pt}},boxrule=1pt, boxsep=0pt,left=6pt,
  right=6pt,top=2pt,bottom=2pt}

\newcommand{\subsubsubsection}[1]{\paragraph{#1}\mbox{}\\}

\usepackage{soul}
\soulregister\cite7
\soulregister\ref7

\definecolor{cvprblue}{rgb}{0.21,0.49,0.74}
\usepackage[pagebackref,breaklinks,colorlinks,citecolor=cvprblue]{hyperref}

\def\paperID{7373} 
\def\confName{CVPR}
\def\confYear{2024}

\title{FISBe: A real-world benchmark dataset for instance segmentation of \\ long-range thin filamentous structures}

\author{
\textbf{Lisa Mais$^{1,2,4,*,\text{\Letter}}$, Peter Hirsch$^{1,2,*}$, Claire Managan$^3$, Ramya Kandarpa$^3$,}\\
\textbf{Josef Lorenz Rumberger$^{1,2}$, Annika Reinke$^{2,5}$, Lena Maier-Hein$^{2,5}$,}\\
\textbf{Gudrun Ihrke$^3$, Dagmar Kainmueller$^{1,2,4,\text{\Letter}}$}\\
$^1$ Max-Delbrueck-Center for Molecular Medicine in the Helmholtz Association (MDC)
$^2$ Helmholtz Imaging\\
$^3$ HHMI Janelia Research Campus\quad
$^4$ University of Potsdam\quad
$^5$ German Cancer Research Center (DKFZ)\\
$^\text{\Letter}$ \texttt{\{firstname.lastname\}@mdc-berlin.de} \quad $^*$ shared first authors
}

\usepackage{capt-of,etoolbox}

\makeatletter
\apptocmd\@maketitle{{\myfigure{}\par}}{}{}
\makeatother
\begin{document}

\newcommand\myfigure{%
\centering
\vspace{-2.2em}
    \begin{tabular}{m{0.18\textwidth}m{0.18\textwidth}m{0.18\textwidth}m{0.18\textwidth}}
        \includegraphics[width=0.2\textwidth]{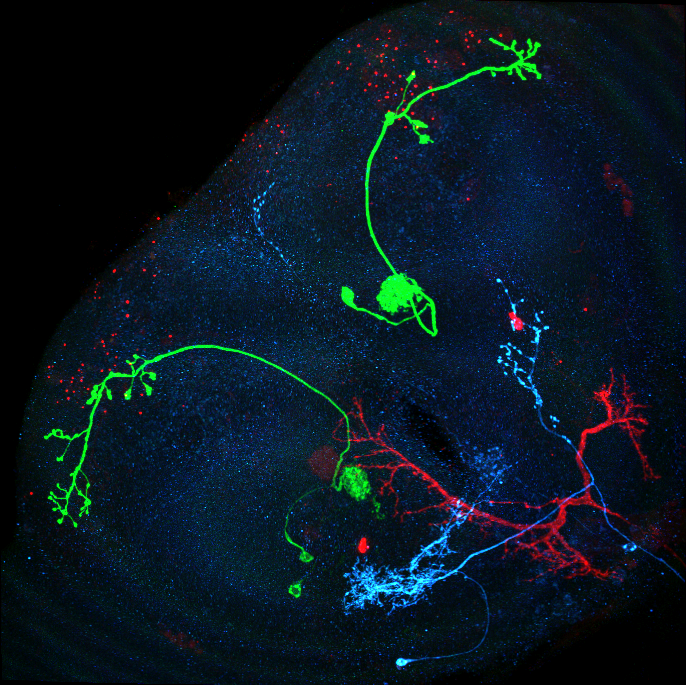} &
        \includegraphics[trim=0 46 0 46, clip, width=0.2\textwidth]{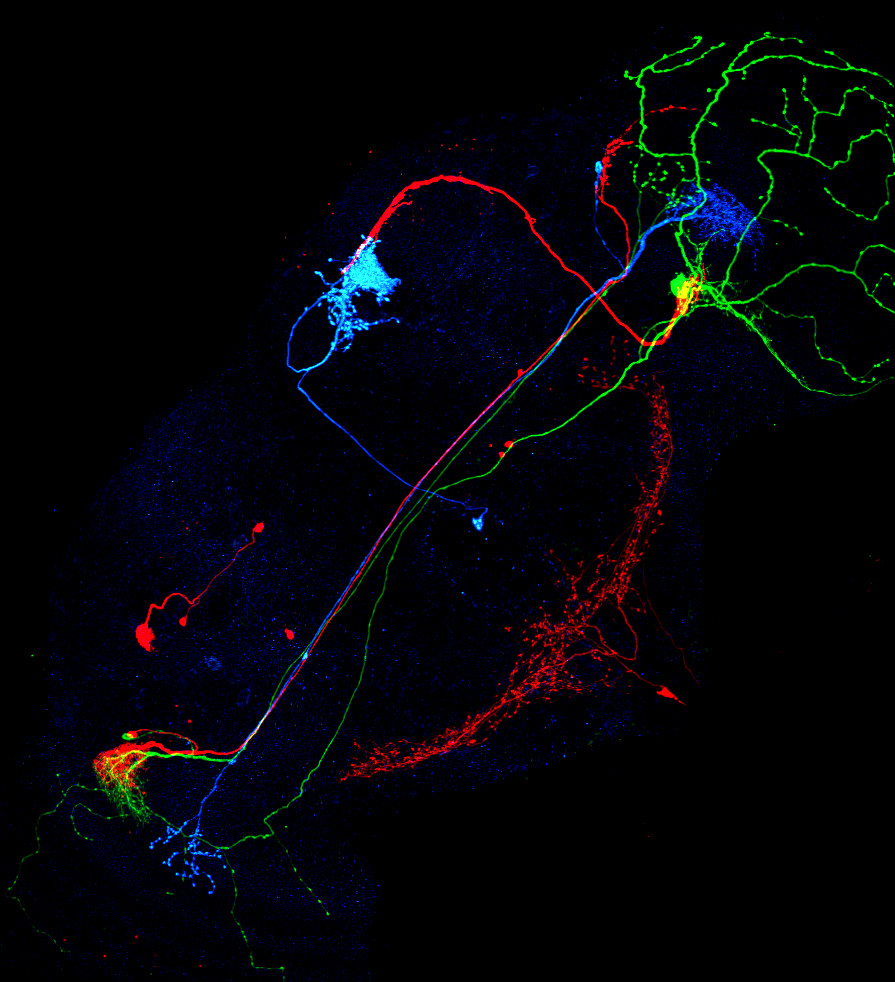} &
        \includegraphics[width=0.2\textwidth]{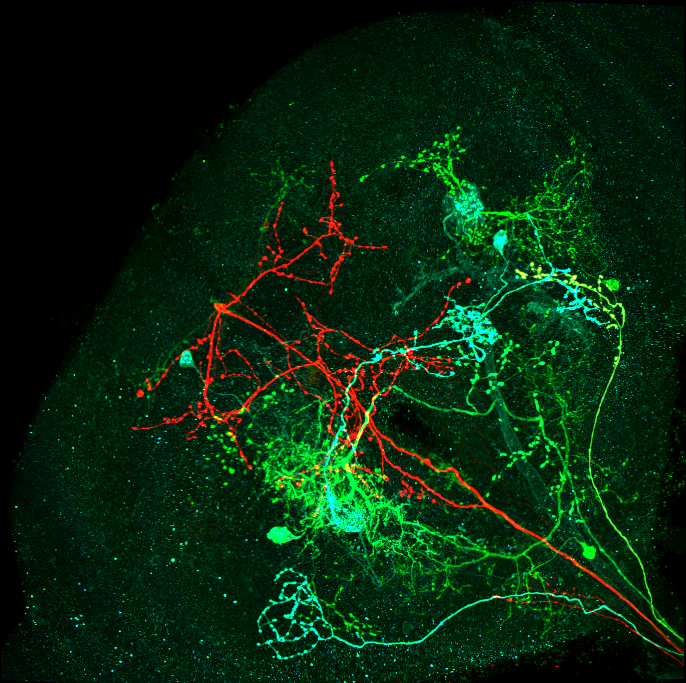} &
        \includegraphics[trim=0 24 0 24, clip, width=0.2\textwidth]{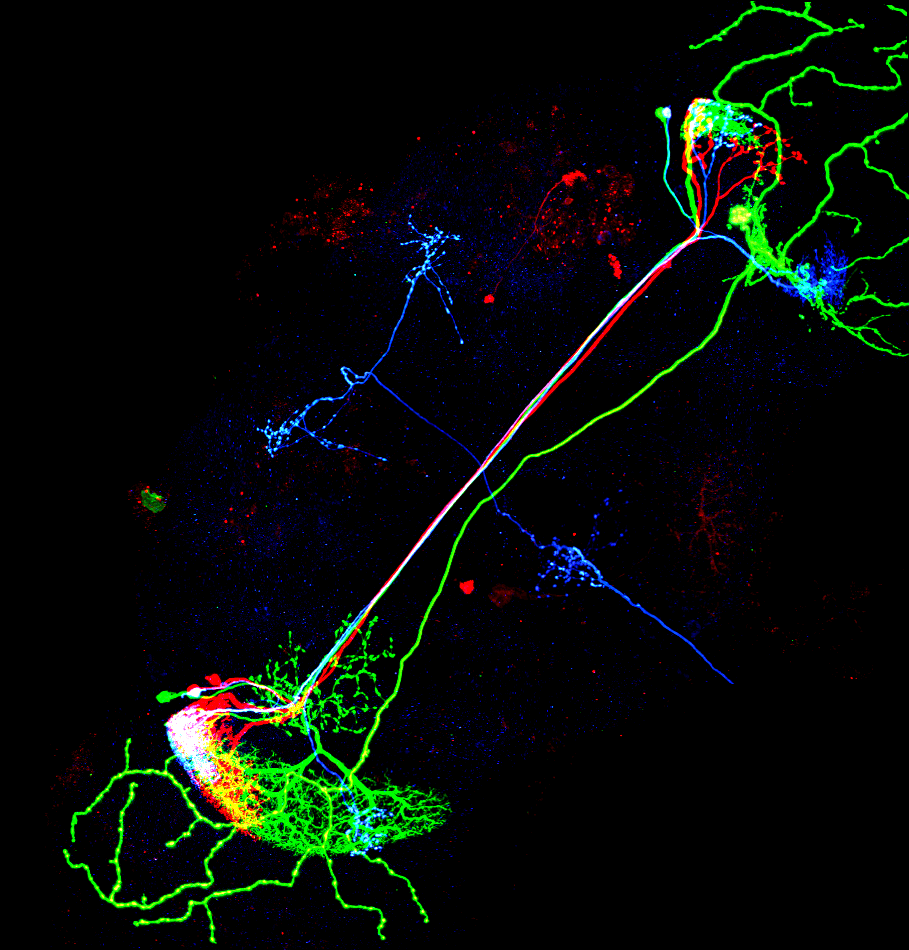}\\
        \includegraphics[width=0.2\textwidth]{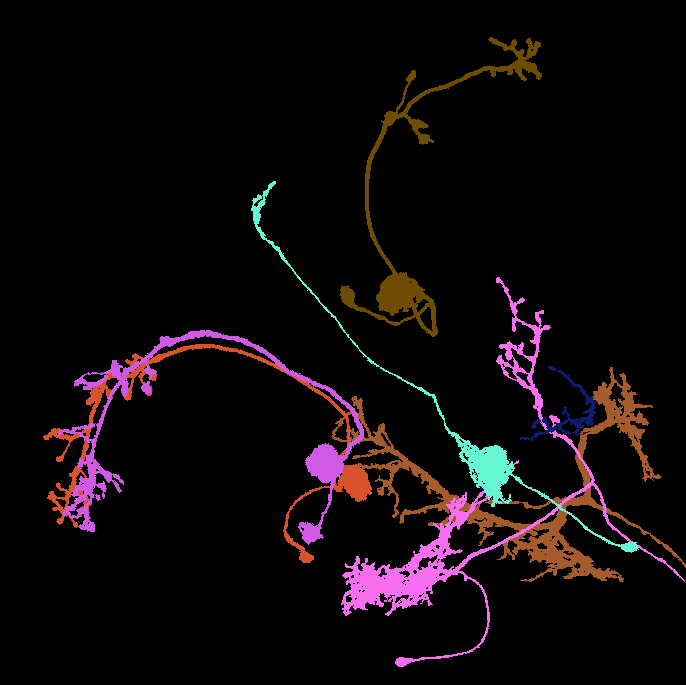} &
        \includegraphics[trim=0 46 0 46, clip, width=0.2\textwidth]{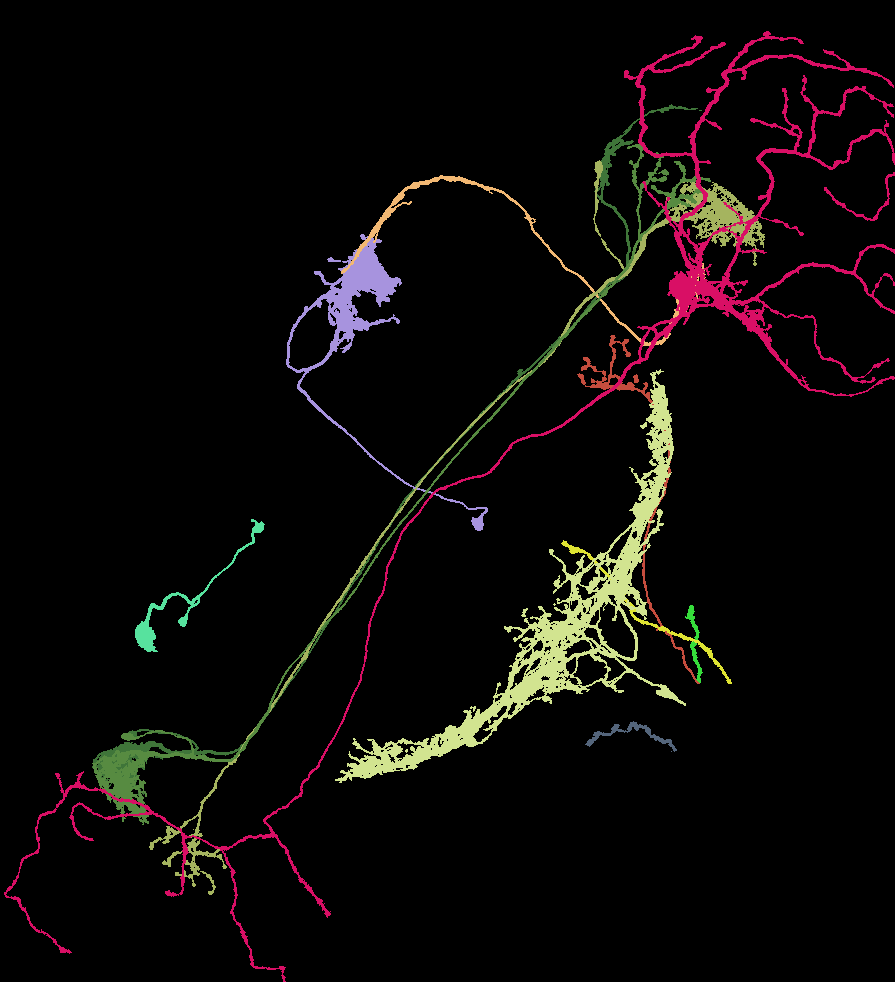} &
        \includegraphics[width=0.2\textwidth]{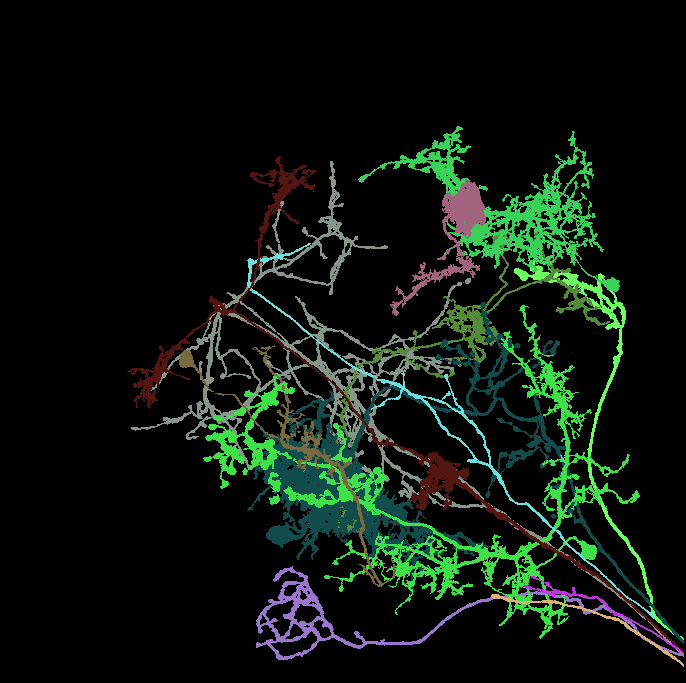} &
        \includegraphics[trim=0 24 0 24, clip, width=0.2\textwidth]{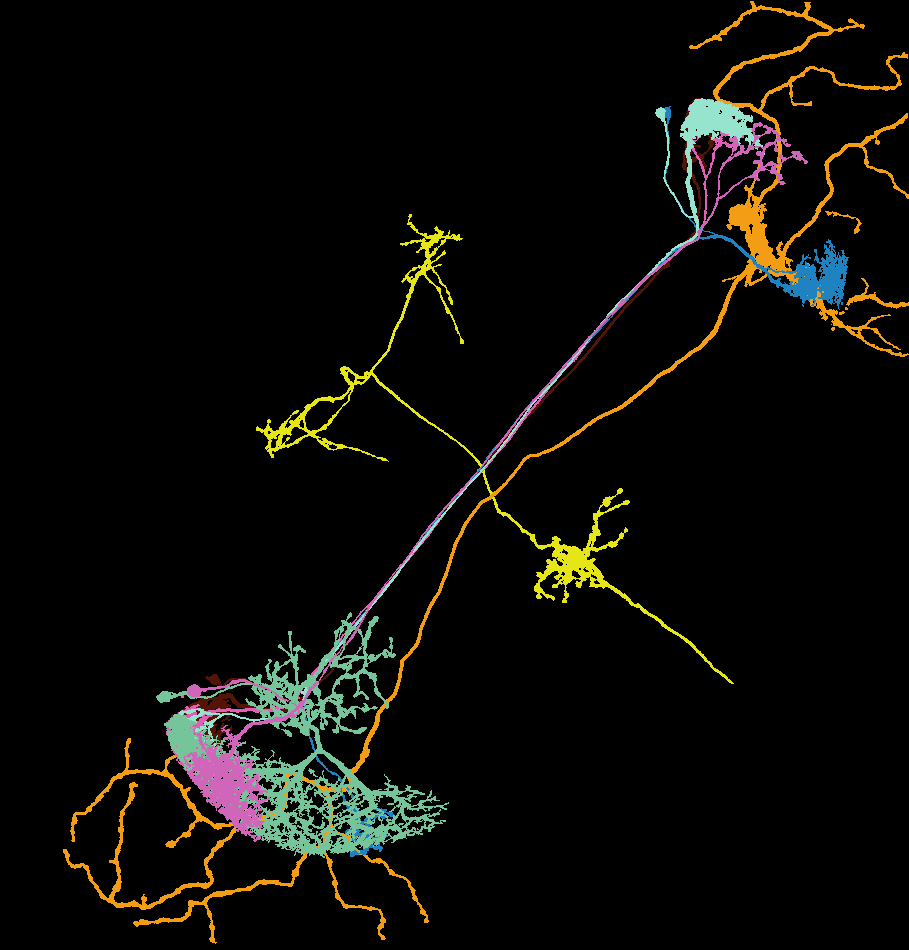}\\
    \end{tabular}
\captionof{figure}{\label{fig:showcases_data}We release the FlyLight Instance Segmentation Benchmark (FISBe) dataset, a 3d multi-color light microscopy dataset of neuronal morphologies in the brain of the fruit fly Drosophila melanogaster, together with high-quality pixel-wise instance segmentations of individual neurons. 
To the best of our knowledge, FISBe constitutes the first publicly available real-world benchmark dataset for instance segmentation of wide-spanning, thin filamentous and tightly interweaving objects. 
Top row: Exemplary FISBe images (3d) visualized in 2d via maximum intensity projection (MIP). Bottom row: Corresponding 2d projections of ground truth instance segmentation masks (3d).}
\vspace{1em}
}
\maketitle

\begin{abstract}
Instance segmentation of neurons in volumetric light microscopy images of nervous systems enables groundbreaking research in neuroscience by facilitating joint functional and morphological analyses of neural circuits at cellular resolution.
Yet said multi-neuron light microscopy data exhibits extremely challenging properties for the task of instance segmentation: 
Individual neurons have long-ranging, thin filamentous and widely branching morphologies, multiple neurons are tightly inter-weaved, and partial volume effects, uneven illumination and noise inherent to light microscopy severely impede local disentangling as well as long-range tracing of individual neurons.
These properties reflect a current key challenge in machine learning research, namely to effectively capture long-range dependencies in the data. While respective methodological research is buzzing, 
to date methods are typically benchmarked on synthetic datasets.
To address this gap, we release the FlyLight Instance Segmentation Benchmark (FISBe) dataset, the first publicly available multi-neuron light microscopy dataset with pixel-wise annotations.
In addition, we define a set of instance segmentation metrics for benchmarking that we designed to be meaningful with regard to downstream analyses. Lastly, we provide three baselines to kick off a competition that we envision to both advance the field of machine learning regarding methodology for capturing long-range data dependencies, 
and facilitate scientific discovery in basic neuroscience.
Project page: \href{https://kainmueller-lab.github.io/fisbe}{https://kainmueller-lab.github.io/fisbe}
\enlargethispage{\baselineskip}
\end{abstract}
\clearpage

\section{Introduction}
\label{sec:intro}
Most existing instance segmentation benchmarks in computer vision are collections of natural images~\cite{cocodataset,ade20k,Cordts2016Cityscapes}.
These are often suitably addressed with proposal-based methods like Mask R-CNN~\cite{rcnn_girshick2014,maskrcnn_he2017}, as the assumption that shapes of objects are well approximated by bounding boxes mostly holds.
However, this key assumption is violated in a range of highly relevant application domains, including neuroscience~\cite{mouselight19,cremi}, the domain the FISBe dataset we contribute in this work stems from.
Here, objects can span large parts of an image and have complex (e.g., tree-like) and intertwined shapes.
Consequently, multiple instances may have very similar, very large bounding boxes. 

Benchmarks from the neuroscientific domain, namely on neuron instance segmentation in electron microscopy (EM) data~\cite{arganda2015crowdsourcing}, have greatly facilitated the development of instance segmentation methodology that is applicable in the face of complex, wide-ranging object shapes. 
Beyond sophisticated object shapes, said benchmarks also call for methodology that applies to very large images, way beyond what current GPU memory can hold~\cite{cremi,wei2021}.
On these benchmarks, proposal-free methods based on CNN backbones~\cite{de2017semantic,chen2019instance,Kulikov_2020_CVPR,mutex_wolf2018,connectome_funke2018,lee2021learning} constitute the current state of the art, and mostly also lend themselves to arbitrarily large images thanks to tile-and-stitch inference. 

However, an object category that generally renders image segmentation very challenging is not represented in the above-mentioned benchmarks, namely objects that exhibit very thin (down to single-pixel width), filamentous structures. 
Benchmarks have been established for \emph{semantic} 
segmentation of very thin filamentous structures in a range of real-world applications, including neuron segmentation in light microscopy (LM) data~\cite{brown2011,bigneuron15,bigneuron23}, blood vessel segmentation in various medical imaging modalities~\cite{drive,Lyu2022reta}, and road extraction from satellite imagery~\cite{spacenet,roadtracer}. However, respective \emph{instance} segmentation benchmarks are currently lacking despite the high relevance of the task, e.g., in basic neuroscience
~\cite{meissner2023}.
The closest related publicly available dataset that exhibits thin and complex object shapes is~\cite{mouselight19}.
Yet it lacks tightly inter-weaved objects by design, and furthermore does not come with pixel-wise ground truth segmentations nor recommended metrics for benchmarking.

Consequently, there is at present a lack of methodology applicable for instance segmentation of wide-ranging thin filamentous intertwined shapes: Only very few deep-learning approaches are potentially suitable, 
among which Flood Filling Networks~\cite{januszewski2018ffn} and PatchPerPix~\cite{ppp2020}.
Most proposal-free instance segmentation methods  do not appear suitable:
Three-label models~\cite{caicedo_eval_deep_learn_nuclei_3label} degenerate in the face of very thin instances because their interior equals their boundary;
models predicting pixel affinities~\cite{fowlkes_aff03,turaga09_maxim_affin_learn_image_segmen} become inappropriate if they rarely encounter foreground in their fixed pixel neighborhoods (as compared to the dense-foreground EM data they were designed for);
metric learning models~\cite{chen2019instance,de2017semantic} lack the capacity to capture long-range connectivity beyond their receptive fields. 
Likewise, methods proposed specifically for cell instance segmentation do not appear suitable: Cellpose~\cite{stringer_2021_cellpose} assumes locally (i.e., within receptive field) visible cues towards some semantically meaningful center point which does not hold true for our dataset;
Stardist~\cite{weigert_2020_stardist} employs  star-convex polygons/polyhedrons as proposals, which do not provide viable approximations of neurons.
As for non-learnt, application-specific methods for neuron separation, some approaches rely on user-defined~\cite{ntracer_roossien2019} or pre-detected anchor points, in particular on cell body detection~\cite{neurogps_quan2015,zhou2021gtree,li2019precise}.
This renders these methods not directly applicable to our data, where cell bodies may lie outside of the imaged volume (namely in the ventral nerve cord). 
Other non-learnt application-specific approaches are based on color clustering~\cite{duan2021,sumbul2026}, which is technically applicable, yet the underlying assumption that each neuron has a unique color is often violated on our data.

A promising recent alternative are query-based methods~\cite{kirillov2023segment,cheng2021per,carion2020_detr}, 
which operate without explicit prior assumptions on object sizes or shapes.
However, e.g., SAM \cite{kirillov2023segment} is not directly applicable as it has not yet been extended to full 3d and it is unclear if and how tile-and-stitch prediction, as would be necessary given the size of individual FISBe images, could be achieved in a seamless manner.
We deem respective potential extensions of SAM a very interesting research topic for which, albeit out of scope, FISBe can serve as benchmark. 
Further recent trends towards explicit modelling of long-range data dependencies appear promising~\cite{gu2021statespace,nguyen2022s4nd,knigge2023modelling}, 
yet so far these models have only been benchmarked on synthetic data~\cite{linsley2018pathfinder,kim2019pathfinder2}, sequence data~\cite{tay2020lra},
and image classification~\cite{deng2009imagenet}, and thus, their potential for improving instance segmentation of long, complex, intertwined objects in real-world tasks has not been assessed.

Our work addresses the gap that, to date, solely synthetic data is available to facilitate methods development towards capturing long-range data dependencies.
To this end, we herewith release the FISBe dataset, a 3d multicolor light microscopy dataset of wide-ranging and tightly interweaving neuronal morphologies in the brain of the fruit fly Drosophila melanogaster, together with high quality expert instance segmentations of individual neurons.
The dataset comprises 101 large, expert-labeled 3d images, of which 30 are completely- and 71 partly labeled, with a total of $\sim$600 pixel-wise neuron instance masks. Exemplary images and instance masks are shown in Fig.~\ref{fig:showcases_data}. 
The novelty of our data entails a gap in evaluation metrics: Metrics commonly employed for benchmarking instance segmentation methodology do not appropriately account for the long, very thin and filamentous object shapes; 
e.g., mean average precision (mAP) with pixel-level IoU for localization is not appropriate for thin structures
~\cite{shit2021,maierhein2023}.
Thus standard metrics may not provide meaningful quantification of segmentation performance. 
To this end, we identify a set of informative evaluation metrics, and contribute a novel aggregate score that we recommend for method benchmarking.
Given our metrics, we evaluate three baseline methods, namely the two learnt methods that are, to our knowledge, technically able to handle the intricacies of our data to date~\cite{ppp2020,januszewski2018ffn}, as well as one non-learnt application-specific method that is technically applicable~\cite{duan2021}.
In summary, we contribute:
\begin{itemize}
    \item The FISBe dataset, to our knowledge the first public benchmark dataset for instance segmentation of real-world, wide-ranging, thin filamentous and tightly interweaving objects.
    \item A set of metrics and a novel ranking score for respective meaningful method benchmarking.
    \item An evaluation of three baseline methods in terms of the above metrics and score.
\end{itemize}
Concerning the size of our dataset, on the one hand, latest 2d natural image datasets are orders of magnitude bigger than ours~\cite{kirillov2023segment} and thus pave the way for particularly data-hungry methods development. 
Such size is, however, far beyond reach for 3d data, let alone for data from
the life sciences where expert knowledge is required for annotation, and acquiring pixel-wise ground truth for image data alike ours has been deemed difficult or infeasible in related work~\cite{friedmann2020mapping,duan2021}. 
On the other hand, numerous benchmark datasets similarly sized as FISBe have proven to greatly boost methods development in the machine learning community, and have likewise boosted respective application-specific scientific discovery~\cite{cremi,wei2021,ji2022amos}. 
We thus foresee our work to be of impact both in advancing the field of machine learning regarding methodology for capturing long-range data dependencies, and in streamlining cell-level analyses of brain function towards advances in basic neuroscience.
We release our data through zenodo (\href{https://zenodo.org/doi/10.5281/zenodo.10875063}{https://zenodo.org/doi/10.5281/zenodo.10875063}) and our project page \href{https://kainmueller-lab.github.io/fisbe}{https://kainmueller-lab.github.io/fisbe}.
\section{Dataset}\label{sec:dataset}
The FISBe dataset consists of 101 3d multicolor LM images of the central nervous system of the fruit fly Drosophila melanogaster. 
The images originate from a large pre-ex\-ist\-ing resource of LM acquisitions~\cite{meissner2023}.
Similar data has already contributed to breakthrough neuroscientific findings, e.g., towards a mechanistic understanding of memory formation and -retrieval in Drosophila on a cellular level~\cite{dolan2018communication}. 
Our work aims at facilitating such findings at scale.
For a more elaborate introduction to the biological background of our data, we refer the interested reader to Suppl.\ Sec.~\ref{suppl:bio_background}.
In the following, we describe the imaging resource FISBe stems from in Sec.~\ref{subsec:image_data}, the selection and annotation process for our dataset in Sec.~\ref{subsec:image_label}, and recommended data splits and evaluation for benchmarking in Sec.~\ref{subsec:benchmark_setup}.

\begin{figure}[t]
	\centering
		\begin{tabular}{m{0.2\columnwidth}m{0.2\columnwidth}m{0.2\columnwidth}m{0.2\columnwidth}}
			\begin{subfigure}[b]{0.24\columnwidth}
				\includegraphics[width=\columnwidth]{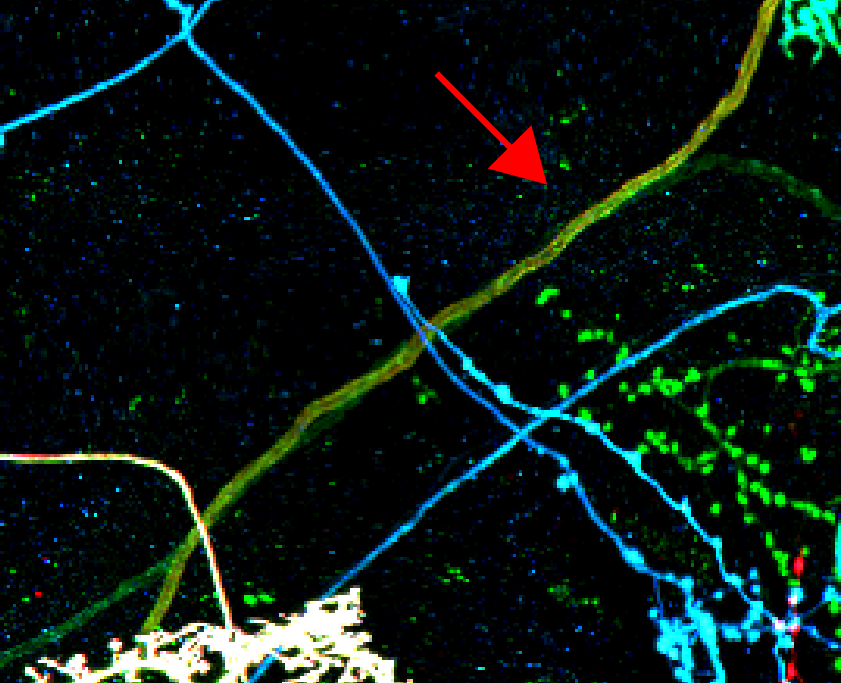}
			\end{subfigure}&
            \begin{subfigure}[b]{0.24\columnwidth}
				\includegraphics[width=\columnwidth]{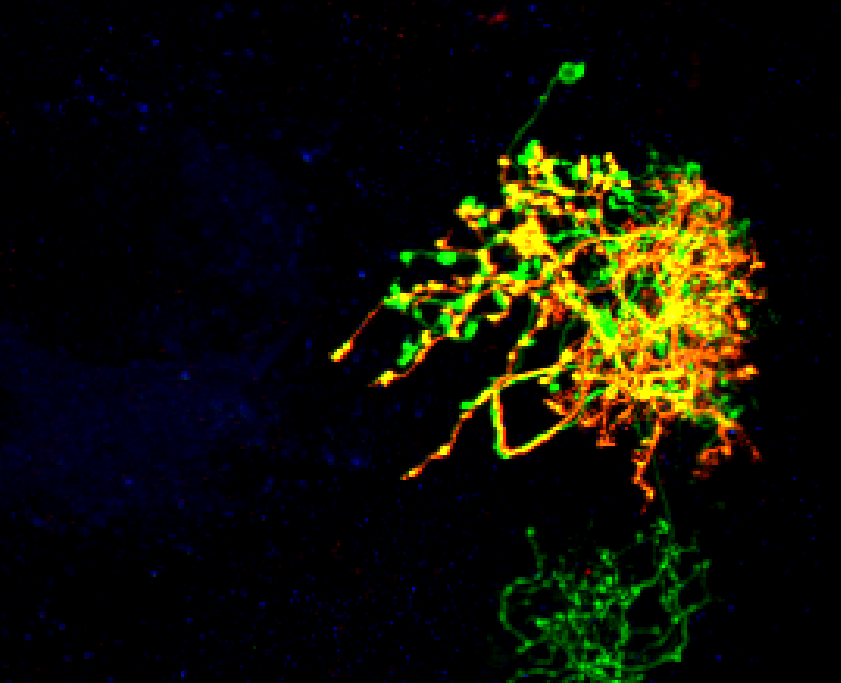}
			\end{subfigure}&
			\begin{subfigure}[b]{0.24\columnwidth}
				\includegraphics[width=\columnwidth]{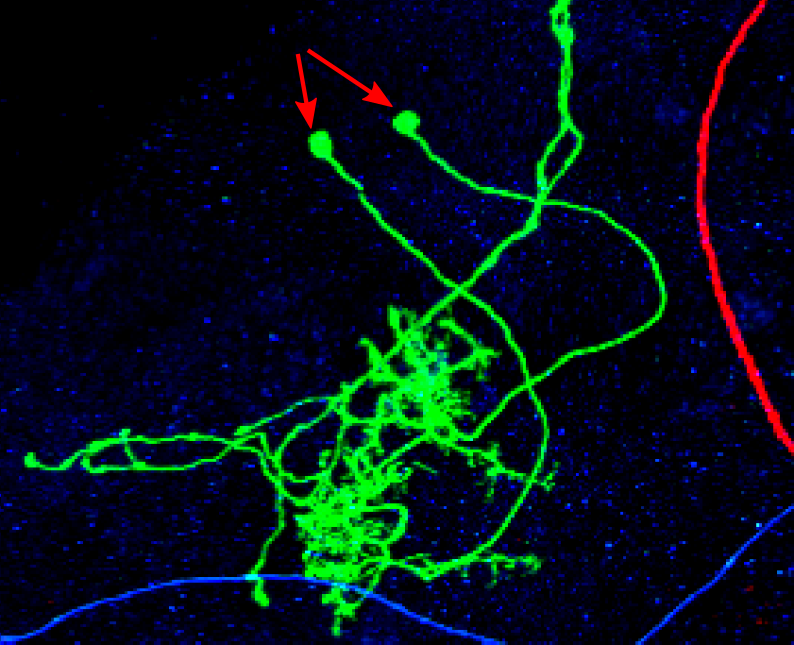}
			\end{subfigure}&
			\begin{subfigure}[b]{0.24\columnwidth}
				\includegraphics[width=\columnwidth]{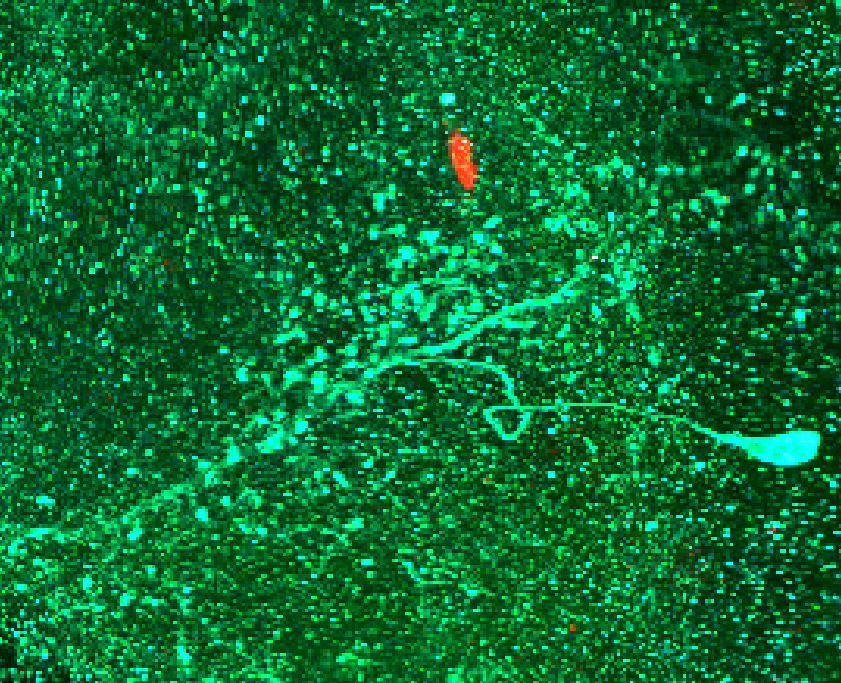}
			\end{subfigure}\\
			\begin{subfigure}[b]{0.24\columnwidth}
				\includegraphics[trim=250 220 165 245, clip, width=\columnwidth]{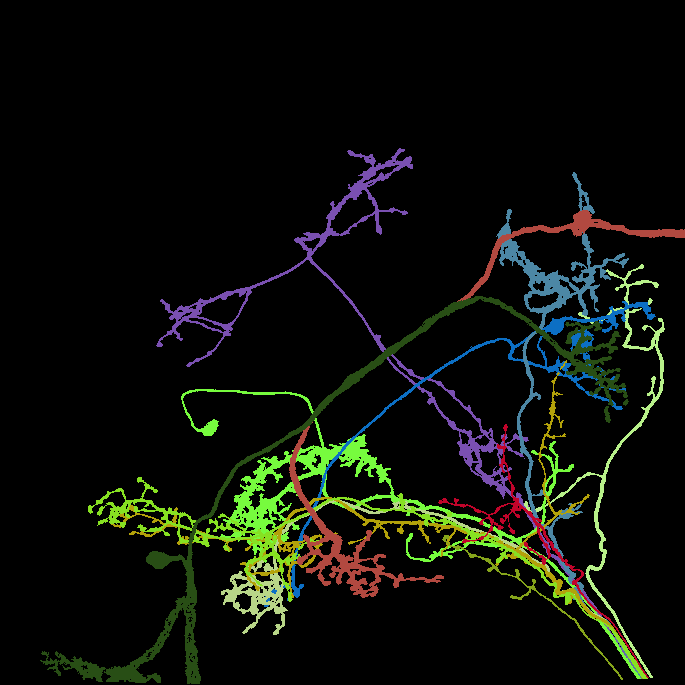}
			\end{subfigure}&
            \begin{subfigure}[b]{0.24\columnwidth}
				\includegraphics[trim=400 400 10 60, clip, width=\columnwidth]{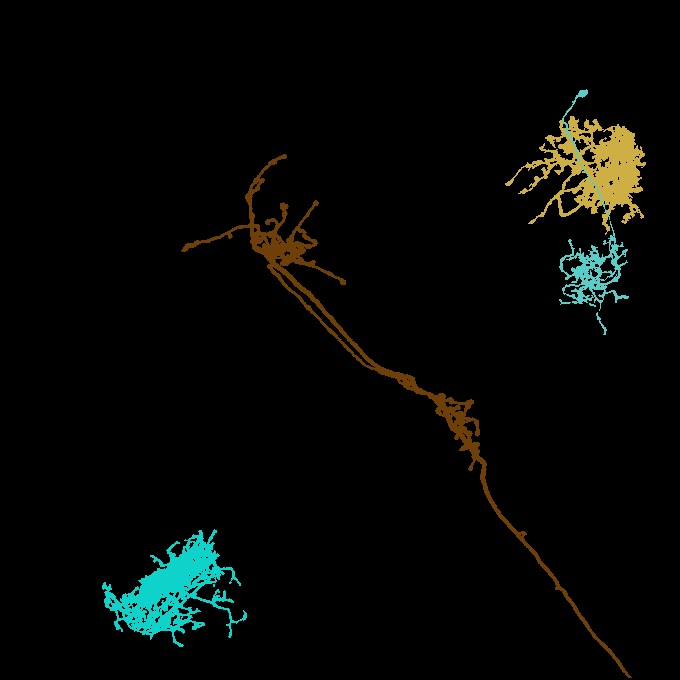}
			\end{subfigure}&
			\begin{subfigure}[b]{0.24\columnwidth}
				\includegraphics[trim=140 292 280 172, clip, width=\columnwidth]{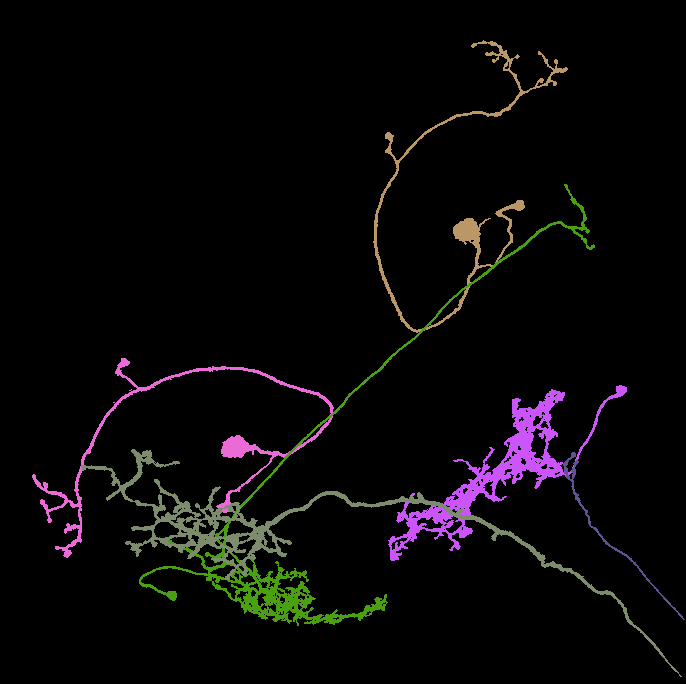}
			\end{subfigure}&
			\begin{subfigure}[b]{0.24\columnwidth}
				\includegraphics[trim=400 290 10 160, clip, width=\columnwidth]{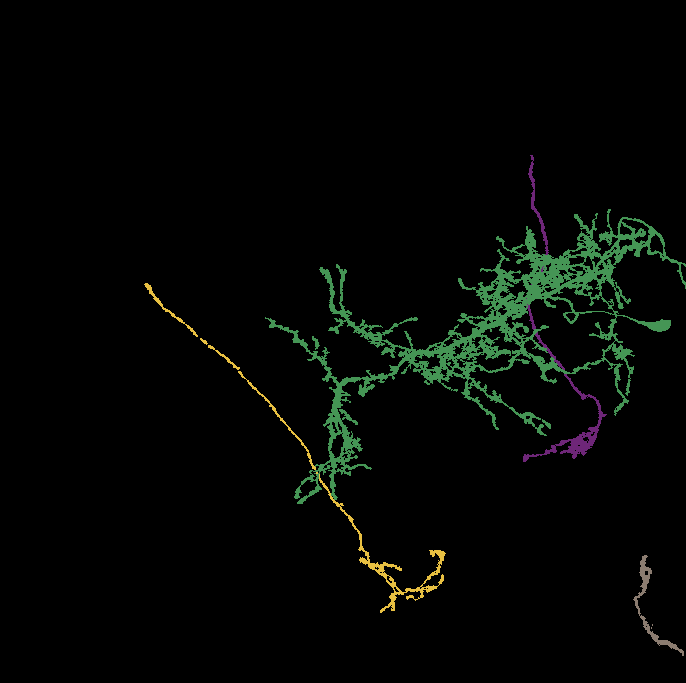}
			\end{subfigure}\\
		\end{tabular}
\vspace*{-11.5em}  
\textcolor{white}{\begin{flushleft}\raggedright \hspace{0.6em} (a) \hspace{4.4em} (b) \hspace{4.2em} (c) \hspace{4.2em} (d)\end{flushleft}}
\vspace*{7.8em}
		\caption{\label{fig:challenges}Exemplary challenging cases for disentangling neurons in FISBe images (top row), and respective expert annotations (bottom row). (a) Long overlap of two neurons running in parallel, (b) two almost completely overlapping neurons in different color (only one could successfully be annotated), (c) two inter-weaved neurons of same color that could not be separated (clearly identified by two somata), and (d) dim neuron in noisy background.}
\end{figure}

\subsection{Image Data Acquisition and Characteristics}\label{subsec:image_data}
Our dataset originates from the FlyLight project~\cite{meissner2023}, where
confocal microscopy images of the nervous systems of $\sim$74,000 flies were acquired with a technique called MultiColor FlpOut (MCFO)~\cite{mcfo_Nern2015}.
This image collection was previously released\footnote{Download (CC BY 4.0 license): \href{https://gen1mcfo.janelia.org}{https://gen1mcfo.janelia.org}}.
We selected images from the "40x Gen1" subset where
images have an isotropic resolution of $0.44 \mu m$, an average size of $\sim$400$\times$700$\times$700 pixels, and three color channels. 
Fig.~\ref{fig:showcases_data} (top row) shows exemplary MCFO images. Note that visualizations of image data shown in this paper are maximum intensity projections along the z axis, if not noted otherwise. For exemplary orthographic views see Suppl. Fig.~\ref{fig:ortho_view}.
For more information on the dataset see our datasheet~\cite{datasheet_gebru2018} in Suppl.\ Sec.~\ref{suppl:datset_doc}.

MCFO images capture the very thin, tree-like morphology of individual neurons as well as the intertwining of multiple neurons.
The number of neurons expressed in individual images varies from extremely dense ($>$50) to very sparse (1-2).
The FlyLight project has sorted images into five categories according to expression density, where 18\% of images express up to 10 neurons (cat.\,1 and 2) and 
55\% express around 20 neurons (cat.\,3, cf.\ Suppl.\ Fig.\ 1 in ~\cite{meissner2023}).

MCFO imagery has sparse, unbalanced foreground signal, low signal-to-noise ratio, and exhibits artifacts like broken structures and intensity shifts.
Intensity varies strongly, per image, per neuron and within neurons. Thus neurons may appear in very different quality, ranging continuously from clearly visible neurons to very dim neurons that are partly indistinguishable from noise.
The MCFO technique causes individual neurons to exhibit random colors, though color diversity per image is often not sufficient to allow for distinguishing all neurons by color. 
Multiple neurons in very close proximity may appear as overlapping due to partial volume effects\footnote{Multiple instances occupy the same 3d pixel (distinct from occlusion)}. 
Neurons in MCFO images are particularly hard to distinguish if neurons of same or similar color form dense clusters or overlap (see Fig.~\ref{fig:challenges}).

\subsection{Image Selection and Labeling Process}\label{subsec:image_label}
Labeling 3d data is generally cumbersome as objects often need to be viewed with different angles, scales and color settings. 
As for the FlyLight data, limited resolution inherent in confocal microscopy makes isolating individual similarly colored, close-by neurons particularly difficult. Furthermore, in some cases poor signal-to-noise ratio makes it difficult to identify the complete wide-spanning extent of neurons. In both cases, expert anatomical knowledge of fruit fly neurons is often crucial for successful annotation.

To form FISBe, expert annotators chose 101 samples from the FlyLight data for which they determined by eye that manual annotation is feasible. Compared to the full FlyLight resource, this introduces a bias in our dataset towards sparser expression densities:
Of our 101 images, one image is of density cat.\,1, 72 are of cat.\,2, and 28 of cat.\,3. 
Two expert annotators manually segmented and proof-read each other to label as many neurons as possible in these 101 images using the interactive rendering tool VVD Viewer\cite{vvd_viewer}.
Annotators were able to segment a total of 590 neurons. Labeling a single neuron took 30-60 min on average, yet for a difficult neuron it could take up to 4 hours.
Not every neuron in every sample could be annotated successfully, thus yielding completely- as well as partly labeled images. 
A third annotator performed a final visual inspection of all labeled neurons and revised the categorization into completely- and partly labeled images.

Our completely labeled dataset comprises 30 MCFO images and a total of 139 labeled neurons;  
see Fig.\ \ref{fig:showcases_data} (1st and 2nd example) and Suppl.\ Fig.\ \ref{suppl_fig:samples_completely}.
Our partly labeled dataset comprises 71 MCFO images and a total of 451 labeled neurons; see Fig.\ \ref{fig:showcases_data} (3rd and 4th example) and Suppl.\ Fig.\ \ref{suppl_fig:samples_partly}. 
These images exhibit unlabeled neuronal morphologies because expert annotators were either unable to disentangle multiple neurons of the same color in a dense cluster, or unable to annotate very dim neurons that are partly indistinguishable from noisy background. 
Note, 61 images contain labeled neurons that overlap due to partial volume effects.

Complementing our new annotated data, the large trove of previously released non-annotated images in the FlyLight resource may serve for self-supervised pre-training.

\subsection{Benchmarking Setup}\label{subsec:benchmark_setup}
We split the completely labeled data into train, validation and test sets with 18, 5 and 7 samples respectively as defined in Suppl.\ Table \ref{tab:samples_completely_labeled}. 
We split the partly labeled data into train, validation and test sets with 43, 12 and 16 samples respectively as defined in Suppl.\ Table \ref{tab:samples_partly_labeled}. 
We recommend evaluation on the combined data (i.e., the union of completely and partly labeled data) as the main benchmarking scenario.
To assess training stability, we recommend to report summary statistics over three training runs in each evaluated scenario.

\section{Evaluation Metrics}\label{sec:metrics}

Our dataset constitutes the first benchmark dataset for \textit{instance} segmentation of thin filamentous structures. Consequently, we need to assess which evaluation metrics are suitable for benchmarking. 
The main requirements for a suitable metric are: (r1) To account for thin filamentous structures, (r2) to be able to handle overlapping instances (both in ground truth and predicted instances), and (r3) to be meaningful with respect to downstream tasks.

Some existing benchmarks for \textit{semantic} segmentation of filamentous structures have employed topology-based metrics, which assess the similarity between graph representations of ground truth- and predicted objects~\cite{gillette2011,roadtracer}.
We deem these not suitable for our data, as obtaining topologically correct (tree) graph representations of neurons is infeasible due to the limited resolution of light microscopy.

Instead, we follow the Metrics Reloaded~\cite{maierhein2023} recommendation for thin filamentous instance segmentation
and apply an instance-level F1 score as one of our main evaluation metrics.
Moreover, we propose to complement the F1 score with a custom metric that we design towards satisfying r3, namely a centerline recall with one-to-many matching, which we refer to as \emph{average ground truth coverage}.
We combine both metrics to derive an aggregate benchmark score.
Finally, we define a set of easily interpretable error measures that may provide additional insight to methods developers and practitioners.
E.g., we extend existing work~\cite{fsfm_caicedo2019,tra_matula15} by defining false split (FS) and false merge (FM) error counts for overlapping instances.
We define our selected metrics in Sec.\ \ref{subsec:score}, ensuring that all apply not only to completely- but also to partly labeled data, and discuss properties and suitability in Sec.\ \ref{subsec:eval_discussion}.
Suppl. Table~\ref{tab:metric_overview} summarizes all metrics with their localization and matching.

\subsection{Metrics Definitions}\label{subsec:score}
Instance segmentation can be phrased as a pixel labeling problem, where pixels with same label form instances.  Note that in FISBe, one pixel can be assigned multiple labels due to overlapping instances.
Segmentation quality is generally assessed via evaluation metrics that capture how well predicted instances overlay with given ground truth (gt) instances. 
We denote a set of gt instances $G = \{g_k\}_{k\in L_G}$ and a set of predicted instances $P = \{p_l\}_{l\in L_P}$ where $L_G$ and $L_P$ represent the sets of labels identifying gt and predicted instances, and $|G|$ and $|P|$ denote the total number of instances of the respective set.
With subscript \(i\) the set is limited to a single image, e.g., \(G_i\), otherwise it refers to the set over all images \(I\).
Gt- and prediction sets exclude background (bg) as instance label if not stated otherwise.

\noindent\textbf{Average F1 score \(avF1\).}
Following \cite{maierhein2023}, a metric consists of three steps: localization, matching and computation.
The localization step employs some function to compute how well each pair of predicted and gt instances are co-localized.
The matching step selects subset of these pairs, resulting in a match of predicted to gt instances.
The computation step computes the value of the metric based on the quality of the previously computed subset of matched instances.

It has been shown that pixel-level IoU or F1 are not suitable for thin structures as small variations on boundaries can have a large effect \cite{maierhein2023,pqreview_foucart2023}.
Thus we employ clDice~\cite{shit2021}, a variation of the Dice score that operates on object centerlines, for the localization step.
Following \cite{shit2021}, we use medial surface axial thinning algorithm~\cite{skel_lee1994} to skeletonize volumetric instance masks and denote it with function $skeletonize(\cdot)$.
Given ground truth and prediction we compute clDice as follows:
\begingroup
\allowdisplaybreaks
\begin{align}
\begin{split}%
    &\text{clPrecision}(p, g) = \frac{|\text{skeletonize}(p) \cap  g|}{|\text{skeletonize}(p)|}\\%
    &\quad\quad\quad\forall p \in P_i, \forall g \in G_i \cup \{bg\}%
\end{split}\\%
\begin{split}%
    &\text{clRecall}(g, p) = \frac{|\text{skeletonize}(g) \cap p|}{|\text{skeletonize}(g)|}\\%
    &\quad\quad\quad\forall g \in G_i, \forall p \in P_i \cup \{bg\}%
\end{split}\\%
\begin{split}%
    &\text{clDice}(g, p) = 2 * \frac{\text{clPrecision}(p, g) * \text{clRecall}(g, p)}{\text{clPrecision}(p, g) + \text{clRecall}(g, p)}\\%
    &\quad\quad\quad\forall g \in G_i, \forall p \in P_i%
\end{split}%
\end{align}
\endgroup
clDice is only computed for foreground pairs as we do not skeletonize the background, but we, e.g., include it for clPrecision to detect predictions mainly located in the gt background.

For the matching step we follow the greedy strategy recommended in \cite{maierhein2023}. To this end we compute clDice for all pairs of predicted and gt instances \(\{\text{clDice}(p,g) ~|~ \forall i \in I: \forall p \in P_i, \forall g \in G_i\}\).
We iterate through all scores in descending order and match the corresponding \((p,g)\)-pair if neither has been assigned before.
Similarly to \cite{tra_matula15}, we denote $p \Subset g$, if the two instances have been matched.
In the computation step, we derive true positives (TP), false positives (FP) and false negatives (FN) for all clDice thresholds \(th\) in the range [0.1,0.9] with step size 0.1:
\begin{itemize}
    \setlength\itemsep{0.5em}
    \item TP: all predicted instances that are assigned to a gt instance with clDice $>$ \(th\):
    \[\text{TP} = |\{p \in P ~|~ \exists g: p \Subset g \land \text{clDice}(p,g)>\text{th}\}|\]
    \item FP: all unassigned predicted instances $\text{FP} = |P| - \text{TP}$
    \item FN: all unassigned gt instances $\text{FN} = |G| - \text{TP}$
\end{itemize}
\vspace{0.5em}
Based on these values we compute the F1 score $F1=\frac{2\text{TP}}{2\text{TP}+\text{FP}+\text{FN}}$ for each threshold. Note that TP, FP, FN are thus computed across all images.
The final \textit{avF1} score is the average of all F1 scores.

\noindent\textbf{Average ground truth coverage \(C\).}
We compute clPrecision scores for all pairs of predicted and gt instances as localization criterion.
We match each prediction to the gt instance with the highest clPrecision score (one-to-many matching).
Then we average clRecall for all gt instances and the union of their matched predictions (to avoid double-counting of pixels with overlapping predictions):
\begin{equation}
    C = 1/|G| \sum_{g \in G} \text{clRecall}(g, \bigcup_{\forall p \in P: p \Subset g} p)
\end{equation}

\noindent\textbf{Aggregate benchmark score \(S\).}
We propose to combine the average F1 score $avF1$ and the average ground truth coverage $C$ to form a primary benchmark ranking score \(S\).
We average both measures to obtain the final ranking score: $S = 0.5 avF1 + 0.5 C$.
Note that we do not multiply them, as a linear increase in segmentation quality should lead to linear increase in the score function \cite{pqreview_foucart2023}.

\noindent\textbf{False splits \(FS\) and false merges \(FM\).} 
False split errors occur if one gt instance is covered by multiple predicted instances.
False merge errors occur if one predicted instance covers more than one gt instance.
We propose to use a greedy many-to-many matching algorithm that naturally handles overlapping instances and based on which we can compute FM and FS directly in a unified way.
For the matching, we iteratively assign predicted and gt instances with the highest clRecall value while keeping track of already matched pixels (see Algorithm~\ref{alg:mn_matching}).
Remaining clRecall values are constantly updated to only include \textit{free} pixels, which are available for further matching.
By doing so, we avoid that predicted instances in overlapping gt regions are assigned multiple times; or that mostly overlapping predicted instances are assigned to the same gt instance (see Suppl. Fig.~\ref{fig:testcases_FM_FS}).
Note that we monitor centerline pixels for gt instances and complete pixelwise masks for predicted instances due to the definition of clRecall.

\begingroup
\SetArgSty{textnormal}
\newcommand{\pluseq}{\mathrel{+}=}
\begin{algorithm}[t]
\SetKwInOut{Input}{input}\SetKwInOut{Output}{output}
\DontPrintSemicolon
\Input{$G$: set of gt instances $g_k$, \\
$P$: set of predicted instances $p_l$, \\
th: clRecall threshold}
\Output{$M$: set of matched $(g_k, p_l)$-instances}
\BlankLine
initialize $M \leftarrow \emptyset$\;
initialize $G_{\text{free}} \leftarrow \{\text{skeletonize}(g_k) ~|~ \forall g_k \in G\}$\; 
initialize $P_{\text{free}} \leftarrow$ copy($P$)\;
$\text{clR} \leftarrow \text{sort}(\{\text{clRecall}(\text{skeletonize}(g_k), p_l) ~|~ $\\ 
\hspace{8em}$\forall g_k \in G, \forall p_l \in P\}, \downarrow)$\;
\While{top(clR) $>$ th}{
    $g_k, p_l \leftarrow \text{pop}(\text{clR})$\;
    $M \pluseq \{(g_k, p_l)\}$\;
    update $g_{\text{free}_{k}} \leftarrow g_{\text{free}_{k}} \setminus p_l$\;
    update $p_{\text{free}_{l}} \leftarrow p_{\text{free}_{l}} \setminus g_k$\;
    \ForAll{ $(g_m, p_n) \in \text{clR}$}{
    \If{$g_m = g_k$}{
        update $\text{clR}(g_m, p_n) = \frac{|g_{\text{free}_{m}} \cap p_n|}{|\text{skeletonize}(g_m)|}$
    }
    \If{$p_n = p_l$}{
        \mbox{update $\text{clR}(g_m, p_n) = \frac{|\text{skeletonize}(g_m) \cap p_{\text{free}_{n}}|}{|\text{skeletonize}(g_m)|}$}
    }
    }
    $\text{sort}(\text{clR}, \downarrow)$\;
}
\caption{Greedy Many-to-many Matching}
\label{alg:mn_matching}
\end{algorithm}
\endgroup

We then count for each gt instance the additional number of assigned predicted instances apart from one correctly matched instance to compute false splits (with \(\text{th}=0.05\))
\begin{equation}
    \text{FS} = \sum_{g \in G} \text{max}((\sum_{p \in P: p \Subset g} 1 ) - 1, 0),
\end{equation}
and analogously for false merges (with \(\text{th}=0.1\))
\begin{equation}
    \text{FM} =  \sum_{p \in P} \text{max}((\sum_{g \in G: g \Subset p} 1 ) - 1, 0).
\end{equation}

\noindent\textbf{True positives clDice.}
We report average centerline Dice for uniquely matched instances \(\text{clDice}_{\text{TP}}\) to provide a measure of how well true positives
are reconstructed.
We re-use the matching computed for $avF1$, employ a threshold of 0.5 for the definition of TP and define
\begin{equation}
    \text{clDice}_{\text{TP}} = \frac{1}{|\text{TP}_{0.5}|} \sum_{(p, g): \\p \in \text{TP}_{0.5} \land p \Subset g} \text{clDice}(p, g).
\end{equation}

\noindent\textbf{Evaluating partly labeled samples.}
In partly labeled samples only a subset of neurons is annotated.
For unlabeled pixels we do not know if there is background or other neuronal structures and for labeled neurons if they partly overlap with a non-annotated one.
This has no influence on average ground truth coverage as well as false split and false merge counts, although the reported measures only reflect parts of the whole volume.
However, for the F1 score the false positive count cannot be computed.
Therefore, we only count predicted instances that are not one-to-one matched based on clDice, but that primarily lie within a gt instance and not the background:
\begin{align}
    \text{FP}_{\text{partly}} &= |\{ p \in P ~|~ \nexists g: p \Subset g \land \\
    &\argmax_{g \in G \cup \{\text{bg}\}} \text{clPrecision}(p, g) \neq \text{bg}\}|\nonumber
\end{align}
We use this error count as an approximation of false positives and adapt the formula to $F1=\frac{2\text{TP}}{2\text{TP}+\text{FP}_{\text{partly}}+\text{FN}}$.
All other calculation steps remain unchanged.
Note that thus the F1 scores of completely and partly labeled image sets cannot be compared directly.
To evaluate the full dataset, we average the results for the completely and partly labeled set (for normalized measures, counting measures such as FS, FM and TP are summed up).

\noindent\textbf{Evaluating challenging cases.}
Main challenges are dim and overlapping neurons (cf. Fig.~\ref{fig:challenges}).
To evaluate how well such subsets of neurons are segmented, 
we report gt coverage \(C\) and 
the relative number of TPs (\(\text{tp} =\frac{\text{TP}_{S,0.5}}{|G_S|}\), with greedy one-to-one matching and clDice $>$ 0.5) for the respective subsets ($G_{\text{dim}}$/$G_{\text{ovlp}}$).
Dim neuron instances of validation and test sets are listed within the dataset.

\subsection{Discussion}\label{subsec:eval_discussion}

\begin{figure*}
    \centering
    \includegraphics[width=0.92\textwidth]{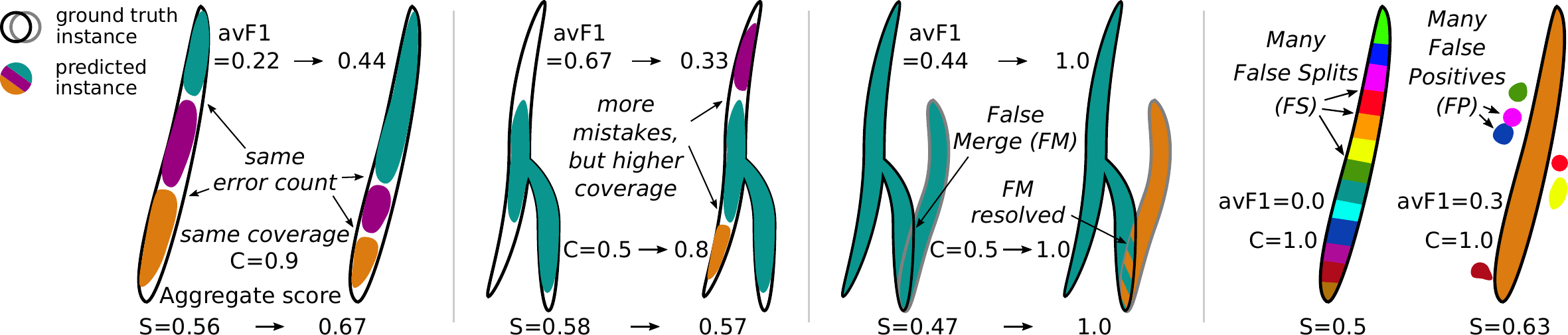}
    \vspace*{-12em}  
    \textcolor{black}{\begin{flushleft}\raggedright \hspace{10em} (a) \hspace{8em} (b) \hspace{9em} (c) \hspace{10.2em} (d)\end{flushleft}}
    \vspace*{9em}
    \caption{Visualization of segmentation examples to assess suitable evaluation metrics:
      (a) Depending on the split position, avF1 can vary significantly at identical gt coverage and error count.
      (b) Using the avF1 score alone would favor lower coverage over more false split errors. This might be disadvantageous in downstream analysis tasks.
      In (c) resolving the merge leads to a large improvement in the overall score.
      In (d) both cases achieve a perfect score wrt.\ the gt coverage C. By penalizing FP and FS errors in the F1 score the limitations of these predictions are reflected in the overall score.
      For more edge cases and the full quantitative numbers please see Suppl. Fig.~\ref{fig:edge_cases}.
      \label{fig:testcases}}
\end{figure*}

We satisfy requirement (r1) by using clPrecision, clRecall and clDice in all of our measures. 
They are defined in such a way that they handle overlaps in both predicted and gt instances, thus satisfying requirement (r2).

The $avF1$ score considers all error types equally.
However, FP and FS errors are more likely to occur in MCFO segmentations due to low signal-to-noise ratio and broken structures.
They are often induced by only a limited number of incorrect pixels, whereas FN errors are limited by the number of neurons and require that all or large parts of neurons are missing.
This can lead to disproportionately low scores that do not reflect how we would visually rate segmentation quality (see Fig.~\ref{fig:testcases}~(a)+(b)).

To mitigate this, we complement the $avF1$ score by the average gt coverage \(C\),
resulting in an improved balancing of the different types of errors.
$C$ provides an intuitive way for measuring how comprehensively gt instances are segmented by the model.
$C$ strongly penalizes FN and FM errors, an important property for downstream tasks.
Consider the edge case of a perfect foreground segmentation.
If FM are not penalized, assigning the same label to each connected component would lead to a perfect score, despite merges occurring at every overlap and point of contact.
Thus, as desired, if a model achieves to split a previously occurring FM correctly, this will lead to a large improvement as an additional gt instance will get matched (see Fig.~\ref{fig:testcases}~(c)).

However, $C$ does not incorporate FP and FS errors. 
Consider the same edge case as before, but now we assign a separate instance label to each foreground pixel.
This, as a consequence of the one-to-many matching still leads to a perfect score for $C$.
But the $avF1$ score will be exceedingly low.
Similarly if the gt instances are indeed segmented perfectly, but the outside noise is segmented as well (see Fig~\ref{fig:testcases}~(d)).
Note that this highlights why $C$ should not be used as a standalone metric. FP and FS errors are accounted for by the \(avF1\) score, and \(C\) penalizes FN and FM errors. Thus $C$ and $avF1$ nicely complement each other. 
For additional edge cases and the corresponding quantitative evaluation please see Suppl. Fig.~\ref{fig:edge_cases}.

Concerning (r3) there are two highly relevant downstream tasks we took into account: 1. Morphological analysis of neurons, and 2. the task of searching for a given neuronal morphology within all MCFO images.
For the first task neurons of interest need to be reconstructed in their entirety.
In this regard,
FN and only partly annotated neurons are critical, as their manual curation is very time-consuming.
FP, FM and FS are less relevant as, up to a point, they can relatively easily be corrected with a few clicks.
This importance of completeness is strongly reflected in $C$.
For the second task,
the above considerations still hold, except that FP are more problematic.
This is accounted for by the F1 score, which, as the number of FP can often be much higher than the number of TP and FN (see Table~\ref{tab:results_both_sets}), heavily penalizes such cases.

Overall, our benchmark score $S$ is in accordance with all our requirements.
However, as a single benchmark score most likely cannot do justice to every possible use case, we report additional metrics that can be resorted to for alternative downstream tasks as well as for considerations on method improvements.
The additional metrics all satisfy r1 and r2.
Finally, our proposed many-to-many matching algorithm is very generic, despite adhering to our quite specific requirements.
There is no special treatment of overlapping regions which strongly facilitates the matching.
The algorithm can be applied to other object shapes by replacing clRecall with other overlap-based metrics like IoU or IoR.


\section{Baselines}\label{sec:baseline}

\begin{table*}[bpht]
	\centering
    \caption{\label{tab:results_both_sets}Quantitative results on the full FISBe validation and test sets (i.e., completely and partly labeled data combined; for results on the respective subsets see Suppl.\ Table~\ref{suppl_tab:results_both_sets}). We compare PatchPerPix (ppp, \cite{ppp2020}), Flood Filling Networks (FFN, \cite{januszewski2018ffn}) trained on the completely labeled and the full training dataset (+partly) and Duan et al.'s color clustering \cite{duan2021}. We report mean and standard de\-vi\-a\-tion (\std{}) over three independent runs (except for Duan et al.'s as it is non-learnt). For all scores except FS and FM higher values are better.}
    \begin{tblr}{width=\linewidth,rows={abovesep=1pt,belowsep=1pt},colspec={X[0.4,l] X[1.4,l] J[1.2] J[1.2] J[1.2] J[1.2] J[1.1] J[1.0] J[1.2] J[1.2] J[1.2] J[1.2] J[1.2]},row{1}={guard}}
        \midrule
		  Split & Method & \(S\) & \(avF1\) & \(C\) & \(clDice_{TP}\) & \(FS\) & \(FM\) & \(C_{dim}\) & \(C_{ovlp}\) & \(tp\) & \(tp_{dim}\)  & \(tp_{ovlp}\)\\
        \midrule
		\SetCell[r=4]{m}Val & ppp & 0.38\std{0.02} & 0.41\std{0.02} & 0.35\std{0.01} & 0.75\std{0.02} & 6.0\std{0.8} & 24\std{1.6} & 0.12\std{0.01} & 0.38 \std{0.04} & 0.46 \std{0.01} & 0.16 \std{0.04} & 0.39 \std{0.03} \\
        & FFN & 0.25\std{0.01} & 0.27\std{0.01} & 0.23\std{0.01} & 0.79\std{0.01} & 7.0\std{2.9} & 12\std{2.0} & 0.03\std{0.01} & 0.30 \std{0.01} & 0.32 \std{0.01} & 0.04 \std{0.01} & 0.37 \std{0.02} \\
        & FFN+partly & 0.27\std{0.01} & 0.29\std{0.01} & 0.24\std{0.01} & 0.79\std{0.01} & 7.7\std{2.6} & 14\std{0.8} & 0.02\std{0.01} & 0.33 \std{0.02} & 0.34 \std{0.03} & 0.03 \std{0.00} & 0.38 \std{0.04} \\
        & \mbox{Duan et al.} & 0.24 & 0.26 & 0.22 & 0.70 & 14 & 13 & 0.02 & 0.28 & 0.37 & 0.03 & 0.42 \\
        \midrule
        \SetCell[r=4]{m}Test & ppp & 0.35\std{0.00} & 0.34\std{0.01} & 0.35\std{0.01} & 0.80\std{0.00} & 19\std{2.9} & 52\std{3.4} & 0.16\std{0.03} & 0.27 \std{0.04} & 0.36 \std{0.01} & 0.19 \std{0.04} & 0.19 \std{0.03} \\
        & FFN & 0.25\std{0.03} & 0.22\std{0.04} & 0.29\std{0.02} & 0.80\std{0.01} & 17\std{1.7} & 39\std{5.3} & 0.03\std{0.01} & 0.26 \std{0.03} & 0.32 \std{0.03} & 0.00 \std{0.00} & 0.24 \std{0.05} \\
        & FFN+partly & 0.27\std{0.01} & 0.24\std{0.02} & 0.31\std{0.00} & 0.80\std{0.01} & 18\std{3.7} & 36\std{3.6} & 0.04\std{0.01} & 0.28 \std{0.01} & 0.36 \std{0.01} & 0.03 \std{0.00} & 0.28 \std{0.01} \\
        & \mbox{Duan et al.} & 0.30  & 0.27 & 0.33 & 0.77 & 45 & 29 & 0.03 & 0.36 & 0.37 & 0.03 & 0.34 \\
		\bottomrule
	\end{tblr}
    \begin{tblr}{width=\linewidth,rows={abovesep=1pt,belowsep=1pt},colspec={X[1.5,l] J[1.2] J[1.2] J[1.2] J[1.2] J[1.2] J[1.2] J[1.2] J[1.2] J[1.2]},row{1}={guard}}
        \midrule
		  Test & F1$_{0.1}$ & F1$_{0.2}$ & F1$_{0.3}$ & F1$_{0.4}$ & F1$_{0.5}$ & F1$_{0.6}$ & F1$_{0.7}$ & F1$_{0.8}$ & F1$_{0.9}$\\
		\midrule
        ppp & 0.50\std{0.01} & 0.48\std{0.01} & 0.44\std{0.01} & 0.41\std{0.02} & 0.35\std{0.02} & 0.29\std{0.02} & 0.26\std{0.01} & 0.19\std{0.02} & 0.12\std{0.01}\\
        FFN & 0.34\std{0.05} & 0.31\std{0.04} & 0.28\std{0.04} & 0.25\std{0.05} & 0.22\std{0.04} & 0.20\std{0.04} & 0.17\std{0.03} & 0.12\std{0.01} & 0.07\std{0.01}\\
        FFN+partly & 0.36\std{0.02} & 0.32\std{0.02} & 0.30\std{0.02} & 0.27\std{0.03} & 0.25\std{0.03} & 0.21\std{0.03} & 0.18\std{0.02} & 0.15\std{0.02} & 0.09\std{0.01}\\
        Duan et al. & 0.43 & 0.38 & 0.35 & 0.33 & 0.31 & 0.29 & 0.20 & 0.12 & 0.06\\
		\bottomrule
	\end{tblr}
\end{table*}

\begin{figure*}[htbp]
	\centering%
    \begin{subfigure}[b]{\textwidth}%
    \hfill%
    \begin{subfigure}[b]{0.19\textwidth}%
        \includegraphics[width=\textwidth]{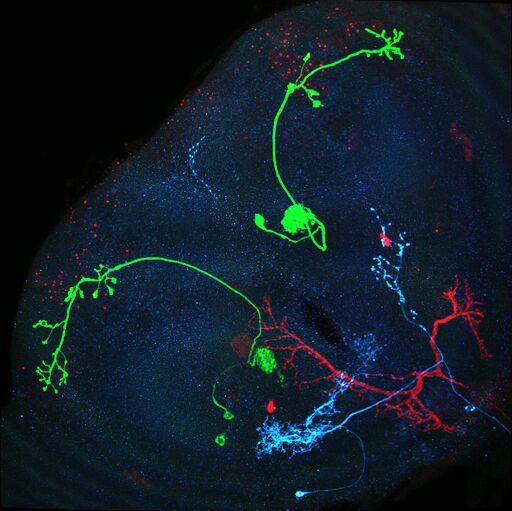}%
    \end{subfigure}%
    \hfill%
    \begin{subfigure}[b]{0.19\textwidth}%
        \includegraphics[width=\textwidth]{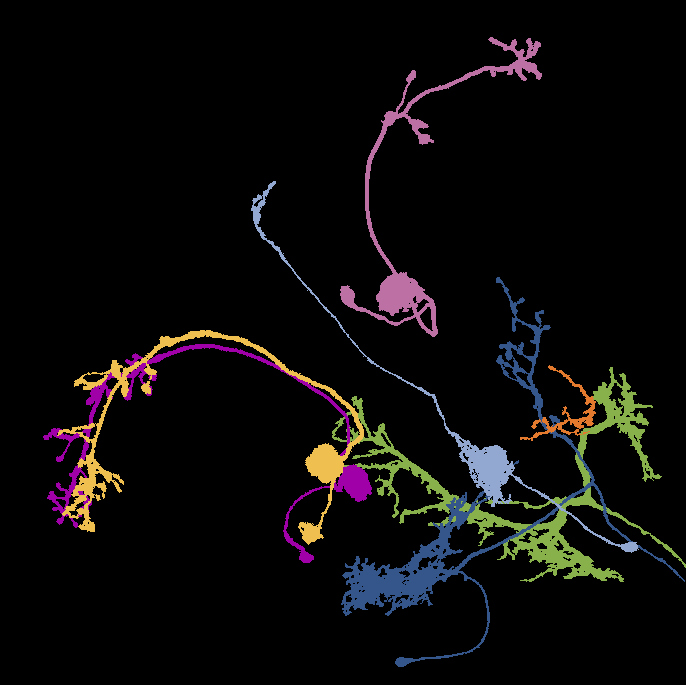}%
    \end{subfigure}%
    \hfill%
    \begin{subfigure}[b]{0.19\textwidth}%
        \includegraphics[width=\textwidth]{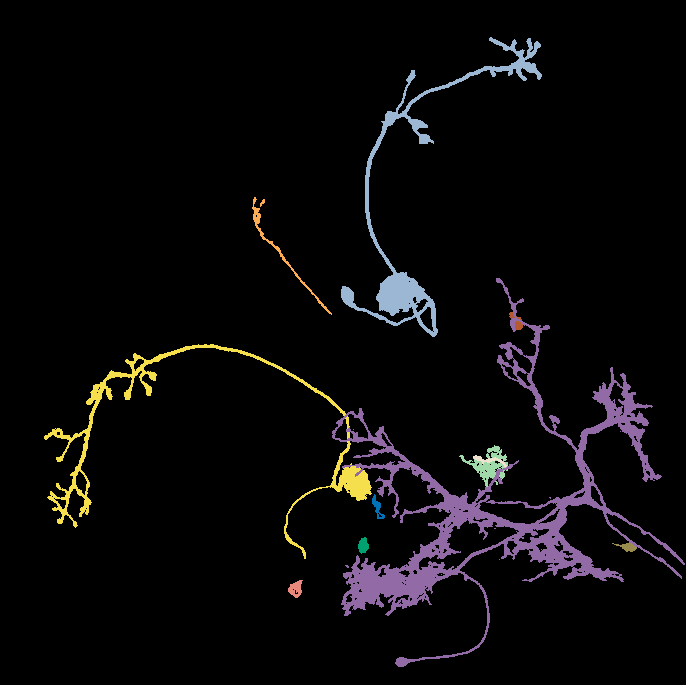}%
    \end{subfigure}%
    \hfill%
    \begin{subfigure}[b]{0.19\textwidth}%
        \includegraphics[width=\textwidth]{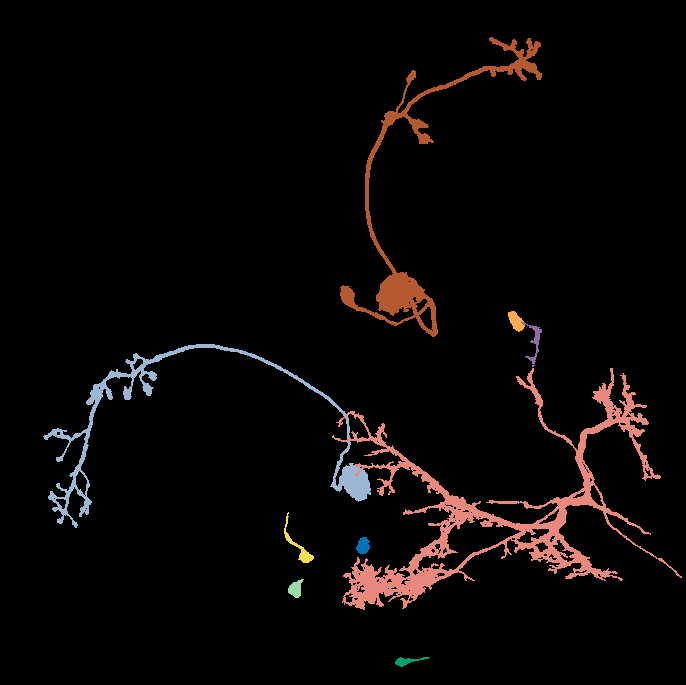}%
    \end{subfigure}%
    \hfill%
    \begin{subfigure}[b]{0.19\textwidth}%
        \includegraphics[width=\textwidth]{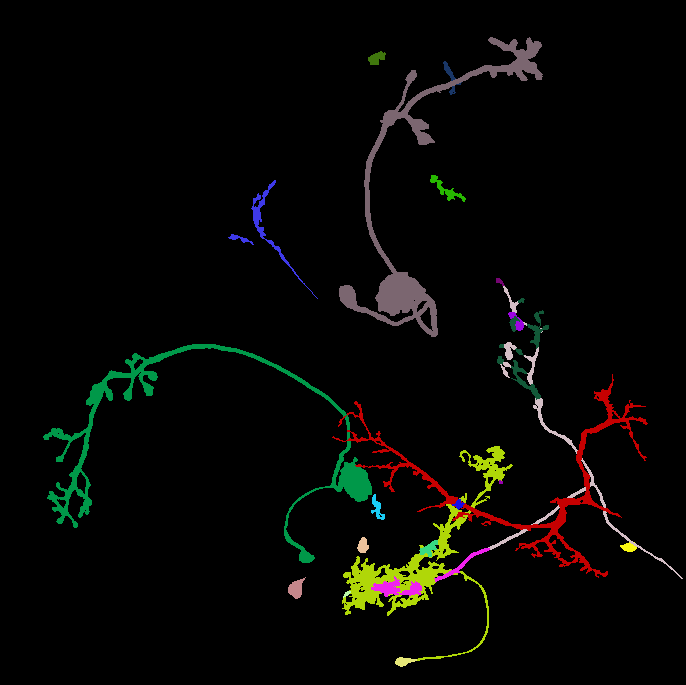}
    \end{subfigure}%
    \hfill%
    \hfill%
    \end{subfigure}%

    \begin{subfigure}[b]{\textwidth}%
    \hfill%
    \begin{subfigure}[b]{0.19\textwidth}%
        \includegraphics[width=\textwidth]{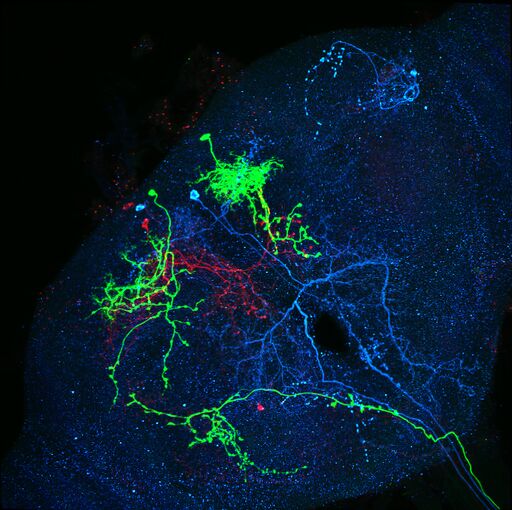}%
        \phantomsubcaption\label{fig:qual_res:mcfo}%
    \end{subfigure}%
    \hfill%
    \begin{subfigure}[b]{0.19\textwidth}%
        \includegraphics[width=\textwidth]{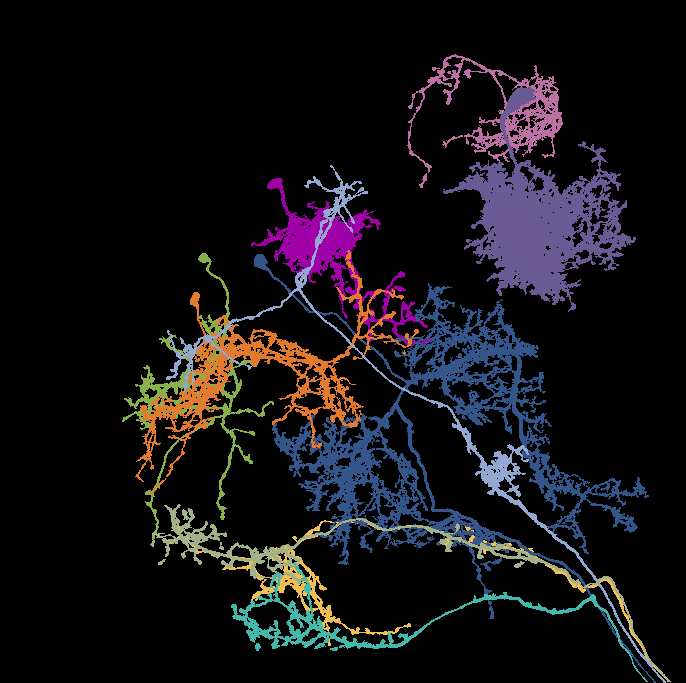}%
        \phantomsubcaption\label{fig:qual_res:gt}%
    \end{subfigure}%
    \hfill%
    \begin{subfigure}[b]{0.19\textwidth}%
        \includegraphics[width=\textwidth]{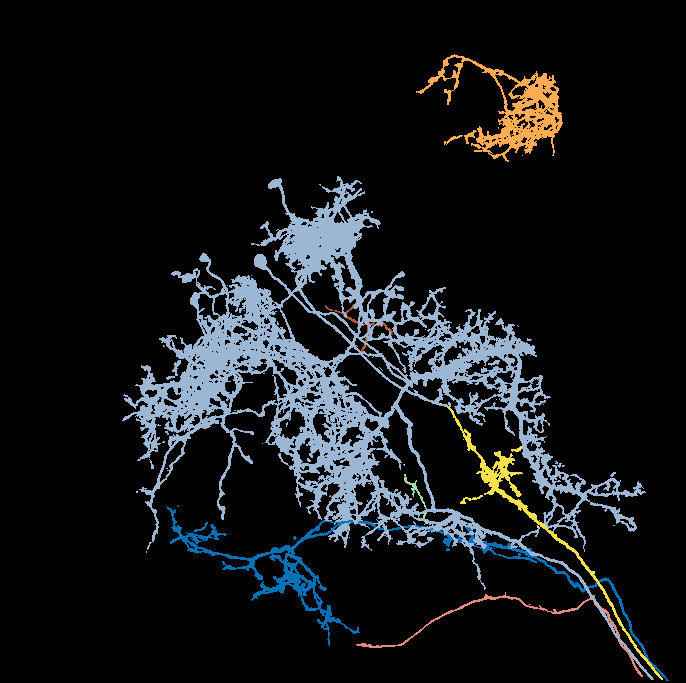}%
        \phantomsubcaption\label{fig:qual_res:ppp}%
    \end{subfigure}%
    \hfill%
    \begin{subfigure}[b]{0.19\textwidth}%
        \includegraphics[width=\textwidth]{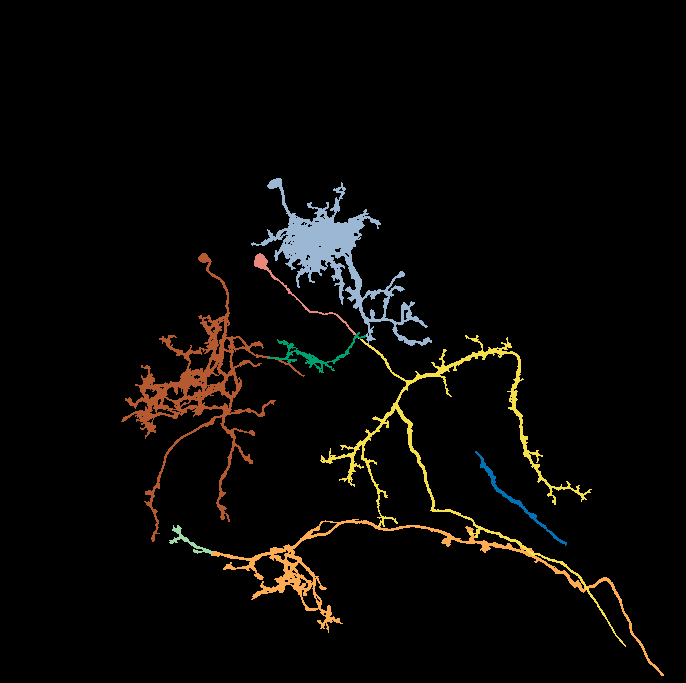}%
        \phantomsubcaption\label{fig:qual_res:ffn}%
    \end{subfigure}%
    \hfill%
    \begin{subfigure}[b]{0.19\textwidth}%
        \includegraphics[width=\textwidth]{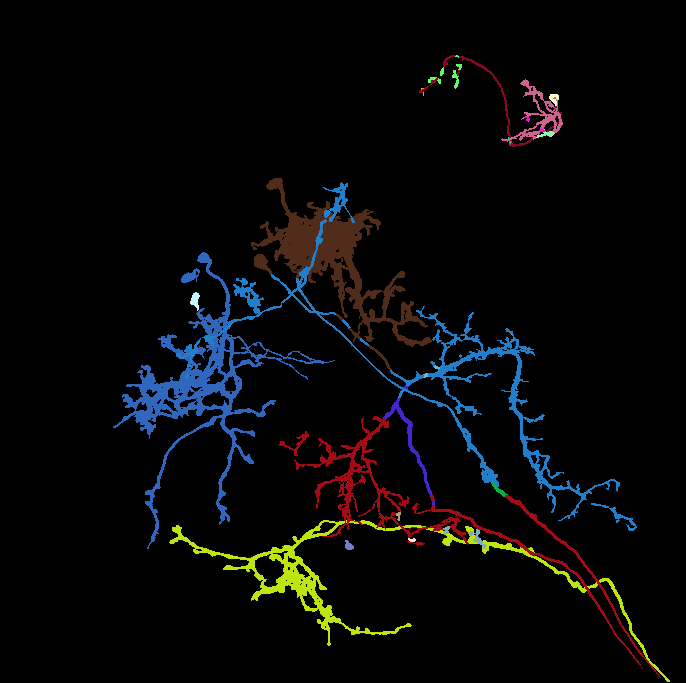}%
        \phantomsubcaption\label{fig:qual_res:duan}%
    \end{subfigure}%
    \hfill%
    \hfill%
    \end{subfigure}%
    
\vspace*{-21.0em}  
\textcolor{white}{\begin{flushleft}\raggedright \hspace{2.7em} (a) MCFO \hspace{6.0em} (b) gt \hspace{6.7em} (c) ppp \hspace{6.5em} (d) FFN \hspace{5.0em} (e) Duan et al.\end{flushleft}}
\vspace*{18em}

\vspace*{-14em}  
\textcolor{white}{\begin{flushleft}\raggedright\footnotesize\quad VT011145-20171222\_63\_I2\end{flushleft}}

\vspace*{5.7em}  
\textcolor{white}{\begin{flushleft}\raggedright\footnotesize\quad R14A02-20180905\_65\_A6\end{flushleft}}
\vspace*{0.3em}
    \caption{Qualitative results for our three baseline methods: PatchPerPix (ppp), Flood Filling Networks (FFN) and Duan et al.'s color clustering.
    In the top row all three methods yield few correctly segmented neurons (the two green neurons), but ppp and FFN merge the blue and the red one, and Duan et al.'s splits the blue neuron while nicely segmenting the red one.
    In the bottom row ppp merges many neurons of different color; FFN segments three neurons, but has low coverage; and Duan et al.'s also merges different colored neurons.}\label{fig:comparative_results}
\end{figure*}

To showcase the FISBe dataset together with our selection of metrics, we provide evaluation results for three baseline methods, namely PatchPerPix~\cite{ppp2020}, Flood Filling Network (FFN)~\cite{januszewski2016ffn, januszewski2018ffn} and a non-learnt application-specific color clustering from Duan et al.\ \cite{sumbul2026, duan2021}.
For information on all models, including training and validation details, see Suppl.\ Sec.~\ref{suppl:ext_baseline}.
Quantitative results for the full dataset are shown in Table~\ref{tab:results_both_sets}, for completely and partly sets in Suppl.\ Table~\ref{suppl_tab:results_both_sets} and \ref{suppl_tab:results_both_sets_f_scores}. 
Qualitative results are shown in Fig.~\ref{fig:comparative_results}, Suppl.\ Fig.~\ref{suppl_fig:comparative_results} and \ref{suppl_fig:vis_res_ppp_completely}.
The results show that all three baseline methods yield large fragments for clearly visible and easily separable neurons.
There are, however, many segmentation errors as reflected by the low avF1 scores (maximum value of 0.34\std{0.01}).
PatchPerPix achieves best results for all metrics, except for false merges.
Inspecting PatchPerPix results visually shows that many touching neurons are falsely merged even with different color.
FFNs and Duan et al.\ perform similar, although Duan et al.\ has the highest number of false splits and lowest number of false merges. Thus it separates best touching neurons.

As the training of PatchPerPix is not directly applicable to the partly labeled data we only report results for models trained on the completely labeled data.
FFNs, on the other hand, operate in a one-versus-all fashion and can thus by design train on partly labeled data without any modifications.
Training FFN on the full dataset shows increases in all metrics, especially on the test set.
PatchPerPix intrinsically handles overlapping instances
but it can only bridge small overlaps up to the used patch size.
FFN does not support overlaps out of the box but could, with some modifications for efficient inference, be extended to this end.
None of the learnt methods models long-range data dependencies.
In summary, all three baseline methods do yield some true positive neuron reconstructions, but extensive further method development is necessary to be able to achieve high quality instance segmentation on this dataset.

\section{Conclusion}\label{sec:conclusion}

With this work we release the FISBe dataset, which is, to the best of our knowledge, the first real-world benchmark dataset for instance segmentation of wide-ranging thin filamentous intertwined objects.
In addition to the data we contribute a set of metrics for meaningful method benchmarking and three baselines.
A limitation of FISBe is its bias towards sparser samples from the FlyLight MCFO image resource it stems from. This entails that  in general, benchmarking on FISBe does not serve to gauge method performance on denser samples. 
A possible avenue to mitigate this issue is to define proxy evaluation metrics on denser samples with the help of domain-specific downstream tasks for which higher-level annotations exist (see Sec.~\ref{subsec:eval_discussion} for examples of such tasks).
Main limitations of our baseline methods are that they handle no or only small overlaps and are computationally very demanding.
In future work, we are excited to see recently proposed methods for long-range data dependencies, such as structured state space models~\cite{gu2021statespace,nguyen2022s4nd} and continuous CNNs~\cite{knigge2023modelling}, applied to this real-world dataset.
In addition, the huge set of already available non-annotated images should lend itself perfectly for self-supervised pretraining.
We believe that the new challenging FISBe dataset is a great resource to the computer vision community as it might reveal blind spots of current methods. 
Thus, we hope that it will lead to new methods development for capturing long range data dependencies, while at the same time advancing cell-level analyses in basic neuroscience. 
\vspace{-1em}

\paragraph{Acknowledgements.}
We thank Aljoscha Nern for providing unpublished MCFO images as well as Geoffrey W. Meissner and the entire FlyLight Project Team for valuable discussions. 
P.H., L.M. and D.K. were supported by the HHMI Janelia Visiting Scientist Program.

{
    \small
    \bibliographystyle{ieeenat_fullname}
    \bibliography{main}
}

\clearpage
\graphicspath{{./supplement/}}
\maketitlesupplementary
\setcounter{secnumdepth}{4}

\appendix
\section{Appendix}\label{sec:appendix}
\subsection{Dataset Documentation}\label{suppl:datset_doc}

\subsubsection{Datasheet}\label{suppl:datasheet}

In this section we answer the Datasheet for Datasets questionnaire \cite{datasheet_gebru2018} to document FISBe, the FlyLight Instance Segmentation Benchmark dataset. It contains information about motivation, composition, collection, preprocessing, usage, licensing as well as hosting and maintenance plan.

\definecolor{darkblue}{RGB}{0,0,0}

\newcommand{\dssectionheader}[1]{%
   \noindent\framebox[\columnwidth]{%
      {\textbf{\textcolor{darkblue}{#1}}}
   }
}

\newcommand{\dsquestion}[1]{%
    {\noindent \textcolor{darkblue}{\textbf{#1}}}
}

\newcommand{\dsquestionex}[2]{%
    {\noindent \textcolor{darkblue}{\textbf{#1} #2 \smallskip}}
}

\newcommand{\dsanswer}[1]{%
   {\noindent #1 \medskip}
}

\subsubsubsection{Motivation}

\dsquestionex{For what purpose was the dataset created?}{Was there a specific task in mind? Was there a specific gap that needed to be filled? Please provide a description.}

\dsanswer{Segmenting individual neurons in multi-neuron light microscopy (LM) recordings is intricate due to the long, thin filamentous and widely branching morphology of individual neurons, the tight interweaving of multiple neurons, and LM-specific imaging characteristics like partial volume effects and uneven illumination.
These properties reflect a current key challenge for deep-learning models across domains, namely to efficiently capture long-range dependencies in the data. While methodological research on this topic is buzzing in the machine learning community, to date, respective methods are typically benchmarked on synthetic datasets.
To fill this gap, we created the FlyLight Instance Segmentation Benchmark dataset, to the best of our knowledge, the first publicly available multi-neuron LM dataset with pixel-wise ground truth and the first real-world benchmark dataset for instance segmentation of long thin filamentous objects.
}

\dsquestion{Who created this dataset (e.g., which team, research group) and on behalf of which entity (e.g., company, institution, organization)?}

\iftoggle{cvprfinal}{
\dsanswer{This dataset was created in a collaboration of the Max-Delbrueck-Center for Molecular Medicine in the Helmholtz Association (MDC) and the Howard Hughes Medical Institute Janelia Research Campus. More precisely, the Kainmueller lab at the MDC and the Project Technical Resources Team at Janelia.
}}{
\dsanswer{\textit{Hidden during review process}}}

\dsquestionex{Who funded the creation of the dataset?}{If there is an associated grant, please provide the name of the grantor and the grant name and number.}

\iftoggle{cvprfinal}{
\dsanswer{Howard Hughes Medical Institute Janelia Research Campus and Max-Delbrueck-Center for Molecular Medicine in the Helmholtz Association (MDC) funded the creation of the dataset.
}
}{
\dsanswer{\textit{Hidden during review process}}}

\subsubsubsection{Composition}

\dsquestionex{What do the samples\footnote{We changed \textit{instances} to \textit{samples} when refering to images of the dataset to not use the term ambiguously; instead we only use \textit{instances} to refer to \textit{object instances} in images.} that comprise the dataset represent (e.g., documents, photos, people, countries)?}{ Are there multiple types of samples (e.g., movies, users, and ratings; people and interactions between them; nodes and edges)? Please provide a description.}

\dsanswer{The dataset consists of 3d multi-neuron multicolor light microscopy images and their respective pixel-wise instance segmentation masks.
The "raw" light microscopy data shows neurons of the fruit fly Drosophila Melanogaster acquired with a technique called MultiColor FlpOut (MCFO) \cite{mcfo_Nern2015,meissner2023}.
Fruit fly brains of different transgenic lines (e.g. GAL4 lines \cite{gal4_jenett2012}) were imaged, each transgenic line tags a different set of neurons.
There are multiple MCFO images of the same transgenic line, where each MCFO image expresses (shows) a stochastic subset of the tagged neurons.
The neurons contained in each image were manually annotated by trained expert annotators.
The dataset is split into a \textit{completely} labeled (all neurons in the image are manually segmented) and a \textit{partly} labeled (a subset of neurons in the image is manually segmented) set.
}

\dsquestion{How many samples/instances are there in total (of each type, if appropriate)?}

\dsanswer{The \textit{completely} labeled set comprises 30 images with 139 labeled neurons in total, and the \textit{partly} labeled set comprises 71 images with 451 labeled neurons in total.
}

\dsquestionex{Does the dataset contain all possible samples or is it a subset (not necessarily random) of samples from a larger set?}{ If the dataset is a sample, then what is the larger set? Is the sample representative of the larger set (e.g., geographic coverage)? If so, please describe how this representativeness was validated/verified. If it is not representative of the larger set, please describe why not (e.g., to cover a more diverse range of instances, because instances were withheld or unavailable).}

\dsanswer{The dataset contains a subset of 101 images from the "40x Gen1" set of \cite{meissner2023}.
The full "40x Gen1" set consists of 46,791 images of 4575 different transgenic lines. 
From this set, we selected relatively sparse images in terms of number of expressed neurons which seemed feasible for manual annotation.
Thus, our dataset is not representative for the full "40x Gen1" MCFO collection.}

\dsquestionex{What data does each sample consist of? “Raw” data (e.g., unprocessed text or images) or features? Is there a label or target associated with each sample?}{Please provide a description.}

\dsanswer{Each sample consists of a single 3d MCFO image of neurons of the fruit fly.
For each image, we provide a pixel-wise instance segmentation for all separable neurons.
Each sample is stored as a separate \textit{zarr} file ("Zarr is a file storage format for chunked, compressed, N-dimensional arrays based on an open-source specification." \href{https://zarr.readthedocs.io}{https://zarr.readthedocs.io}).
The image data ("raw") and the segmentation ("gt\_instances") are stored as two arrays within a single zarr file.
The segmentation mask for each neuron is stored in a separate channel.
The order of dimensions is CZYX.
In Python the data can, for instance, be opened with:}
\begin{minipage}{\linewidth}
\begin{lstlisting}[language=python]
import zarr
raw = zarr.open(
  <path_to_zarr>, 
  path="volumes/raw")
seg = zarr.open(
  <path_to_zarr>, 
  path="volumes/gt_instances")
\end{lstlisting}
\end{minipage}
\dsanswer{Zarr arrays are read lazily on-demand.
Many functions that expect numpy arrays also work with zarr arrays.
The arrays can also explicitly be converted to numpy arrays with:}
\begin{minipage}{\linewidth}
\begin{lstlisting}[language=python]
import numpy as np
raw_np = np.array(raw)
\end{lstlisting}
\end{minipage}

\dsquestionex{Is any information missing from individual samples?}{If so, please provide a description, explaining why this information is missing (e.g., because it was unavailable). This does not include intentionally removed information, but might include, e.g., redacted text.}

\dsanswer{Not all neuronal structures could be segmented within all images of the provided dataset. 
Mainly, there are two reasons: (1) there are overlapping neurons with the same or a similar color that could not be separated due to the partial volume effect, and (2) some neuronal structures cannot be delineated correctly in the presence of noisy background in the same color as the neuron itself.
In the \textit{completely} labeled set all neuronal structures have been segmented, in the \textit{partly} labeled set some structures are missing.
}

\dsquestionex{Are relationships between individual samples made explicit (e.g., users’ movie ratings, social network links)?}{If so, please describe how these relationships are made explicit.}

\dsanswer{Yes, one transgenic line is often imaged multiple times as only a stochastic subset of all tagged neurons is visible per MCFO image.
Moreover, the same neuron might be tagged in multiple transgenic lines.}

\dsquestionex{Are there recommended data splits (e.g., training, development/validation, testing)?}{If so, please provide a description of these splits, explaining the rationale behind them.}

\dsanswer{Yes, we provide a recommended data split for training, validation and testing.
The files in the provided download are presorted according to this recommendation.
When splitting the data into sets, we made sure that images of the same transgenic lines are in the same split and paid attention to having similar proportions of images with overlapping neurons as well as having a similar average number of neurons per image in each split.
}

\dsquestionex{Are there any errors, sources of noise, or redundancies in the dataset?}{If so, please provide a description.}

\dsanswer{There might be uneven illumination resulting in gaps within neurons in the raw microscopy images as well as the corresponding annotations. This is intrinsic to this kind of light microscopy images.
}

\dsquestionex{Is the dataset self-contained, or does it link to or otherwise rely on external resources (e.g., websites, tweets, other datasets)?}{If it links to or relies on external resources, a) are there guarantees that they will exist, and remain constant, over time; b) are there official archival versions of the complete dataset (i.e., including the external resources as they existed at the time the dataset was created); c) are there any restrictions (e.g., licenses, fees) associated with any of the external resources that might apply to a future user? Please provide descriptions of all external resources and any restrictions associated with them, as well as links or other access points, as appropriate.}

\dsanswer{Yes, the dataset is self-contained.
There is an external, additional source of raw images that could potentially be used for self-supervised learning.
The raw images in our dataset are a subset of the released MCFO collection of the FlyLight project \cite{meissner2023}. The whole collection can be downloaded at \href{https://gen1mcfo.janelia.org}{https://gen1mcfo.janelia.org}.
Note though that there are no segmentation masks available for these images.
}

\dsquestionex{Does the dataset contain data that might be considered confidential (e.g., data that is protected by legal privilege or by doctor-patient confidentiality, data that includes the content of individuals non-public communications)?}{If so, please provide a description.}

\dsanswer{No.}

\subsubsubsection{Collection Process}

\dsquestionex{How was the data associated with each sample acquired?}{Was the data directly observable (e.g., raw text, movie ratings), reported by subjects (e.g., survey responses), or indirectly inferred/derived from other data (e.g., part-of-speech tags, model-based guesses for age or language)? If data was reported by subjects or indirectly inferred/derived from other data, was the data validated/verified? If so, please describe how.}

\dsanswer{The content of the raw images was directly recorded using confocal microscopes.
The annotations were created manually.
}

\dsquestionex{What mechanisms or procedures were used to collect the data (e.g., hardware apparatus or sensor, manual human curation, software program, software API)?}{How were these mechanisms or procedures validated?}

\dsanswer{Imaging was performed using eight Zeiss LSM 710 or 780 laser scanning confocal microscopes (for more information on the imaging process see~\cite{meissner2023}).
Two trained expert annotators manually segmented and proof-read each other to segment the neurons in these images using the interactive rendering tool VVD Viewer (\href{https://github.com/JaneliaSciComp/VVDViewer}{https://github.com/JaneliaSciComp/VVDViewer}).
}

\dsquestion{If the dataset is a sample from a larger set, what was the sampling strategy (e.g., deterministic, probabilistic with specific sampling probabilities)?}

\dsanswer{We manually selected images from the larger "40x Gen1" collection. 
We chose images that contained a sparse set of neurons and that contained neurons that preferably were not contained in previously selected images.
}

\dsquestion{Who was involved in the data collection process (e.g., students, crowdworkers, contractors) and how were they compensated (e.g., how much were crowdworkers paid)?}

\dsanswer{The data collection process was done by full time employees at \iftoggle{cvprfinal}{the Howard Hughes Medical Institute Janelia Research Campus and the Max-Delbrueck-Center for Molecular Medicine in the Helmholtz Association (MDC)}{\textit{Hidden during review process}}.}

\dsquestionex{Over what timeframe was the data collected? Does this timeframe match the creation timeframe of the data associated with the samples (e.g., recent crawl of old news articles)?}{If not, please describe the timeframe in which the data associated with the samples was created.}

\dsanswer{MCFO selection and manual annotation were mainly done in 2018 and 2019. 
The respective acquisition date of the MCFO sample is noted within the sample name in "YYYYMMDD" format. Most samples of our dataset were acquired in 2017 and 2018.
}

\dsquestionex{Were any ethical review processes conducted (e.g., by an institutional review board)?}{If so, please provide a description of these review processes, including the outcomes, as well as a link or other access point to any supporting documentation.}

\dsanswer{There was no ethical review process conducted as we did not record any new animal data, the dataset does not relate to people and it does not contain confidential data.
}

\dsquestionex{Does the dataset relate to people?}{If not, you may skip the remaining questions in this section.}

\dsanswer{No.
}

\subsubsubsection{Preprocessing/cleaning/labeling}

\dsquestionex{Was any preprocessing/cleaning/labeling of the data done (e.g., discretization or bucketing, tokenization, part-of-speech tagging, SIFT feature extraction, removal of instances, processing of missing values)?}{If so, please provide a description. If not, you may skip the remainder of the questions in this section.}

\dsanswer{The following preprocessing was done for each image:
The central brain and part of the ventral nerve cord (VNC) were recorded in tiles by the light microscope.
The tiles were stitched together and distortion corrected (for more information see \cite{stitching_yu2011}).}

\dsquestionex{Was the “raw” data saved in addition to the preprocessed/cleaned/labeled data (e.g., to support unanticipated future uses)?}{If so, please provide a link or other access point to the “raw” data.}

\dsanswer{The original images are available at \href{https://gen1mcfo.janelia.org}{https://gen1mcfo.janelia.org}.}

\dsquestionex{Is the software used to preprocess/clean/label the samples available?}{If so, please provide a link or other access point.}

\dsanswer{The image processing, such as distortion correction and stitching, is done by using the open-source software Janelia Workstation~\cite{workstation_Rokicki2019}.
}

\subsubsubsection{Uses}

\dsquestionex{Has the dataset been used for any tasks already?}{If so, please provide a description.}

\iftoggle{cvprfinal}{
\dsanswer{In \cite{ppp2020}, an earlier, unpublished version of our dataset has been used to qualitatively evaluate \mbox{PatchPerPix}, a deep learning-based instance segmentation method.
The trained model was then applied to $\sim$40.000 samples of the MCFO collection~\cite{mais2021patchperpixmatch, meissner2023} to search for given neuronal structures extracted from electron microscopy (EM) data~\cite{scheffer2020connectome}.
\mbox{PatchPerPix} is also used as one of three baselines to showcase this published version of our dataset.
}
}{
\dsanswer{\textit{Hidden during review process}}}

\dsquestionex{Is there a repository that links to any or all papers or systems that use the dataset?}{If so, please provide a link or other access point.}

\dsanswer{As they are getting published, we will reference them at \iftoggle{cvprfinal}{\href{https://kainmueller-lab.github.io/fisbe}{https://kainmueller-lab.github.io/fisbe}}{\textit{Hidden during review process}}}

\dsquestion{What (other) tasks could the dataset be used for?}

\dsanswer{The dataset can be used for a wide range of method development tasks such as capturing long-range dependencies, segmentation of thin filamentous structures, self- and semi-supervised training or denoising.
Advances in these areas can in turn facilitate scientific discoveries in basic neuroscience by providing improved neuron reconstructions for morphological and functional analyses.
}

\subsubsubsection{Distribution}

\dsquestionex{Will the dataset be distributed to third parties outside of the entity (e.g., company, institution, organization) on behalf of which the dataset was created?}{If so, please provide a description.}

\dsanswer{The dataset will be publicly available.}

\dsquestionex{How will the dataset be distributed (e.g., tarball on website, API, GitHub)}{Does the dataset have a digital object identifier (DOI)?}

\dsanswer{The dataset will be distributed through \href{https://zenodo.org}{zenodo} (DOI: \href{https://zenodo.org/doi/10.5281/zenodo.10875063}{10.5281/zenodo.10875063}) and our project page \href{https://kainmueller-lab.github.io/fisbe}{https://kainmueller-lab.github.io/fisbe}.}

\dsquestion{When will the dataset be distributed?}

\dsanswer{With publication of the accompanying paper.}

\dsquestionex{Will the dataset be distributed under a copyright or other intellectual property (IP) license, and/or under applicable terms of use (ToU)?}{If so, please describe this license and/or ToU, and provide a link or other access point to, or otherwise reproduce, any relevant licensing terms or ToU, as well as any fees associated with these restrictions.}

\dsanswer{The dataset will be distributed under the Creative Commons Attribution 4.0 International (CC BY 4.0) license (\href{https://creativecommons.org/licenses/by/4.0/}{https://creativecommons.org/licenses/by/4.0/}).}

\dsquestionex{Have any third parties imposed IP-based or other restrictions on the data associated with the samples?}{If so, please describe these restrictions, and provide a link or other access point to, or otherwise reproduce, any relevant licensing terms, as well as any fees associated with these restrictions.}

\dsanswer{All MCFO images have previously been made publicly available by~\cite{meissner2023} under the same license (CC BY 4.0) at \href{https://gen1mcfo.janelia.org}{https://gen1mcfo.janelia.org}.}

\subsubsubsection{Maintenance}

\dsquestion{Who will be supporting/hosting/maintaining the dataset?}

\dsanswer{\iftoggle{cvprfinal}{Lisa Mais}{\textit{Hidden during review process}} supports and maintains the dataset.
}

\dsquestion{How can the owner/curator/manager of the dataset be contacted (e.g., email address)?}

\dsanswer{\iftoggle{cvprfinal}{Lisa Mais and Dagmar Kainmueller}{\textit{Hidden during review process}} can be contacted at \iftoggle{cvprfinal}{\{firstname.lastname\}@mdc-berlin.de}{\textit{Hidden during review process}}.}

\dsquestionex{Is there an erratum?}{If so, please provide a link or other access point.}

\dsanswer{Errata will be published at \iftoggle{cvprfinal}{\href{https://kainmueller-lab.github.io/fisbe}{https://kainmueller-lab.github.io/fisbe}}{\textit{Hidden during review process}}.}

\dsquestionex{Will the dataset be updated (e.g., to correct labeling errors, add new samples, delete samples)?}{If so, please describe how often, by whom, and how updates will be communicated to users (e.g., mailing list, GitHub)?}

\dsanswer{The dataset will be updated to correct erroneous segmentation and potentially to add new samples and annotations.
It will be updated when a relevant number of updates has accumulated.
Updates will be communicated through \iftoggle{cvprfinal}{\href{https://kainmueller-lab.github.io/fisbe}{https://kainmueller-lab.github.io/fisbe}}{\textit{Hidden during review process}}.}

\dsquestionex{Will older versions of the dataset continue to be supported/hosted/maintained?}{If so, please describe how. If not, please describe how its obsolescence will be communicated to users.}

\dsanswer{We publish our dataset on zenodo.
Zenodo supports versioning, including DOI versioning.
Older versions of the dataset will thus stay available.}

\dsquestionex{If others want to extend/augment/build on/contribute to the dataset, is there a mechanism for them to do so?}{If so, please provide a description. Will these contributions be validated/verified? If so, please describe how. If not, why not? Is there a process for communicating/distributing these contributions to other users? If so, please provide a description.}

\dsanswer{We welcome contributions to our dataset.
Errata, new samples and annotations and other contributions can be contributed via \textit{github issues} at \iftoggle{cvprfinal}{\href{https://kainmueller-lab.github.io/fisbe}{https://kainmueller-lab.github.io/fisbe}}{\textit{Hidden during review process}}.
We will verify such contributions and update the dataset accordingly.}
\subsubsection{How to Open and View Image Files}\label{suppl:howto}

We recommend viewing the FISBe dataset with napari~\cite{napari}. 
The following instructions have been tested with Linux.
While they should also work for Windows and MacOS, they might require some small changes. 
Please follow the official installation instructions (\href{https://napari.org/stable/}{https://napari.org/stable/}):
\begin{lstlisting}[language=bash]
  conda create -y -n napari-env -c \
    conda-forge python=3.9
  conda activate napari-env
  pip install "napari[all]" zarr
\end{lstlisting}
Then save the following Python script (also included in the provided download of our dataset):
\begin{lstlisting}[language=python]
import zarr, sys, napari

raw = zarr.load(
    sys.argv[1], path="volumes/raw")
gt = zarr.load(
    sys.argv[1], path="volumes/gt_instances")

viewer = napari.Viewer(ndisplay=3)
for idx, gt in enumerate(gts):
    viewer.add_labels(
        gt, rendering='translucent', 
        blending='additive', name=f'gt_{idx}')
viewer.add_image(raw[0], colormap="red", 
    name='raw_r', blending='additive')
viewer.add_image(raw[1], colormap="green",  
    name='raw_g', blending='additive')
viewer.add_image(raw[2], colormap="blue", 
    name='raw_b', blending='additive')
napari.run()
\end{lstlisting}
Execute it from the command line to view the image:
\begin{lstlisting}[language=bash]
  python <script_name.py> <path-to-file>/R9F03-20181030_62_B5.zarr
\end{lstlisting}
\subsection{Extended Metrics Information}\label{suppl:ext_metrics}

\begin{figure}
    \centering
		\includegraphics[width=\linewidth]{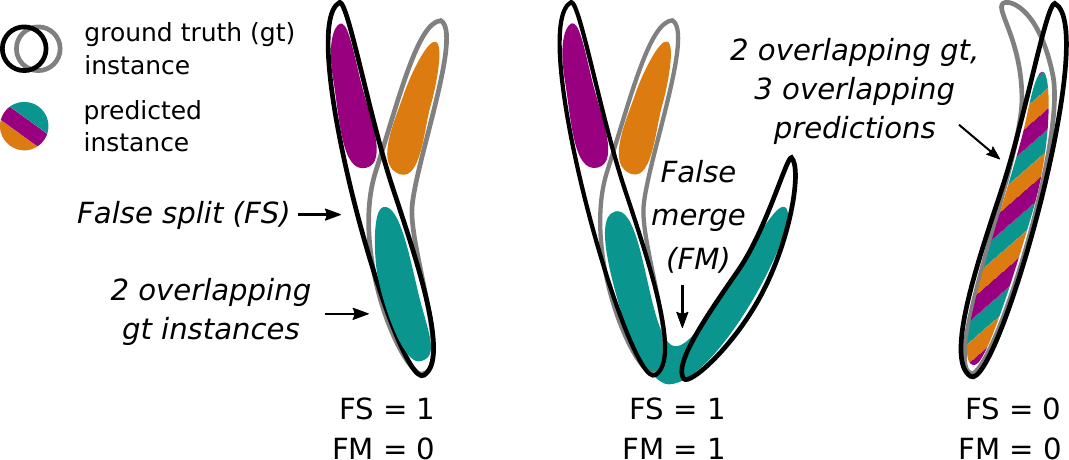}
        \begin{tabular}{m{0.19\columnwidth}m{0.21\columnwidth}m{0.23\columnwidth}m{0.3\columnwidth}}
			\begin{subfigure}[b]{0.19\columnwidth}
			\end{subfigure}&\hfill
            \begin{subfigure}[b]{0.21\columnwidth}
				\caption{}
			\end{subfigure}&\hfill
            \begin{subfigure}[b]{0.23\columnwidth}
				\caption{}
			\end{subfigure}&\hfill
			\begin{subfigure}[b]{0.3\columnwidth}
				\caption{}
			\end{subfigure}
   \end{tabular}
	\caption{Exemplary challenges for many-to-many matching with overlaps.
 (a) One predicted instance lies completely within an overlapping gt region, but it should only be assigned to one of them;
 (b) one predicted instance covers one gt and merges with an overlapping gt region, here it should be assigned to the single gt and one of the overlapping ones;
 and (c) three overlapping predicted instances cover two overlapping gt instances, here only two predicted instances should be matched to the two gt instances respectively (the other predicted instance should rather only count as false positive than as false split).
 As there are plenty of scenarios how gt and predicted instances can overlap,
 special treatment for overlapping regions is difficult and error-prone.
 However, our proposed algorithm 
 (see Alg. 1 in the main paper)
 naturally handles such overlaps by keeping track of already matched pixels (as opposed to only on the level of instances).
 }
	\label{fig:testcases_FM_FS}
\end{figure}

\begin{figure*}[t]
	\hfill%
    \begin{subfigure}[b]{0.1\textwidth}%
		\centering%
		\includegraphics[width=\textwidth]{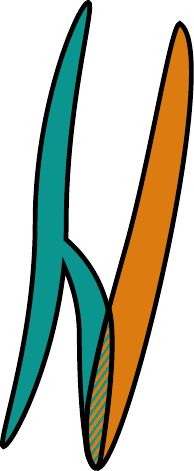}%
        \caption{\label{fig:edge_case_perfect}}%
	\end{subfigure}%
 	\hfill%
    \begin{subfigure}[b]{0.1\textwidth}%
		\centering%
		\includegraphics[width=\textwidth]{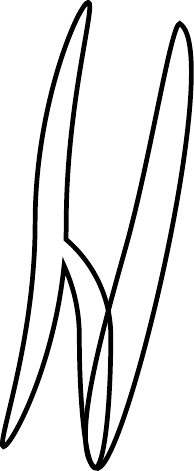}%
        \caption{\label{fig:edge_case_no_pred}}%
	\end{subfigure}%
    \hfill%
    \begin{subfigure}[b]{0.1\textwidth}%
		\centering%
		\includegraphics[width=\textwidth]{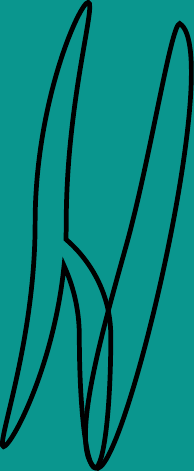}%
        \caption{\label{fig:edge_case_one_inst_per_img}}%
	\end{subfigure}%
     \hfill%
    \begin{subfigure}[b]{0.1\textwidth}%
		\centering%
		\includegraphics[width=\textwidth]{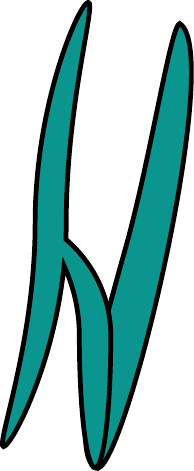}%
        \caption{\label{fig:edge_case_one_inst_per_cc}}%
	\end{subfigure}%
    \hfill%
     \begin{subfigure}[b]{0.1\textwidth}%
		\centering%
		\includegraphics[width=\textwidth]{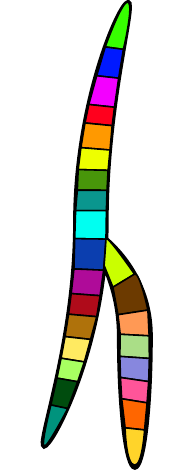}%
        \caption{\label{fig:edge_case_one_inst_per_pix}}%
	\end{subfigure}%
     \hfill%
     \begin{subfigure}[b]{0.1\textwidth}%
		\centering%
		\includegraphics[width=\textwidth]{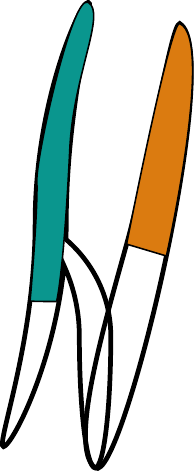}%
        \caption{\label{fig:edge_case_half_seg_two_inst}}%
	\end{subfigure}%
     \hfill%
     \begin{subfigure}[b]{0.1\textwidth}%
		\centering%
		\includegraphics[width=\textwidth]{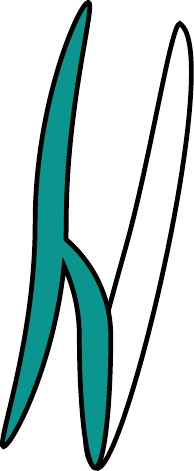}%
        \caption{\label{fig:edge_case_half_seg_one_full}}%
	\end{subfigure}%
    \hfill%
    \begin{subfigure}[b]{0.1\textwidth}%
		\centering%
		\includegraphics[width=\textwidth]{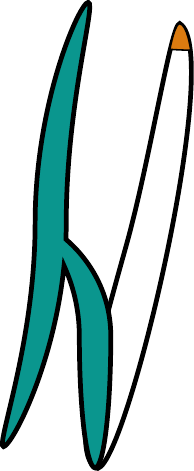}%
        \caption{\label{fig:edge_case_half_seg_one_full_one_tiny}}%
	\end{subfigure}%
    \hfill%
    \begin{subfigure}[b]{0.1\textwidth}%
		\centering%
		\includegraphics[width=\textwidth]{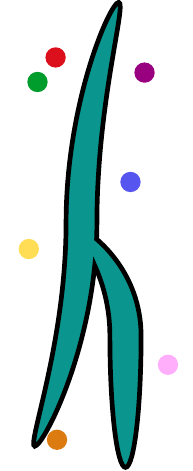}%
        \caption{\label{fig:edge_case_perfect_plus_fps}}%
	\end{subfigure}%
    \hfill%
    \hfill%

    \SetTblrInner{rowsep=1pt}
    \begin{tblr}{width=\linewidth,colspec={
    X[1.0,l] 
    X[1.0,si={table-format=1.2,table-number-alignment=center},c]
    J[1.2] J[1.2] J[1.2]
    Q[co=1,1.0,si={table-format=2,table-number-alignment=center},c] 
    J[2] J[2] J[2] J[2] J[2]
    },row{1}={guard}}
        \toprule
    	   &      \(S\) & \(avF1\) & \(C\) & \(\text{clDice}_{\text{TP}}\) & \#Pred & \(TP\) & \(FP\) & \(FS\) & \(FN\) & \(FM\)\\
    	\midrule
        (a) & 1.0  & 1.0  &      1.0 &      1.0 & 2      &  2 &  0 &  0 &  0 &  0\\
        (b) & 0.0  & 0.0  &      0.0 &      0.0 & 0      &  0 &  0 &  0 &  2 &  0\\
        (c) & 0.0  & 0.0  &      0.0 &      0.0 & 1      &  0 &  1 &  0 &  2 &  1\\
        (d) & 0.47 & 0.44 &      0.5 &      0.67& 1      &  1 &  0 &  0 &  1 &  1\\
        (e) & 0.5  & 0.0  &      1.0 &      0.0 & 31     &  0 & 31 & 30 &  2 &  0\\
    	  (f) & 0.58 & 0.67 &      0.51 &     0.68& 2      &  2 &  0 &  0 &  0 &  0\\
    	  (g) & 0.58 & 0.67 &      0.5 &      1.0 & 1      &  1 &  0 &  0 &  1 &  0\\
    	  (h) & 0.5  & 0.5  &      0.52&      1.0 & 2      &  1 &  1 &  0 &  1 &  0\\
        (i) & 0.68 & 0.36 &      1.0 &      1.0 & 9      &  2 &  7 &  0 &  0 &  0\\
    	\bottomrule
 	\end{tblr}

    \begin{tblr}{width=\linewidth,colspec={X[1.0,l] J[1.2] J[1.2] J[1.2] J[1.2] J[1.2] J[1.2] J[1.2] J[1.2] J[1.2]},row{1}={guard}}
        \midrule
		& F1$_{0.1}$ &F1$_{0.2}$ &F1$_{0.3}$ &F1$_{0.4}$ &F1$_{0.5}$ &F1$_{0.6}$ &F1$_{0.7}$ &F1$_{0.8}$ &F1$_{0.9}$\\
		\midrule
		(d)&    0.67 &      0.67 &      0.67 &      0.67 &      0.67 &      0.67 &      0.0  &      0.0  &      0.0 \\
		(f)&    1.0  &      1.0  &      1.0  &      1.0  &      1.0  &      1.0  &      0.0  &      0.0  &      0.0 \\
		(g)&    0.67 &      0.67 &      0.67 &      0.67 &      0.67 &      0.67 &      0.67 &      0.67 &      0.67\\
		(h)&    0.5  &      0.5  &      0.5  &      0.5  &      0.5  &      0.5  &      0.5  &      0.5  &      0.5 \\
        (i)&    0.36  &      0.36  &      0.36  &      0.36  &      0.36  &      0.36  &      0.36  &      0.36  &      0.36 \\
        \bottomrule
	\end{tblr}
 
	\caption{Quantitative assessment of a number of different edge cases of our evaluation metrics (outline: ground truth, color: predictions, $\text{th}=0.5$), highlighting their applicability and validity for FISBe.\label{fig:edge_cases}
    In (a) we have a perfect prediction, the score is perfect and there are no errors.
    In (b) we have no prediction, the score is zero and we have as many FN as there are instances.
    In (c) we have one prediction that covers the whole image; as the clDice value is too low, there is no TP, so avF1 is still zero.
    When computing the clPrecision for the predicted instance, the corresponding skeleton will likely have the largest overlap with the ground truth background and will be matched to it. Thus, C will be zero as well.
    In (d) we have a perfect foreground segmentation but the two ground truth instances are merged; the predicted instance is assigned to one of the ground truth instances, resulting in $\text{C} = 0.5$. Assuming clDice $=0.67$ for the one match (and thus $\text{clDice}_{\text{TP}} = 0.67$), we have $\text{F1}=0.67$ for $\text{th}<0.7$ and $0$ otherwise.
    In (e) we again have a perfect foreground segmentation but there are many small instances; clDice for each pair of predicted and ground truth instances is $<0.1$, thus $\text{avF1}=0$ and $\text{clDice}_{\text{TP}}=0$; however, $\text{C} = 1.0$ because both instances are completely covered (and multiple predicted instances can be matched to one ground truth instance).
    In (f), (g) and (h) overall slightly more than half of the total ground truth is covered; in (f) both instances are covered slightly more than half; in (g) one instance is covered completely and the other is not; in (h) one instance is covered completely and only a tiny part of the other; to distinguish the cases quantitatively, one has to look at the details: about the same amount of ground truth is covered, thus C has a similar value; in (f) $\text{clDice}_{\text{TP}}$ is worst as both predicted instances are counted as TP, yet both only cover just over half of their respective ground truth instance; furthermore, while avF1 is identical for (f) and (g),
    when looking at the full range of F1$_{\text{th}}$ values there are more differences: in (f) there are 2 TP for $\text{th}<0.7$, resulting in F1 being equal to 1 for smaller thresholds and equal to 0 for larger thresholds; in (g) there is 1 TP for the full range of thresholds, but also 1 FN; in both cases this results in $\text{avF1}=0.67$;
    finally in (h) there is 1 TP for the full range of thresholds, 1 FN as in (g), but also 1 FP, resulting in $\text{avF1}=0.5$.
    In (i) we have a perfect prediction as in (a), but in addition we have a number of small FP, due to noise categorized as foreground; the coverage values are not affected, but the avF1 value drops.
    One could argue that (h) should be better than (g) as more is detected; however, if a prediction is too small, it is, in general, more likely to be noise.
    One could also argue that (d) should better than (g), as both neurons are detected, just merged; however, for downstream tasks having one fully correct instance that can directly be used is often more valuable than first having to manually fix errors.
    }
\end{figure*}

Table~\ref{tab:metric_overview} summarizes all used metrics with their localization criterion and matching.
Fig.~\ref{fig:testcases_FM_FS} highlights some of the challenges of computing a consistent many-to-many matching for overlapping instances.
Fig.~\ref{fig:edge_cases} visualizes and quantifies a comprehensive set of different edge cases of our evaluation metrics.

\newcommand{\pluseq}{\mathrel{+}=}

\begin{table}[htbp]
\caption{Overview of localization criterion and matching algorithm for used scores. Last line shows cardinalities of the ground truth-to-prediction relationships.}
\centering
\begin{tblr}{width=1.0\linewidth,rows={abovesep=1pt,belowsep=1pt},colspec={X[1.3,l] | X[1.1,c] X[1.4,c] X[1.4,c] X[1.4,c] X[1.6,c]}}
    Score & avF1 & C & FS & FM & clDice$_{TP}$ \\
    \midrule
    Loc. & clDice & clPrec. & clRecall & clRecall & clDice \\
    \SetCell[r=2]{}Match. & greedy & greedy & greedy & greedy & greedy\\
    & 1:1 & 1:n & n:m & n:m & 1:1 \\
\end{tblr}
\label{tab:metric_overview}
\end{table}

\subsection{Extended Baseline Information and Results}\label{suppl:ext_baseline}

\begin{figure*}[htbp]
	\centering%
    \begin{subfigure}[b]{\textwidth}%
    \caption*{1) VT058571-20170926\_64\_G6}%
    \hfill%
    \begin{subfigure}[b]{0.19\textwidth}%
        \includegraphics[width=\textwidth]{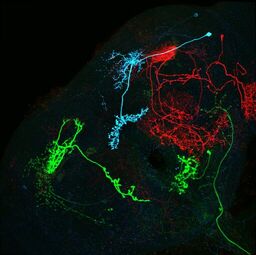}%
    \end{subfigure}%
    \hfill%
    \begin{subfigure}[b]{0.19\textwidth}%
        \includegraphics[width=\textwidth]{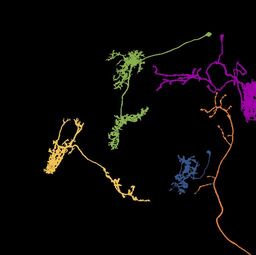}%
    \end{subfigure}%
    \hfill%
    \begin{subfigure}[b]{0.19\textwidth}%
        \includegraphics[width=\textwidth]{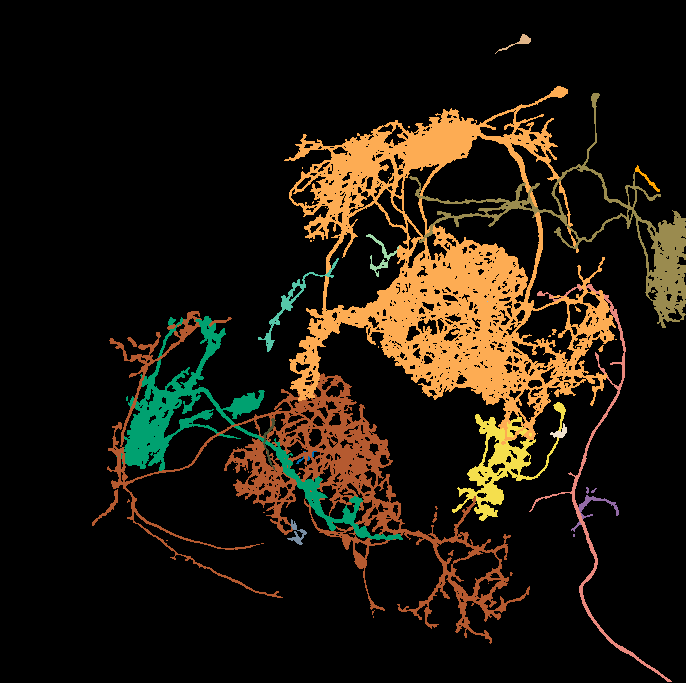}%
    \end{subfigure}%
    \hfill%
    \begin{subfigure}[b]{0.19\textwidth}%
        \includegraphics[width=\textwidth]{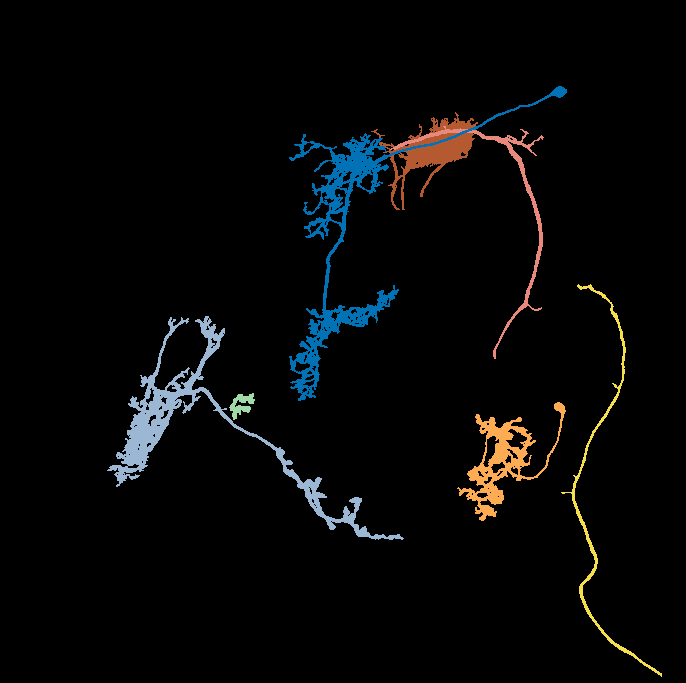}%
    \end{subfigure}%
    \hfill%
    \begin{subfigure}[b]{0.19\textwidth}%
        \includegraphics[width=\textwidth]{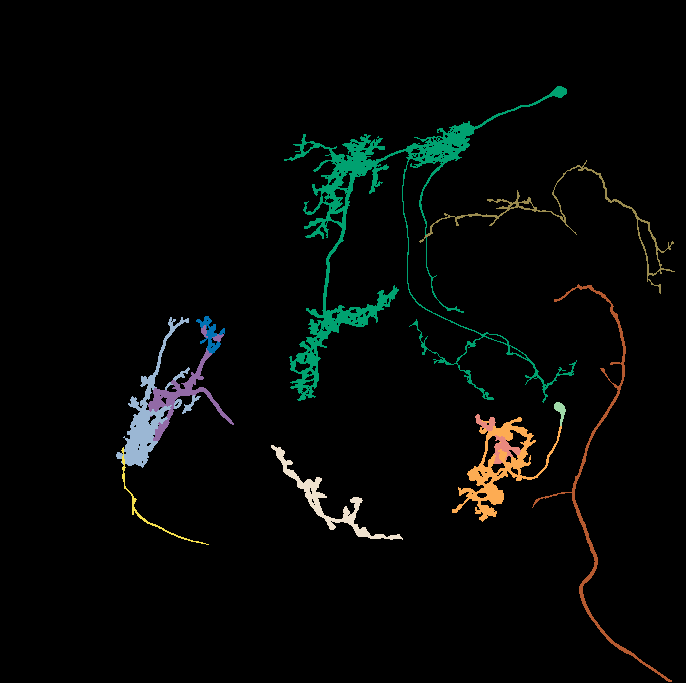}
    \end{subfigure}%
    \hfill%
    \hfill%
    \end{subfigure}%
    
    \begin{subfigure}[b]{\textwidth}%
    \caption*{2) R73H08-20181030\_62\_G5}%
    \hfill%
    \begin{subfigure}[b]{0.19\textwidth}%
        \includegraphics[width=\textwidth]{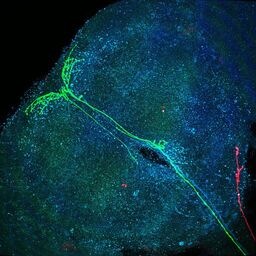}%
    \end{subfigure}%
    \hfill%
    \begin{subfigure}[b]{0.19\textwidth}%
        \includegraphics[width=\textwidth]{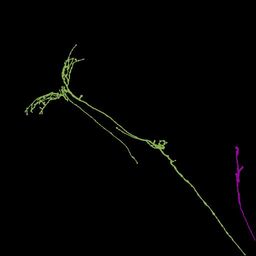}%
    \end{subfigure}%
    \hfill%
    \begin{subfigure}[b]{0.19\textwidth}%
        \includegraphics[width=\textwidth]{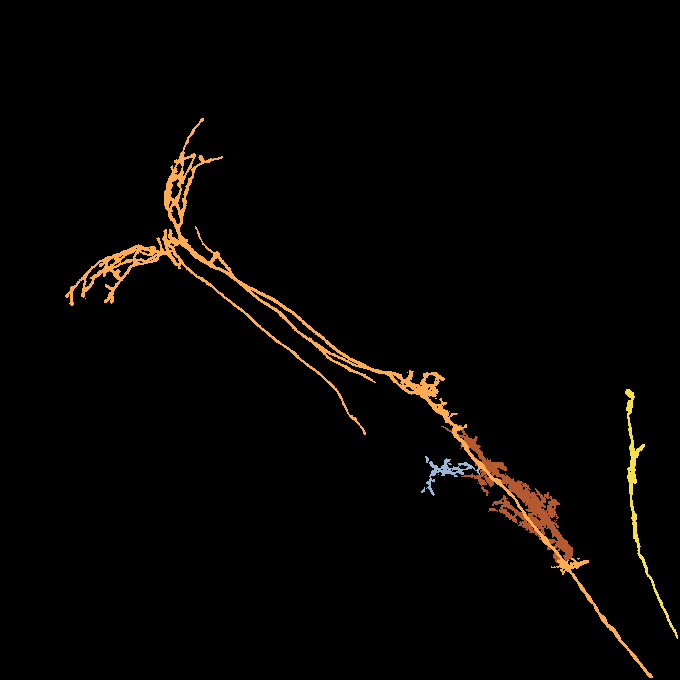}%
    \end{subfigure}%
    \hfill%
    \begin{subfigure}[b]{0.19\textwidth}%
        \includegraphics[width=\textwidth]{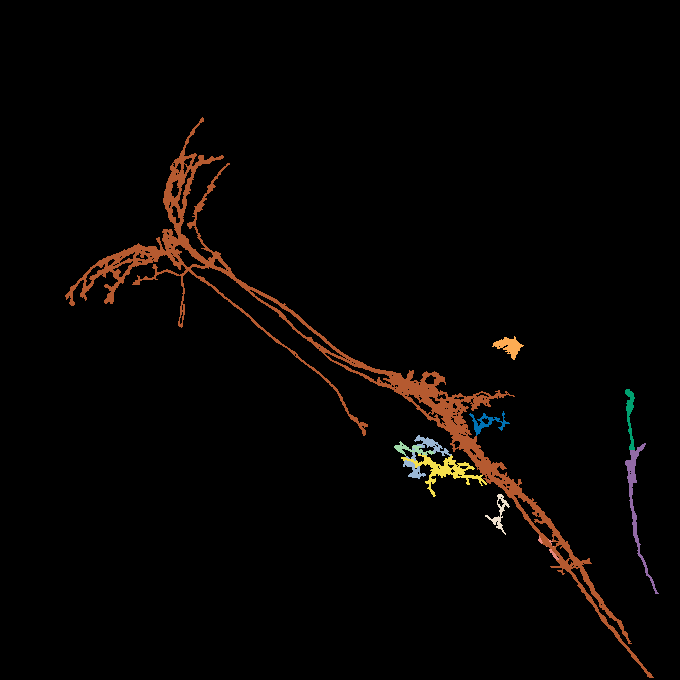}%
    \end{subfigure}%
    \hfill%
    \begin{subfigure}[b]{0.19\textwidth}%
        \includegraphics[width=\textwidth]{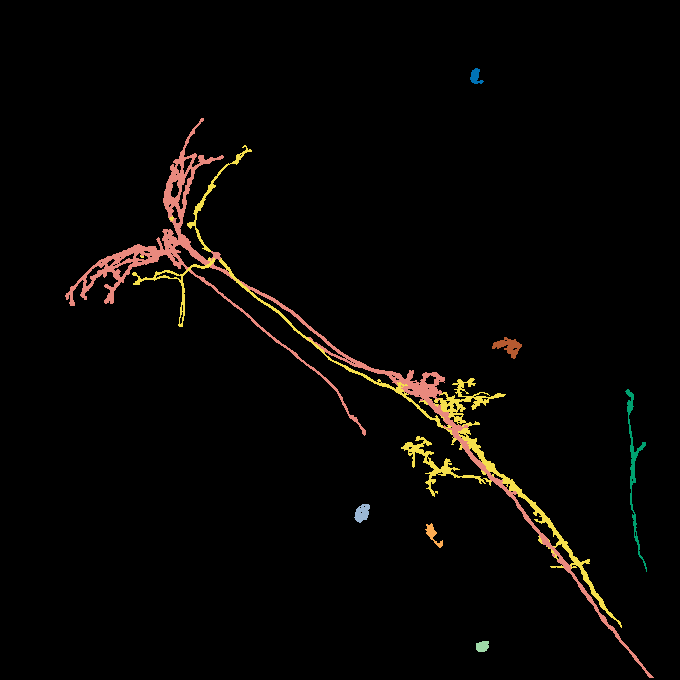}
    \end{subfigure}%
    \hfill%
    \hfill%
    \end{subfigure}%
    
    \begin{subfigure}[b]{\textwidth}%
    \caption*{3) VT027175-20171031\_62\_H6}%
    \hfill%
    \begin{subfigure}[b]{0.19\textwidth}%
        \includegraphics[width=\textwidth]{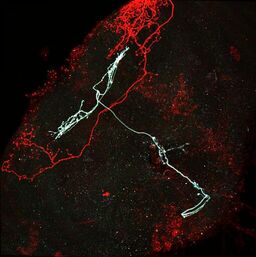}%
    \end{subfigure}%
    \hfill%
    \begin{subfigure}[b]{0.19\textwidth}%
        \includegraphics[width=\textwidth]{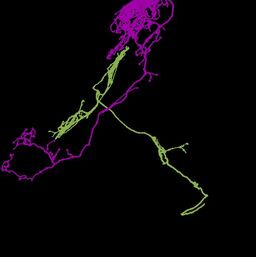}%
    \end{subfigure}%
    \hfill%
    \begin{subfigure}[b]{0.19\textwidth}%
        \includegraphics[width=\textwidth]{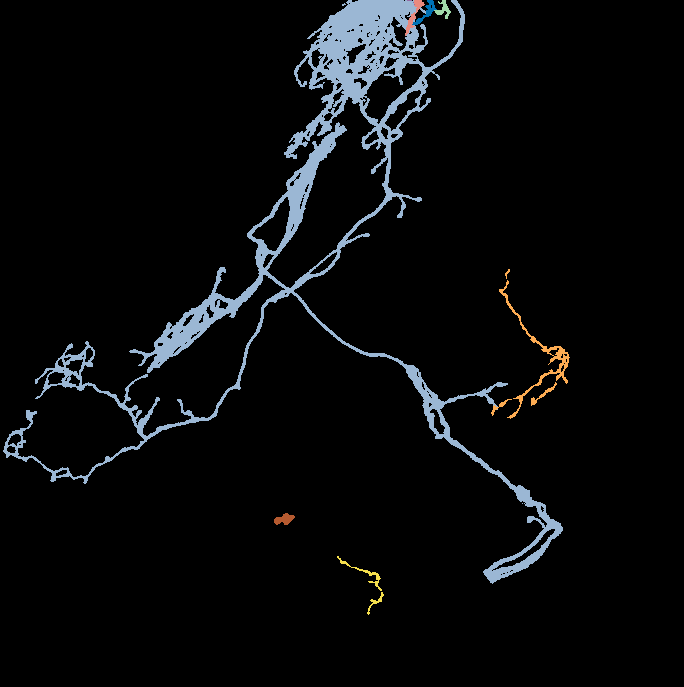}%
    \end{subfigure}%
    \hfill%
    \begin{subfigure}[b]{0.19\textwidth}%
        \includegraphics[width=\textwidth]{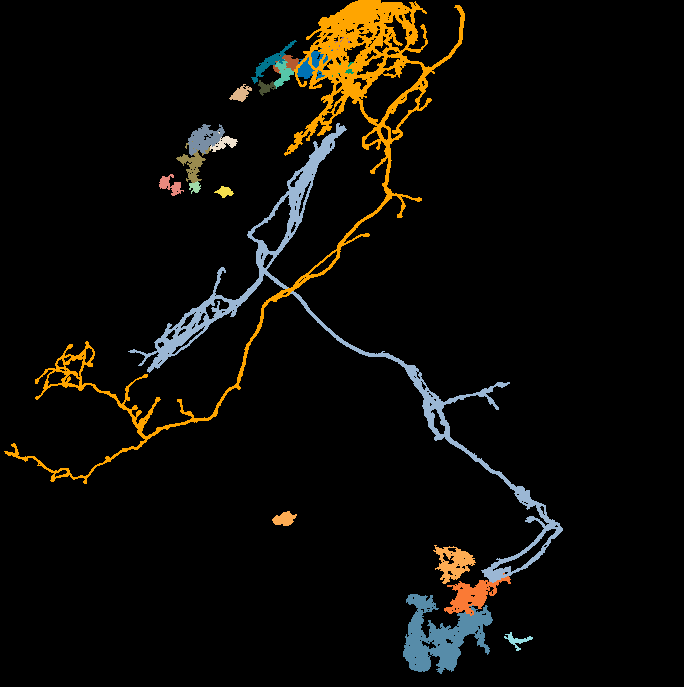}%
    \end{subfigure}%
    \hfill%
    \begin{subfigure}[b]{0.19\textwidth}%
        \includegraphics[width=\textwidth]{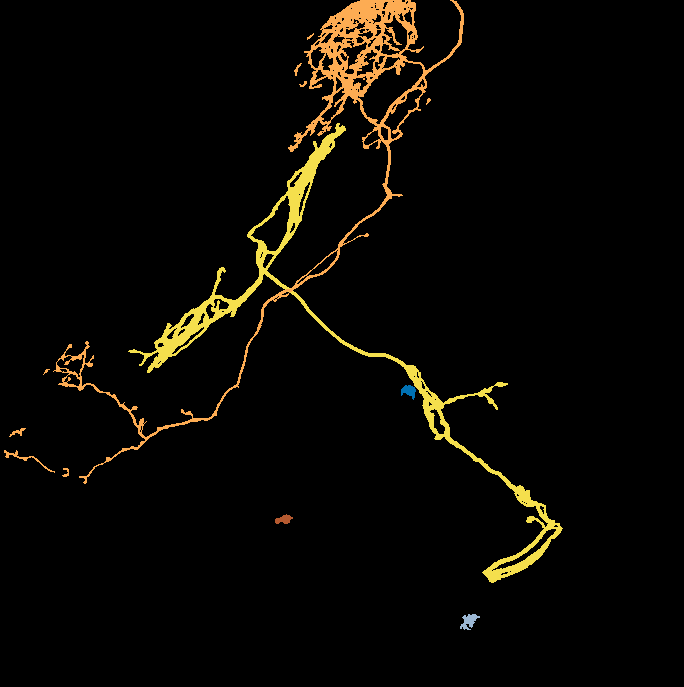}
    \end{subfigure}
    \hfill%
    \hfill%
    \end{subfigure}%

    \begin{subfigure}[b]{\textwidth}%
    \caption*{4) VT028606-20170721\_65\_A3}%
    \hfill%
    \begin{subfigure}[b]{0.19\textwidth}%
        \includegraphics[width=\textwidth]{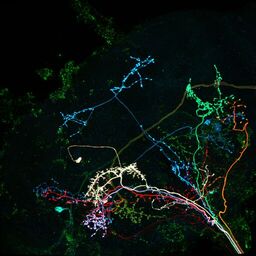}%
        \caption{MCFO}%
    \end{subfigure}%
    \hfill%
    \begin{subfigure}[b]{0.19\textwidth}%
        \includegraphics[width=\textwidth]{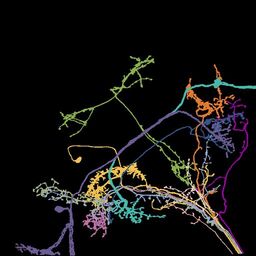}%
        \caption{gt}%
    \end{subfigure}%
    \hfill%
    \begin{subfigure}[b]{0.19\textwidth}%
        \includegraphics[width=\textwidth]{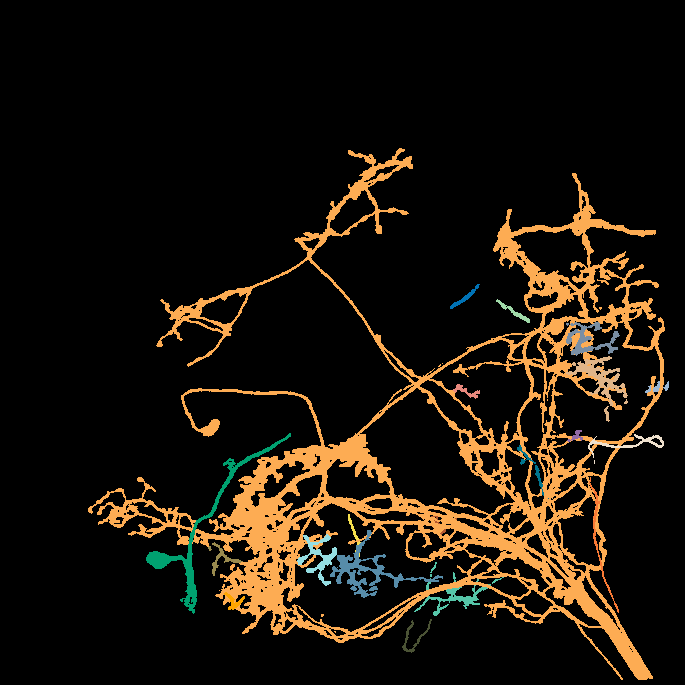}%
        \caption{ppp}
    \end{subfigure}%
    \hfill%
    \begin{subfigure}[b]{0.19\textwidth}%
        \includegraphics[width=\textwidth]{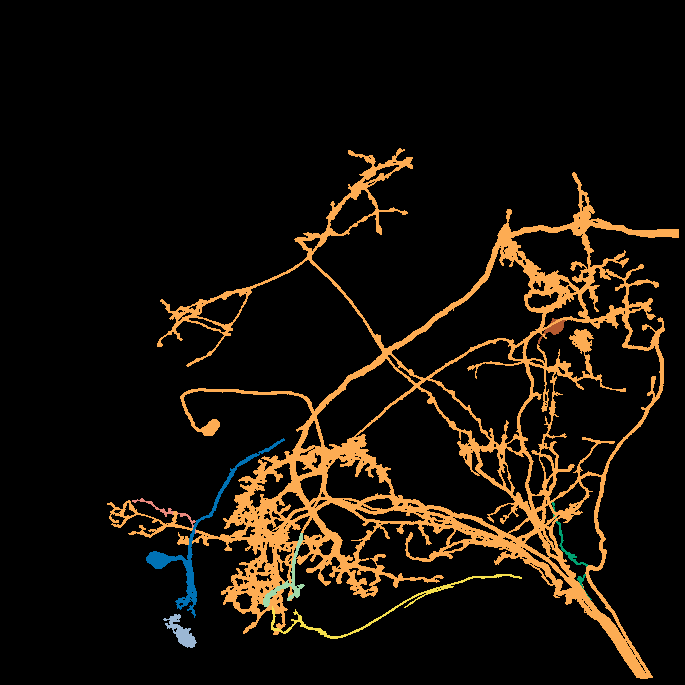}%
        \caption{FFN}%
    \end{subfigure}%
    \hfill%
    \begin{subfigure}[b]{0.19\textwidth}%
        \includegraphics[width=\textwidth]{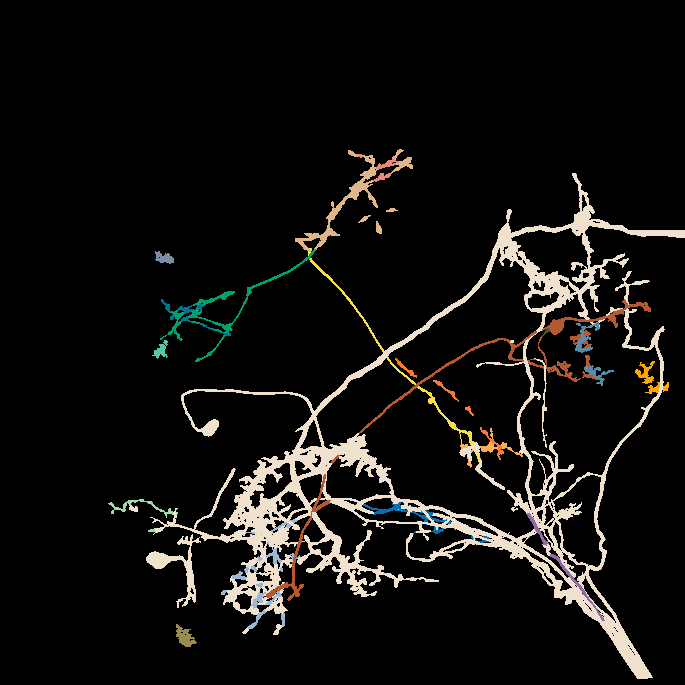}
        \caption{Duan et al.}%
    \end{subfigure}%
    \hfill%
    \hfill%
    \end{subfigure}%
    \caption{Qualitative results for our three baseline methods: PatchPerPix (ppp), Flood Filling Networks (FFN) and Duan et al.'s color clustering. The columns depict the following: (a): Maximum intensity projection of MCFO sample, (b): ground truth segmentation, (c): ppp prediction, (d): FFN prediction, and (e): Duan et al.'s result.
    In (1) all three methods yield some correctly segmented neurons, but ppp merges the blue one and one of the red ones, FFN does not segment most of the red ones and Duan et al.'s merges the blue neuron with parts of the red ones; FFN and Duan et al.'s have lower coverage.
    In (2) the noisy blue channel leads to false positives.
    In (3) ppp merges the two neurons whereas FFN and Duan et al.'s split them correctly; FFN additionally segments some noise.
    In (4) all three methods merge multiple neurons of different color; Duan et al.'s has lower coverage. \label{suppl_fig:comparative_results}}
\end{figure*}

\begin{figure*}[htbp]
	\centering%
    \begin{subfigure}[b]{\textwidth}%
    \caption*{1) VT027175-20171031\_62\_H3}%
    \hfill%
    \begin{subfigure}[b]{0.19\textwidth}%
        \includegraphics[width=\textwidth]{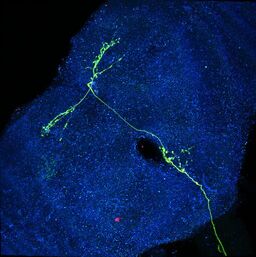}%
    \end{subfigure}%
    \hfill%
    \begin{subfigure}[b]{0.19\textwidth}%
        \includegraphics[width=\textwidth]{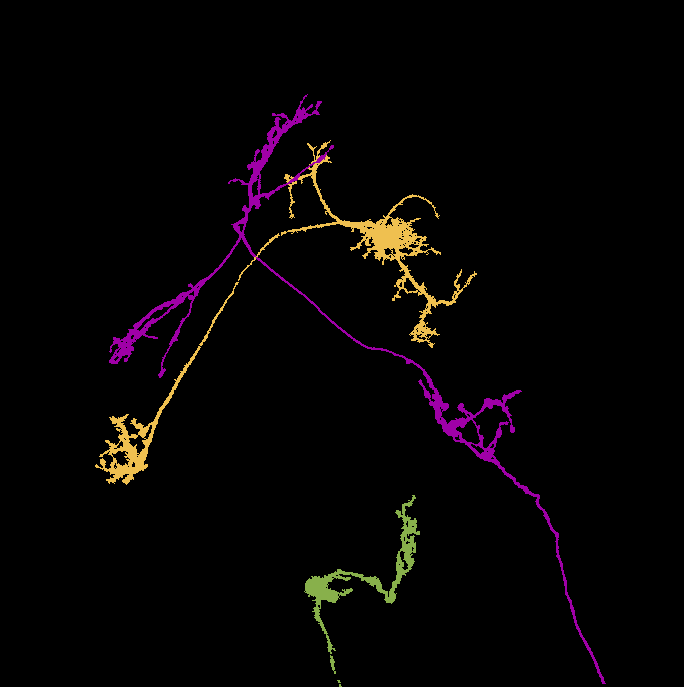}%
    \end{subfigure}%
    \hfill%
    \begin{subfigure}[b]{0.19\textwidth}%
        \includegraphics[width=\textwidth]{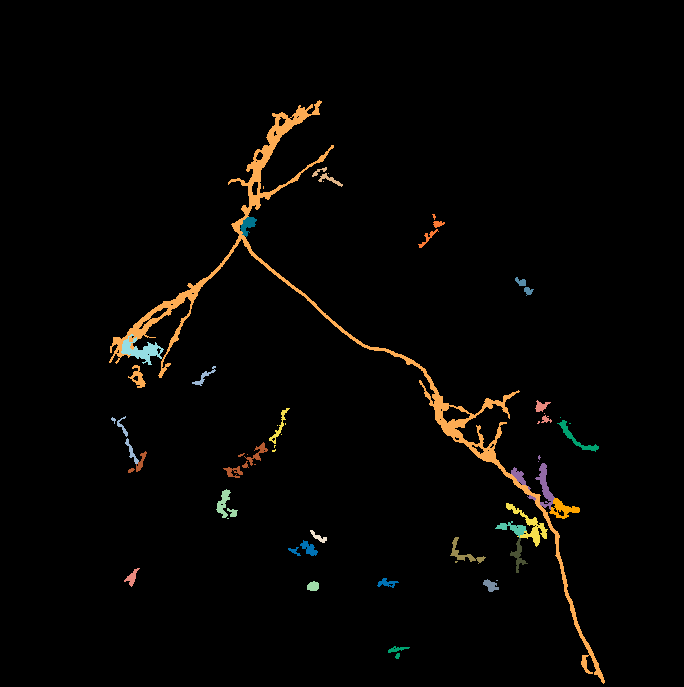}%
    \end{subfigure}%
    \hfill%
    \begin{subfigure}[b]{0.19\textwidth}%
        \includegraphics[width=\textwidth]{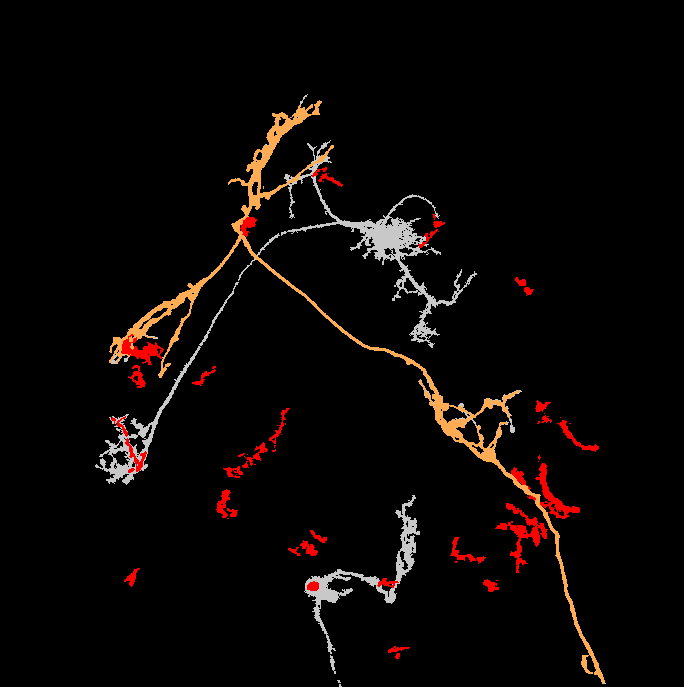}%
    \end{subfigure}%
    \hfill%
    \begin{subfigure}[b]{0.19\textwidth}%
        \includegraphics[width=\textwidth]{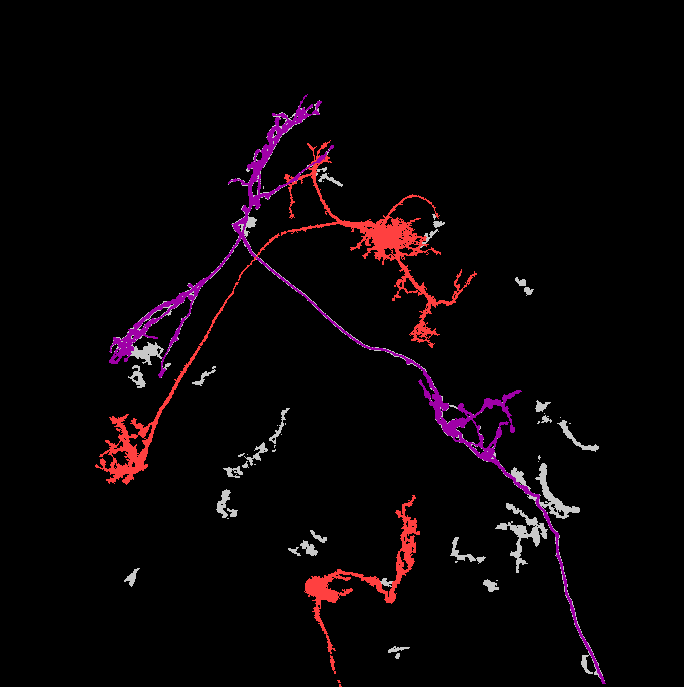}%
    \end{subfigure}%
    \hfill%
    \hfill%
    \end{subfigure}%
    
    \begin{subfigure}[b]{\textwidth}%
    \caption*{2) R14A02-20180905\_65\_A6}%
    \hfill%
    \begin{subfigure}[b]{0.19\textwidth}%
        \includegraphics[width=\textwidth]{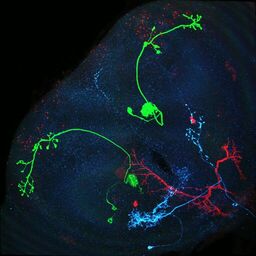}%
    \end{subfigure}%
    \hfill%
    \begin{subfigure}[b]{0.19\textwidth}%
        \includegraphics[width=\textwidth]{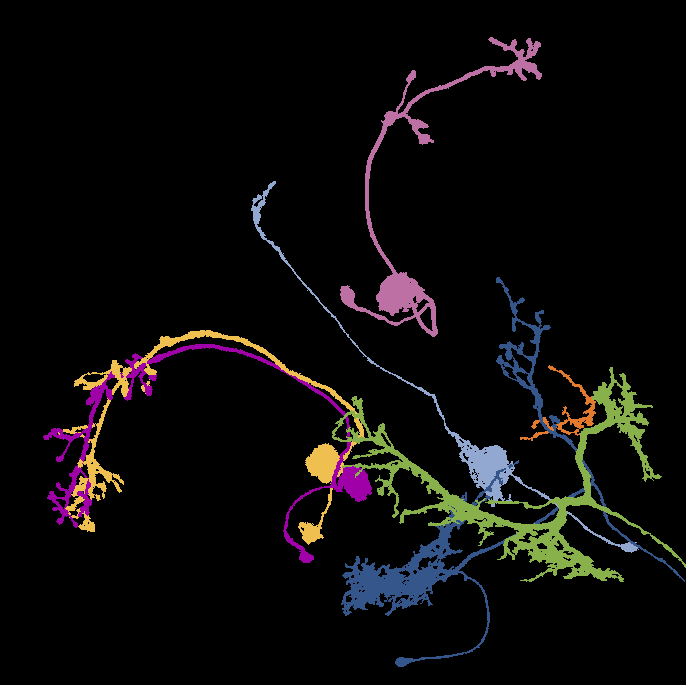}%
    \end{subfigure}%
    \hfill%
    \begin{subfigure}[b]{0.19\textwidth}%
        \includegraphics[width=\textwidth]{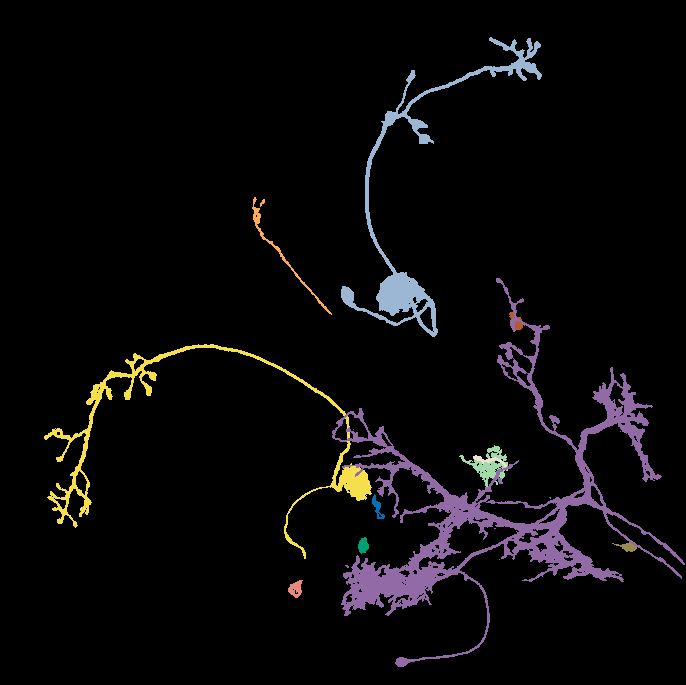}%
    \end{subfigure}%
    \hfill%
    \begin{subfigure}[b]{0.19\textwidth}%
        \includegraphics[width=\textwidth]{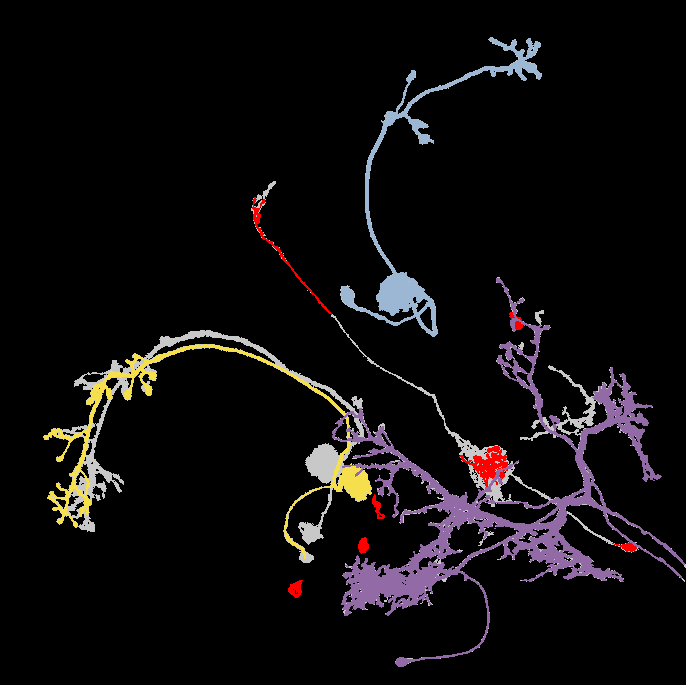}%
    \end{subfigure}%
    \hfill%
    \begin{subfigure}[b]{0.19\textwidth}%
        \includegraphics[width=\textwidth]{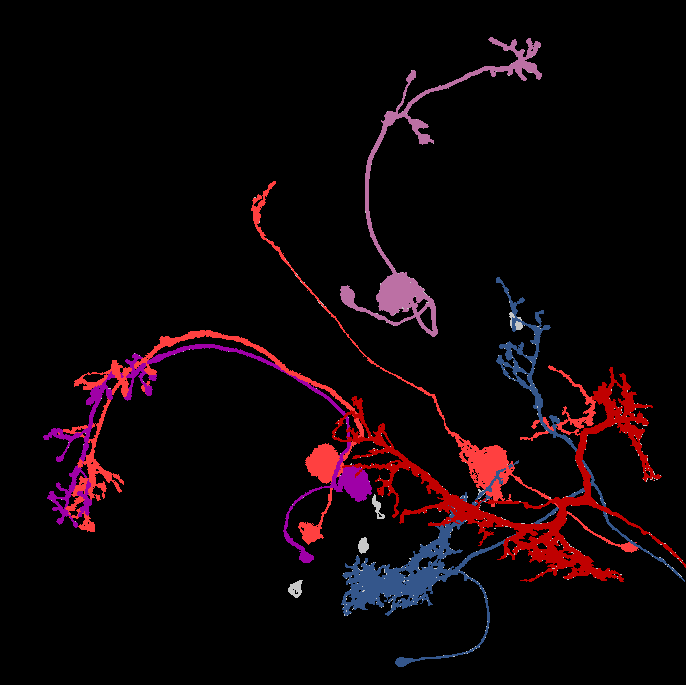}%
    \end{subfigure}%
    \hfill%
    \hfill%
    \end{subfigure}%

    \begin{subfigure}[b]{\textwidth}%
    \caption*{3) VT058571-20170926\_64\_G6}%
    \hfill%
    \begin{subfigure}[b]{0.19\textwidth}%
        \includegraphics[width=\textwidth]{figures/all_samples/partly/test/raw/VT058571-20170926_64_G6.jpg}%
    \end{subfigure}%
    \hfill%
    \begin{subfigure}[b]{0.19\textwidth}%
        \includegraphics[width=\textwidth]{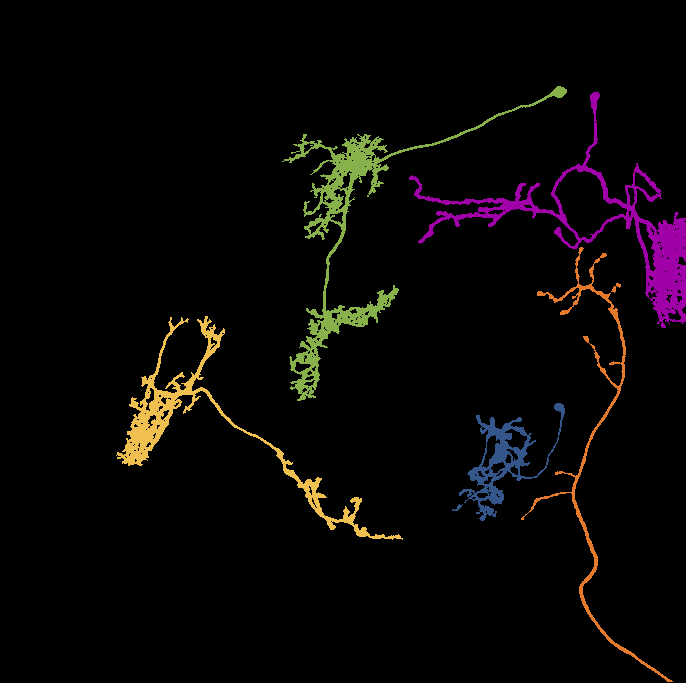}%
    \end{subfigure}%
    \hfill%
    \begin{subfigure}[b]{0.19\textwidth}%
        \includegraphics[width=\textwidth]{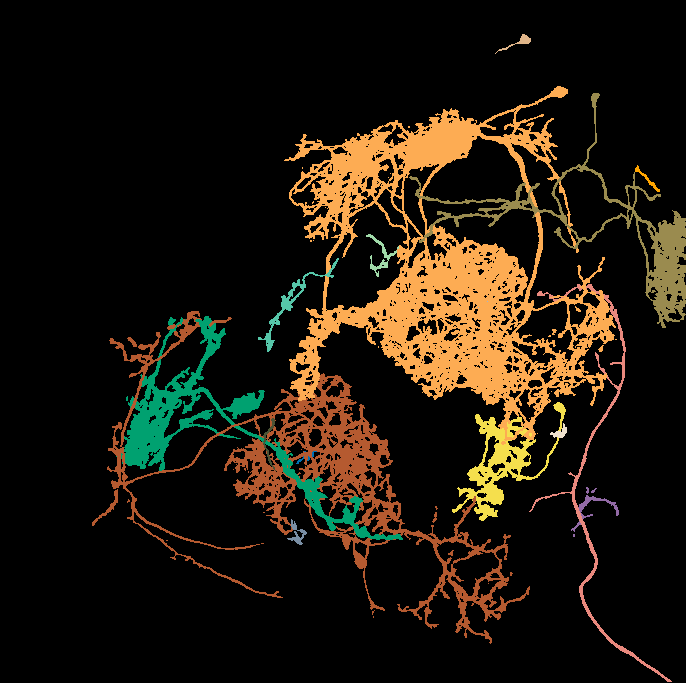}%
    \end{subfigure}%
    \hfill%
    \begin{subfigure}[b]{0.19\textwidth}%
        \includegraphics[width=\textwidth]{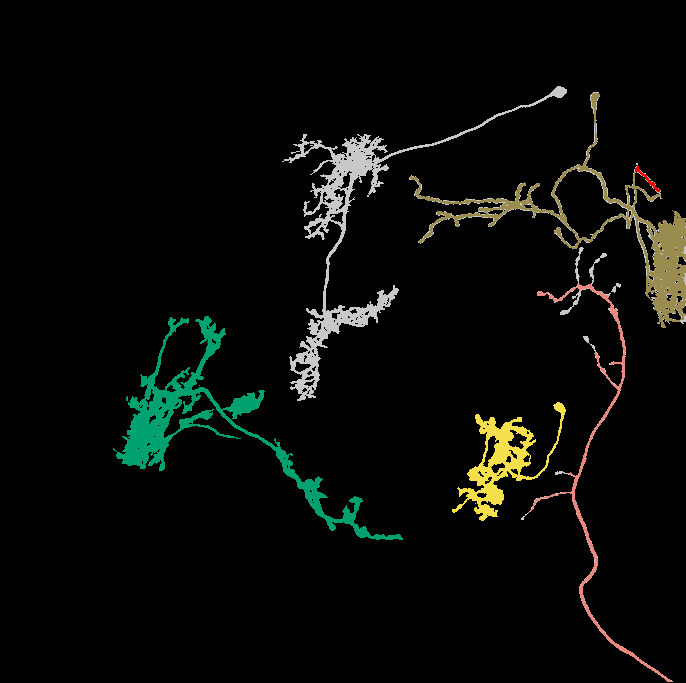}%
    \end{subfigure}%
    \hfill%
    \begin{subfigure}[b]{0.19\textwidth}%
        \includegraphics[width=\textwidth]{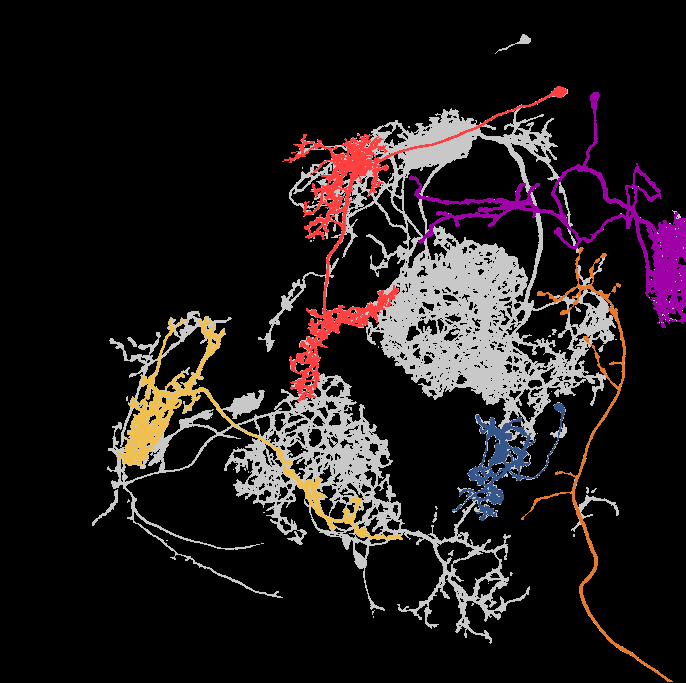}%
    \end{subfigure}%
    \hfill%
    \hfill%
    \end{subfigure}%

    \begin{subfigure}[b]{\textwidth}%
    \caption*{4) VT011145-20171222\_63\_I2}%
    \hfill%
    \begin{subfigure}[b]{0.19\textwidth}%
        \includegraphics[width=\textwidth]{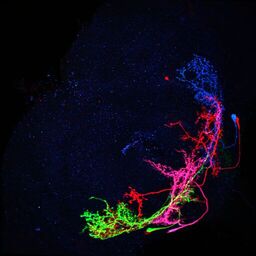}%
        \caption{MCFO}%
    \end{subfigure}%
    \hfill%
    \begin{subfigure}[b]{0.19\textwidth}%
        \includegraphics[width=\textwidth]{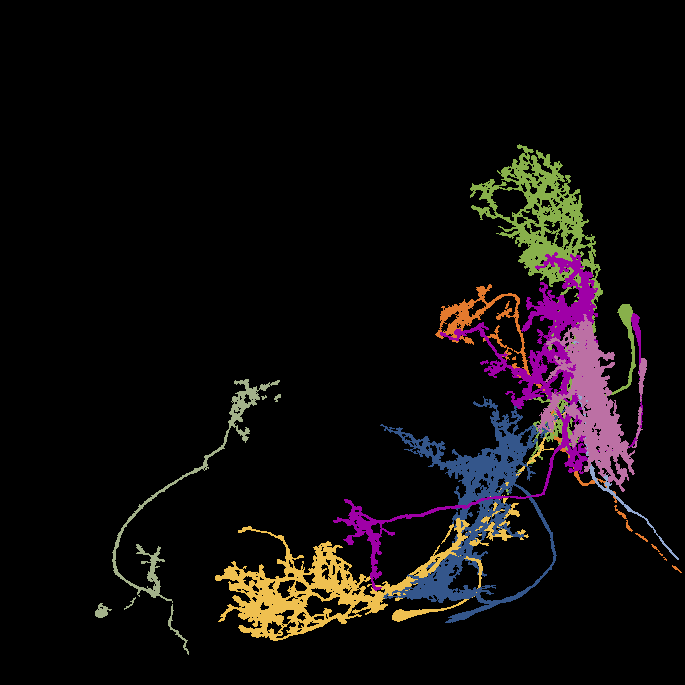}%
        \caption{gt}%
    \end{subfigure}%
    \hfill%
    \begin{subfigure}[b]{0.19\textwidth}%
        \includegraphics[width=\textwidth]{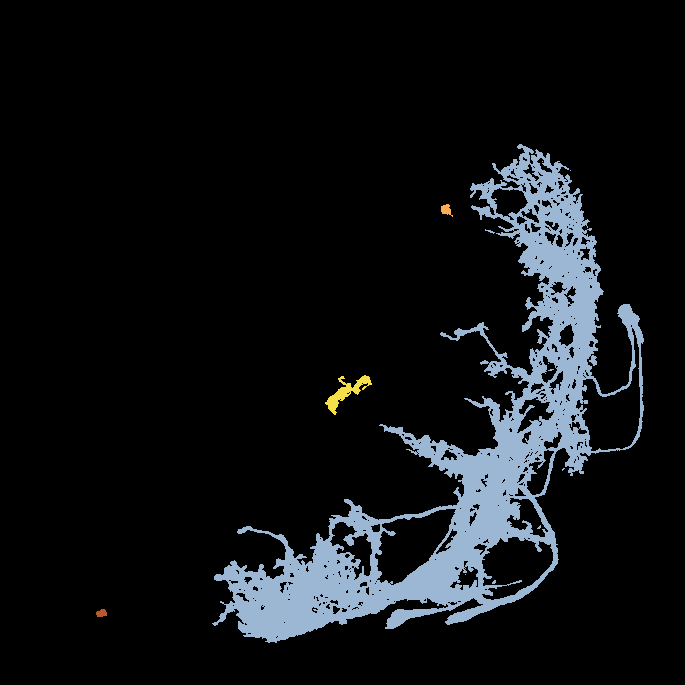}%
        \caption{prediction}
    \end{subfigure}%
    \hfill%
    \begin{subfigure}[b]{0.19\textwidth}%
        \includegraphics[width=\textwidth]{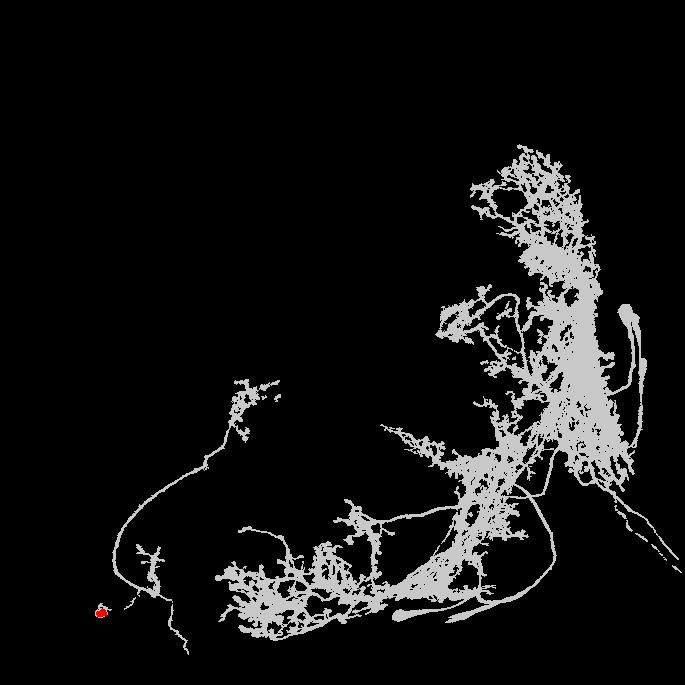}%
        \caption{FP/FS}%
    \end{subfigure}%
    \hfill%
    \begin{subfigure}[b]{0.19\textwidth}%
        \includegraphics[width=\textwidth]{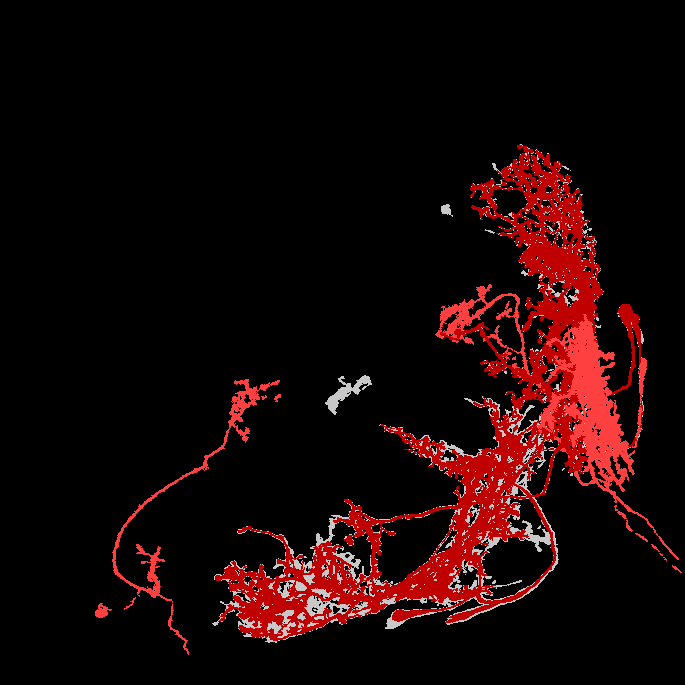}%
        \caption{FN/FM}%
    \end{subfigure}%
    \hfill%
    \hfill%
    \end{subfigure}%
    \caption{\label{suppl_fig:vis_res_ppp_completely}Visualization of the results of our PatchPerPix baseline model for four samples of our completely labeled test set, ground truth-prediction matches are shown for a clDice threshold of 0.5.
    ((a): Maximum intensity projection of MCFO sample; (b): ground truth segmentation; (c): predicted segmentation; (d): TP predicted instances in same color code as in (c), FP and FS in red, ground truth in grey; (e): TP ground truth instances in same color code as in (b), FN in light red, FM in dark red, prediction mask in grey.)
In (1) the bright green neuron is nicely segmented (see (c), in orange). 
However, there are two more, very dim neurons in the image, these were missed (see (e) in red).
In addition, there is a large number of FP (see (d) in red).
In (2) the bright green ones are again nicely segmented (see (c), in yellow and light blue). 
The purple prediction (see (c)) covers most of the blue and the red neurons, unfortunately resulting in a false merge (FM), despite having very different colors. 
The blue one still counts as a TP, the red one though as a FN, more precisely a FM (shown in dark red in (e)). 
There is a very dim red neuron next to the left green one that is missed completely (shown in bright red in (e)). 
There is a dim blue neuron between the two green ones that is mostly missed resulting in a few FS (shown in red in (d)).
In (3) there are a number of unlabeled neurons. 
There are two bright red neurons, only one is labeled (shown by the absence of a label in (b)). 
There are a couple of somewhat dim neurons of different colors (there are still visible relatively well when zooming in). 
We can see that our prediction, as desired, includes the unlabeled neurons (see (c)).
We can also see that, as they are not shown in (d) and again as desired, they are counted neither as TP nor as FP.
There are again some FM, the bright blue neuron is segmented well but unfortunately the prediction is merged with other neurons (note that it is not shown in dark red (FM) in (e) as it is merged with unlabeled neurons, thus it is not possible to automatically tag it as a FM).
(4) shows an extreme FM case. 
There is a cluster of multiple overlapping bright neurons in different colors (see (a) and (b)).
In the prediction they are all merged (shown in dark red in (e)), thus there is no TP (shown by the absence of colored segmentation masks in (d)).
In addition there are a number of dim neurons that have been overlooked by the model (shown in bright red in (e)).}
\end{figure*}

We describe our three baseline methods in the following sections, namely PatchPerPix in Sec.~\ref{suppl:ppp_info}, Flood Filling Networks in Sec.~\ref{suppl:ffn_info} and Duan et al.'s color clustering in Sec.~\ref{suppl:brainbow_info}.
The evaluation code is available here: \iftoggle{cvprfinal}{
\href{https://github.com/Kainmueller-Lab/evaluate-instance-segmentation}{https://github.com/Kainmueller-Lab/evaluate-instance-segmentation}}{\textit{Hidden during review process}}.
Please see Table~\ref{suppl_tab:results_both_sets} and \ref{suppl_tab:results_both_sets_f_scores} for the extended quantitative results.
Qualitative results for all three baselines are shown in Fig.~\ref{suppl_fig:comparative_results} and a visualization of typical error types for PatchPerPix is presented in Fig.~\ref{suppl_fig:vis_res_ppp_completely}.

\begin{table*}[bpht]
	\centering
    \caption{\label{suppl_tab:results_both_sets}Quantitative results of our baseline models on the \textit{combined}, \textit{completely} and \textit{partly} labeled FISBe datasets. We train models both only on the completely labeled data (ppp, FFN), and on the completely and the partly labeled data (FFN+partly). Note that the scores are not directly comparable to each other across datasets (combined, completely, partly), but they are comparable across splits (val, test) and methods within each dataset. We report mean and standard de\-vi\-a\-tion (\std{}) over three independent runs (except for Duan et al.'s as it is non-learnt). For all scores except FS and FM higher values are better. Continued in Table~\ref{suppl_tab:results_both_sets_f_scores}.}
    \begin{tblr}{width=\linewidth,colspec={X[0.5,l] X[1.4,l] J[1.2] J[1.2] J[1.2] J[1.2] J[1.1] J[1.1] J[1.2] J[1.2] J[1.2] J[1.2] J[1.2]},row{1}={guard}}
        \midrule
		  Split & Method & \(S\) & \(avF1\) & \(C\) & \(clDice_{TP}\) & \(FS\) & \(FM\) & \(C_{dim}\) & \(C_{ovlp}\) & \(tp\) & \(tp_{dim}\)  & \(tp_{ovlp}\)\\
        \midrule
        \SetCell[c=13]{c}Combined\\
		\midrule
		\SetCell[r=4]{m}Val & ppp & 0.38\std{0.02} & 0.41\std{0.02} & 0.35\std{0.01} & 0.75\std{0.02} & 6.0\std{0.8} & 24\std{1.6} & 0.12\std{0.01} & 0.38 \std{0.04} & 0.46 \std{0.01} & 0.16 \std{0.04} & 0.39 \std{0.03} \\
        & FFN & 0.25\std{0.01} & 0.27\std{0.01} & 0.23\std{0.01} & 0.79\std{0.01} & 7.0\std{2.9} & 12\std{2.0} & 0.03\std{0.01} & 0.30 \std{0.01} & 0.32 \std{0.01} & 0.04 \std{0.01} & 0.37 \std{0.02} \\
        & FFN+partly & 0.27\std{0.01} & 0.29\std{0.01} & 0.24\std{0.01} & 0.79\std{0.01} & 7.7\std{2.6} & 14\std{0.8} & 0.02\std{0.01} & 0.33 \std{0.02} & 0.34 \std{0.03} & 0.03 \std{0.00} & 0.38 \std{0.04} \\
        & \mbox{Duan et al.} & 0.24 & 0.26 & 0.22 & 0.70 & 14 & 13 & 0.02 & 0.28 & 0.37 & 0.03 & 0.42 \\
        \midrule
        \SetCell[r=4]{m}Test & ppp & 0.35\std{0.00} & 0.34\std{0.01} & 0.35\std{0.01} & 0.80\std{0.00} & 19\std{2.9} & 52\std{3.4} & 0.16\std{0.03} & 0.27 \std{0.04} & 0.36 \std{0.01} & 0.19 \std{0.04} & 0.19 \std{0.03} \\
        & FFN & 0.25\std{0.03} & 0.22\std{0.04} & 0.29\std{0.02} & 0.80\std{0.01} & 17\std{1.7} & 39\std{5.3} & 0.03\std{0.01} & 0.26 \std{0.03} & 0.32 \std{0.03} & 0.00 \std{0.00} & 0.24 \std{0.05} \\
        & FFN+partly & 0.27\std{0.01} & 0.24\std{0.02} & 0.31\std{0.00} & 0.80\std{0.01} & 18\std{3.7} & 36\std{3.6} & 0.04\std{0.01} & 0.28 \std{0.01} & 0.36 \std{0.01} & 0.03 \std{0.00} & 0.28 \std{0.01} \\
        & \mbox{Duan et al.} & 0.30  & 0.27 & 0.33 & 0.77 & 45 & 29 & 0.03 & 0.36 & 0.37 & 0.03 & 0.34 \\
        \bottomrule
        \SetCell[c=13]{c}Completely\\ 
        \midrule
		\SetCell[r=4]{m}Val & ppp & 0.30\std{0.03} & 0.34\std{0.04} & 0.27\std{0.03} & 0.72\std{0.02} & 1.7\std{1.7} & 3.7\std{1.3} & 0.07\std{0.01} & 0.41 \std{0.07} & 0.37 \std{0.06} & 0.06 \std{0.04} & 0.47 \std{0.12} \\
        & FFN & 0.18\std{0.01} & 0.21\std{0.01} & 0.15\std{0.02} & 0.78\std{0.02} & 0.3\std{0.5} & 0.0\std{0.0} & 0.10\std{0.00} & 0.28 \std{0.02} & 0.21 \std{0.02} & 0.00 \std{0.00} & 0.40 \std{0.00} \\
        & FFN+partly & 0.20\std{0.01} & 0.24\std{0.02} & 0.17\std{0.00} & 0.81\std{0.02} & 0.7\std{0.5} & 0.3\std{0.5} & 0.00\std{0.00} & 0.32 \std{0.01} & 0.22 \std{0.02} &  0.00\std{0.00} & 0.40 \std{0.00} \\
        & \mbox{Duan et al.} & 0.16 & 0.17 & 0.15 & 0.65 & 2 &  1 & 0.00 & 0.25 & 0.24 & 0.00 & 0.40 \\
        \midrule
        \SetCell[r=4]{m}Test & ppp & 0.34\std{0.02} & 0.29\std{0.04} & 0.40\std{0.02} & 0.81\std{0.02} & 3.0\std{0.8} & 4.3\std{1.3} & 0.14\std{0.05} & 0.42 \std{0.03} & 0.45 \std{0.01} & 0.19 \std{0.09} & 0.38 \std{0.10} \\
        & FFN & 0.18\std{0.04} & 0.11\std{0.08} & 0.26\std{0.01} & 0.77\std{0.05} & 2.0\std{0.8} & 2.0\std{1.4} & 0.02\std{0.02} & 0.24 \std{0.03} & 0.31 \std{0.06} & 0.00 \std{0.00} & 0.25 \std{0.10}\\
        & FFN+partly & 0.19\std{0.02} & 0.10\std{0.03} & 0.29\std{0.02} & 0.80\std{0.01} & 2.3\std{0.5} & 1.7\std{0.5} & 0.03\std{0.01} & 0.32 \std{0.04} & 0.34 \std{0.06} & 0.02 \std{0.03} & 0.42 \std{0.12} \\
        & \mbox{Duan et al.} & 0.28 & 0.23 & 0.33 & 0.81 & 6 & 1 & 0.02 & 0.43 & 0.38 & 0.00 & 0.50 \\
        \bottomrule
        \SetCell[c=13]{c}Partly\\ 
        \midrule
		\SetCell[r=4]{m}Val & ppp & 0.48\std{0.00} & 0.52\std{0.00} & 0.45\std{0.01} & 0.79\std{0.00} & 3.7\std{0.9} & 20\std{0.0} & 0.19\std{0.01} & 0.35 \std{0.01} & 0.51 \std{0.0} & 0.25 \std{0.02} & 0.37 \std{0.00} \\
		& FFN & 0.34\std{0.01} & 0.36\std{0.01} & 0.32\std{0.01} & 0.79\std{0.01} & 7.3\std{2.9}& 12\std{1.6} & 0.06\std{0.02} & 0.33 \std{0.00} & 0.37 \std{0.02} & 0.06 \std{0.02} & 0.39 \std{0.03} \\
        & FFN+partly & 0.34\std{0.02} & 0.36\std{0.03} & 0.32\std{0.01} & 0.77 \std{0.01} & 8.0\std{1.6} & 13 \std{0.9} & 0.04\std{0.02} & 0.34 \std{0.02} & 0.38 \std{0.03} & 0.03 \std{0.02} & 0.39 \std{0.04}\\
        & \mbox{Duan et al.} & 0.32 & 0.35 & 0.30 & 0.74 & 12 & 12 & 0.04 & 0.31 & 0.41 & 0.04 & 0.43\\
        \midrule
        \SetCell[r=4]{m}Test & ppp & 0.35\std{0.01} & 0.40\std{0.00} & 0.31\std{0.02} & 0.79\std{0.01} & 15\std{1.9}&  46\std{1.7} &  0.15\std{0.02} & 0.15 \std{0.02} & 0.33 \std{0.01} & 0.13 \std{0.02} & 0.17 \std{0.02}\\
		& FFN & 0.33\std{0.02} & 0.35\std{0.02} & 0.32\std{0.02} & 0.80\std{0.01} & 16\std{2.5} & 37\std{4.0} & 0.03\std{0.01} & 0.23 \std{0.03} & 0.34 \std{0.02} & 0.00 \std{0.00} & 0.23 \std{0.04} \\
        & FFN+partly & 0.34\std{0.02} & 0.36\std{0.03} & 0.33\std{0.02} & 0.82\std{0.01} & 17\std{2.5} & 35\std{3.7} & 0.04\std{0.01} & 0.25 \std{0.03} & 0.34 \std{0.02} & 0.05 \std{0.00} & 0.25 \std{0.01} \\
        & \mbox{Duan et al.} & 0.33 & 0.32 & 0.34 & 0.73 & 39 & 28 & 0.04 & 0.28 & 0.37 & 0.05 & 0.32 \\
        \bottomrule
    \end{tblr}
 \end{table*}

\begin{table*}[bpht]
	\centering
    \caption{\label{suppl_tab:results_both_sets_f_scores}Quantitative results of our baseline models on \textit{combined}, \textit{completely} and \textit{partly} labeled FISBe datasets (continuation of Table~\ref{suppl_tab:results_both_sets}).}
    \begin{tblr}{width=\linewidth,colspec={X[0.4,l] X[1.3,l] J[1.2] J[1.2] J[1.2] J[1.2] J[1.2] J[1.2] J[1.2] J[1.2] J[1.2]},row{1}={guard}}
        \midrule
		  Split & Method & F1$_{0.1}$ & F1$_{0.2}$ & F1$_{0.3}$ & F1$_{0.4}$ & F1$_{0.5}$ & F1$_{0.6}$ & F1$_{0.7}$ & F1$_{0.8}$ & F1$_{0.9}$\\
		\midrule
        \SetCell[c=11]{c}Combined\\
		\midrule
		\SetCell[r=4]{m}Val  & ppp & 0.62\std{0.03} & 0.57\std{0.03} & 0.53\std{0.01} & 0.50\std{0.02} & 0.49\std{0.02} & 0.43\std{0.01} & 0.28\std{0.03} & 0.22\std{0.05} & 0.10\std{0.02}\\
        & FFN & 0.38\std{0.01} & 0.36\std{0.01} & 0.34\std{0.01} & 0.32\std{0.00} & 0.30\std{0.01} & 0.27\std{0.01} & 0.22\std{0.02} & 0.17\std{0.01} & 0.07\std{0.01}\\
        & FFN+partly & 0.40\std{0.02} & 0.38\std{0.02} & 0.37\std{0.02} & 0.35\std{0.01} & 0.34\std{0.01} & 0.31\std{0.01} & 0.24\std{0.01} & 0.18\std{0.01} & 0.07\std{0.01}\\
        & Duan et al. & 0.38 & 0.37 & 0.34 & 0.33 & 0.33 & 0.27 & 0.18 & 0.09 & 0.03\\
        \midrule
        \SetCell[r=4]{m}Test & ppp & 0.50\std{0.01} & 0.48\std{0.01} & 0.44\std{0.01} & 0.41\std{0.02} & 0.35\std{0.02} & 0.29\std{0.02} & 0.26\std{0.01} & 0.19\std{0.02} & 0.12\std{0.01}\\
        & FFN & 0.34\std{0.05} & 0.31\std{0.04} & 0.28\std{0.04} & 0.25\std{0.05} & 0.22\std{0.04} & 0.20\std{0.04} & 0.17\std{0.03} & 0.12\std{0.01} & 0.07\std{0.01}\\
        & FFN+partly & 0.36\std{0.02} & 0.32\std{0.02} & 0.30\std{0.02} & 0.27\std{0.03} & 0.25\std{0.03} & 0.21\std{0.03} & 0.18\std{0.02} & 0.15\std{0.02} & 0.09\std{0.01}\\
        & Duan et al. & 0.43 & 0.38 & 0.35 & 0.33 & 0.31 & 0.29 & 0.20 & 0.12 & 0.06\\
		\midrule
        \SetCell[c=11]{c}Completely\\
		\midrule
		\SetCell[r=4]{m}Val & ppp & 0.57\std{0.07} & 0.50\std{0.05} & 0.46\std{0.02} & 0.43\std{0.05} & 0.41\std{0.07} & 0.34\std{0.03} & 0.19\std{0.02} & 0.14\std{0.07} & 0.04\std{0.02}\\
        & FFN & 0.31\std{0.04} & 0.28\std{0.04} & 0.26\std{0.03} & 0.24\std{0.01} & 0.24\std{0.01} & 0.22\std{0.03} & 0.16\std{0.03} & 0.13\std{0.01} & 0.04\std{0.03}\\
        & FFN+partly & 0.32\std{0.03} & 0.30\std{0.01} & 0.28\std{0.02} & 0.26\std{0.04} & 0.26\std{0.04} & 0.24\std{0.03} & 0.23\std{0.02} & 0.15\std{0.02} & 0.08\std{0.03}\\
        & Duan et al. & 0.24 & 0.24 & 0.24 & 0.24 & 0.24 & 0.20 & 0.10 & 0.00 & 0.00\\
        \midrule
		\SetCell[r=4]{m}Test & ppp & 0.40\std{0.04} & 0.38\std{0.03} & 0.37\std{0.03} & 0.34\std{0.05} & 0.30\std{0.04} & 0.27\std{0.06} & 0.23\std{0.05} & 0.18\std{0.04} & 0.11\std{0.01}\\
        & FFN & 0.16\std{0.11} & 0.16\std{0.11} & 0.15\std{0.10} & 0.14\std{0.10} & 0.12\std{0.08} & 0.09\std{0.06} & 0.08\std{0.05} & 0.05\std{0.04} & 0.02\std{0.02}\\
        & FFN+partly & 0.15\std{0.04} & 0.15\std{0.04} & 0.14\std{0.04} & 0.12\std{0.04} & 0.11\std{0.04} & 0.10\std{0.03} & 0.08\std{0.02} & 0.07\std{0.03} & 0.03\std{0.01}\\
        & Duan et al. & 0.31 & 0.29 & 0.27 & 0.27 & 0.27 & 0.27 & 0.20 & 0.14 & 0.06\\
		\midrule
        \SetCell[c=11]{c}Partly\\
		\midrule
		\SetCell[r=4]{m}Val & ppp & 0.73\std{0.01} & 0.68\std{0.01} & 0.65\std{0.01} & 0.61\std{0.01} & 0.59\std{0.00} & 0.54\std{0.02} & 0.41\std{0.03} & 0.28\std{0.00} & 0.19\std{0.01}\\
        & FFN & 0.49\std{0.01} & 0.47\std{0.01} & 0.45\std{0.02} & 0.41\std{0.01} & 0.40\std{0.01} & 0.37\std{0.01} & 0.29\std{0.03} & 0.22\std{0.02} & 0.10\std{0.02}\\
        & FFN+partly & 0.50\std{0.04} & 0.47\std{0.04} & 0.46\std{0.05} & 0.42\std{0.05} & 0.41\std{0.06} & 0.36\std{0.05} & 0.26\std{0.02} & 0.22\std{0.00} & 0.09\std{0.02}\\
        & Duan et al. & 0.51 & 0.49 & 0.44 & 0.42 & 0.42 & 0.34 & 0.25 & 0.18 & 0.06\\
		\midrule
		\SetCell[r=4]{m}Test & ppp & 0.62\std{0.02} & 0.58\std{0.01} & 0.51\std{0.01} & 0.47\std{0.00} & 0.40\std{0.01} & 0.33\std{0.01} & 0.28\std{0.01} & 0.21\std{0.02} & 0.17\std{0.01}\\
        & FFN & 0.55\std{0.04} & 0.49\std{0.01} & 0.44\std{0.02} & 0.40\std{0.02} & 0.36\std{0.02} & 0.32\std{0.03} & 0.27\std{0.02} & 0.21\std{0.01} & 0.11\std{0.01}\\
        & FFN+partly & 0.57\std{0.06} & 0.51\std{0.04} & 0.47\std{0.03} & 0.40\std{0.04} & 0.36\std{0.03} & 0.31\std{0.03} & 0.28\std{0.03} & 0.23\std{0.02} & 0.14\std{0.02}\\
        & Duan et al. & 0.55 & 0.48 & 0.43 & 0.39 & 0.35 & 0.31 & 0.19 & 0.09 & 0.06\\
		\bottomrule
    \end{tblr}
\end{table*}

\subsubsection{PatchPerPix}\label{suppl:ppp_info}
We use PatchPerPix \cite{ppp2020} with a 3-level 3d U-Net~\cite{ronneberger2015u,cicek16_3d_unet} with 20 initial feature maps, tripled at each downsampling layer. 
The predicted patches are of size $7\times 7 \times 7$ pixels.
We use the base model without the additional patch decoder.
In addition to the patches the model is trained to predict how many instances there are per pixel (\textit{numinst} in the code) modelled as a categorical prediction task with the categories: zero, one and more than one instance.
We use PyTorch~\cite{Paszke_PyTorch_An_Imperative_2019} in combination with gunpowder~\cite{gunpowder} for training with the following standard random augmentations: Elastic, Intensity, Flipping.
We add the following augmentations: Overlay (overlaying two random image crops to simulate denser images), Permute (randomly permute color channels), Hue (random rotation of the color wheel).
We train the model only on the completely labeled data as the training is not directly applicable to partly labeled data.

The models are trained for 300k iterations with a learning rate of 0.0001 using Adam\cite{adam}, storing weight checkpoints every 10k iterations.
We use a batch size of 2 and train on random crops.
As most images are in large part background, we sample foreground and areas where neurons overlap with higher probability.
The exact ratios depend on the model and are detailed in the respective provided configuration files.
The training code is available here: 
\iftoggle{cvprfinal}{
\href{https://github.com/Kainmueller-Lab/PatchPerPix}{https://github.com/Kainmueller-Lab/PatchPerPix}}{\textit{Hidden during review process}}.

We select the best checkpoint, the best patch threshold and the best threshold for the \textit{numinst} prediction based on the validation set and report both the validation results and the final results on the test set (both combined and separately for the completely labeled and the partly labeled dataset).
We observed that the models tend to overestimate the case of a single neuron in a given region and underestimate background and neuron overlaps.
To counter this, instead of using a simple \(\argmax\), we additionally select an optimal threshold based on the validation results.

PatchPerPix can only handle overlaps up to the size of the patch size.
As its instance assembly step is computationally demanding for 3d data, the currently applicable patch size is restricted.
In order to be able to handle larger overlaps, PatchPerPix needs to be scaled up in future work.

\subsubsection{Flood Filling Networks}\label{suppl:ffn_info}
For Flood Filling Networks (FFN) we mainly follow the proposed architecture from \cite{januszewski2016ffn} and the publicly available code\footnote{original code: \href{https://github.com/google/ffn}{https://github.com/google/ffn}; and adapted for FISBe: \iftoggle{cvprfinal}{
\href{https://github.com/Kainmueller-Lab/ffn}{https://github.com/Kainmueller-Lab/ffn}}{\textit{Hidden during review process}}}.
We use 12 stacked convolution modules with skip connections in between, where each convolution module consists of two 3d convolution layers.
The field of view (FoV) size, which corresponds to the spatial dimensions of the network's input and output size, is 33$\times$33$\times$33 and we adapted the network to work with three input channel.
FFNs move their current FoV by a short distance after each update to be able to trace the entire object. For this, we use the cuboid movement policy described in \cite{januszewski2018ffn} with step size 8 for each dimension.
We apply standard data augmentation by flipping and permuting spatial axis.
We train the models for 2m iterations with batch size 4.
The sampling strategy is the same as in the original work.

During training we use seeds (starting position of the FoV) generated from ground truth. 
For prediction we create the seeds as follows: 
We convert the three channel input to a grayscale volume and threshold it to obtain a foreground mask, where we filter out small connected components.
Finally, we take local maxima on the corresponding distance transform map.
We determine both thresholds (foreground and small connected components size) during validation.
Aside from that, we choose the best checkpoint, the best FoV movement threshold and the best final segmentation threshold based on validation.

We train and test FFNs both on only the completely labeled dataset and on the full dataset.
In contrast to PatchPerPix, FFNs only consider one instance at a time, which means that there are no changes necessary to train FFNs on partly labeled samples.

\subsubsection{Color Clustering}\label{suppl:brainbow_info}
Duan et al.~\cite{duan2021} propose a non-learnt color clustering algorithm based on \cite{sumbul2026} to segment mouse neurons in Brainbow \cite{livet2007brainbow} images.
Brainbow is a stochastic labeling technique to image neurons in unique colors with light microscopy.
This assumption does not hold for our FISBe dataset, where multiple neurons and abundant noise can be of the same color.
Thus, some steps of the pipeline (denoising, supervoxel generation, color clustering, linkage bridging) need to be adapted to fit our dataset.\footnote{code: \iftoggle{cvprfinal}{
\href{https://github.com/Kainmueller-Lab/brainbow}{https://github.com/Kainmueller-Lab/brainbow}}{\textit{Hidden during review process}}}

Following the original work, we denoise our 3d images with bm4d \cite{maggioni2013bm4d}.
We use $\sigma=0.05$ as noise standard deviation, and normalize and denoise each channel separately.
For supervoxel generation, we threshold the denoised image with foreground threshold $t_{fg}=0.08$, apply distance transform, and run watershed transform with local maxima as seeds and the thresholded foreground as mask. All connected components smaller than threshold $t_{rm}=800$ are removed.
In the next step, all supervoxels are clustered with Gaussian Mixture Models (GMM).
We create an adjacency matrix where supervoxel pairs have a value $>0$, if their spatial and color distance is smaller than certain thresholds ($\delta_s = 5$, $\delta_c = 14$).
Moreover, we use the Bayes Information Criterion (BIC) to determine the number of clusters for the GMM clustering.
Finally, as same colored, but not touching neurons are clustered together, we apply connected component analysis for each GMM cluster with a distance threshold ($\Delta_s = 20$).
Please note, that differing from the original works, we omit supervoxel subdivision and merging, PCA as well as linking bridging, because these steps did not improve performance for our dataset.
We determined $\sigma$, $t_{fg}$, $t_{rm}$, $\delta_s$, $\delta_c$ and $\Delta_s$ during validation.

\subsection{Biological Background and Motivation}
\label{suppl:bio_background}
This section gives a brief overview of the advances our dataset will facilitate in the field of basic neuroscience. 
Based on neuron instance segmentations in MCFO images, neurons can be studied threefold:
(1) Clustering and subsequent statistical analysis of neuron morphologies may yield insights into neuronal cell types and their variability \cite{frechter2019,Ehrhardt2023}.
(2) Locating a neuron morphology of interest in multiple MCFO images facilitates the creation of novel transgenic lines that sparsely express the neuron of interest, which in turn facilitates functional analyses of individual neurons of interest in vivo \cite{splitline_pfeiffer2010}.
(3) Information on neuron connectivity and neuron function can be fused by locating neurons segmented from electron microscopy (EM) data in MCFO images and subsequent in vivo studies as in (2) \cite{meissner2023,Wang2020,Morimoto2020}. 
For these tasks, instance segmentations do not necessarily need to cover all true neurons: Given that MCFO images express only a random subset of neurons in the first place, missing some dim neurons in an instance segmentation, while further reducing recall, does not introduce a categorically new source of error. More specifically, segmenting a subset of neurons with high individual clDice score is preferable to segmenting all neurons but only partly.
We acknowledge these application-derived preferences in our selection of metrics (see Sec. 3 of the main paper).


\subsection{Sample Information and Visualization}\label{suppl:sample_infs}

\begin{figure*}[thbp]
    \centering
    \includegraphics[width=\linewidth]{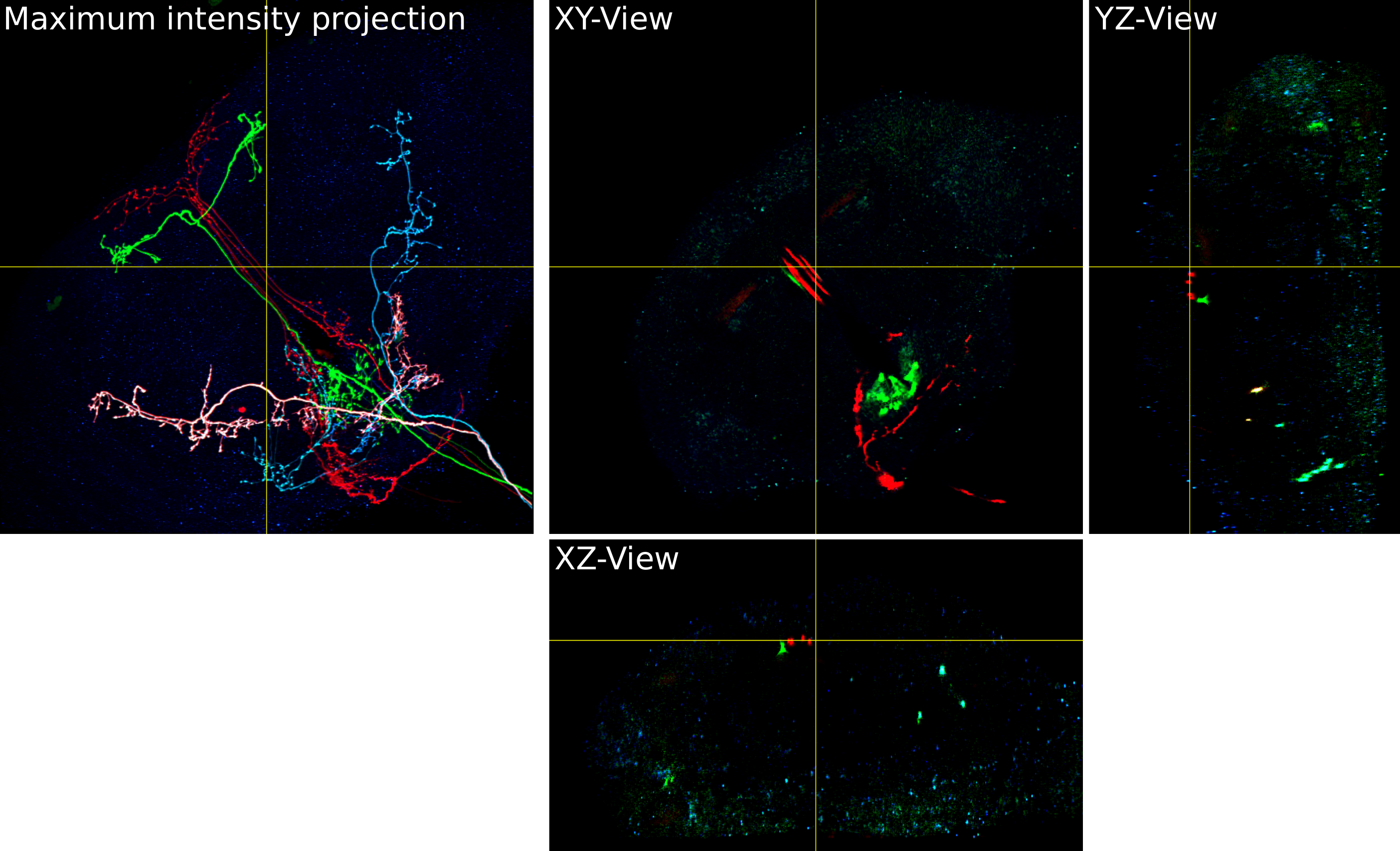}
    \caption{Maximum intensity projection and orthographic views for sample VT007080-20170517\_61\_A2. The orthographic views highlight the overall sparseness of the foreground and thinness of the neuronal structures.
    \label{fig:ortho_view}} 
\end{figure*}

\begin{figure*}
    \centering
    \begin{tabular}{m{0.14\textwidth}m{0.14\textwidth}m{0.14\textwidth}m{0.14\textwidth}m{0.14\textwidth}m{0.14\textwidth}}
        \begin{subfigure}[b]{\textwidth}
        \caption*{Training set}
            \hfil%
			\includegraphics[width=0.16\textwidth]{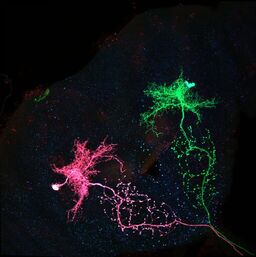}
            \hfil%
			\includegraphics[width=0.16\textwidth]{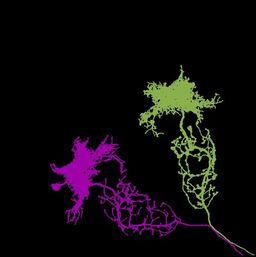}
            \hfil%
            \includegraphics[width=0.16\textwidth]{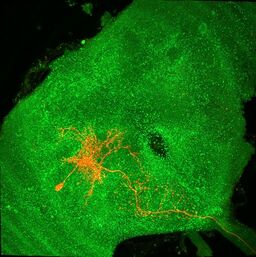}
            \hfil%
			\includegraphics[width=0.16\textwidth]{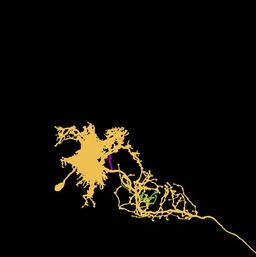}
            \hfil%
            \includegraphics[width=0.16\textwidth]{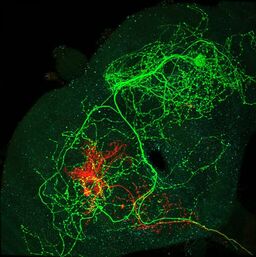}
            \hfil%
			\includegraphics[width=0.16\textwidth]{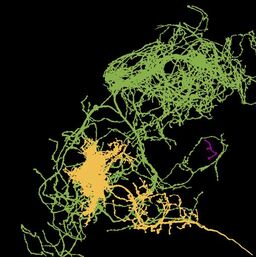}
            \hfil%
        \end{subfigure}\\
        \begin{subfigure}[b]{\textwidth}
            \hfil%
            \includegraphics[width=0.16\textwidth]{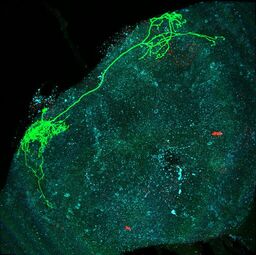}
            \hfil%
            \includegraphics[width=0.16\textwidth]{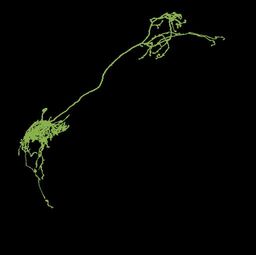}
            \hfil%
            \includegraphics[width=0.16\textwidth]{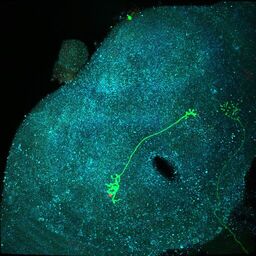}
            \hfil%
            \includegraphics[width=0.16\textwidth]{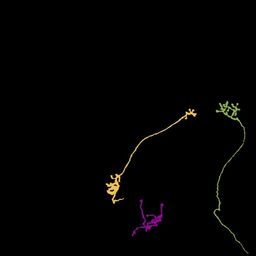}
            \hfil%
            \includegraphics[width=0.16\textwidth]{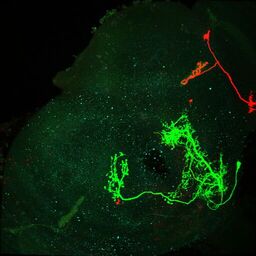}
            \hfil%
            \includegraphics[width=0.16\textwidth]{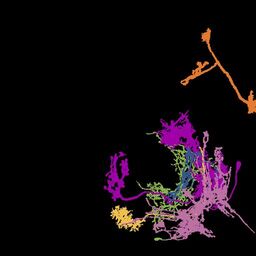}
            \hfil%
        \end{subfigure}\\
        \begin{subfigure}[b]{\textwidth}
            \hfil%
            \includegraphics[width=0.16\textwidth]{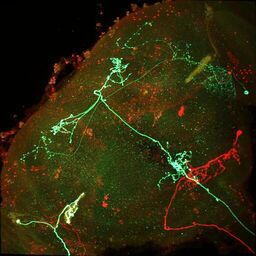}
            \hfil%
            \includegraphics[width=0.16\textwidth]{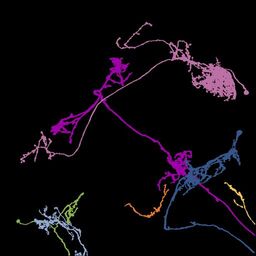}
            \hfil%
            \includegraphics[width=0.16\textwidth]{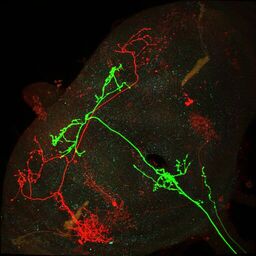}
            \hfil%
            \includegraphics[width=0.16\textwidth]{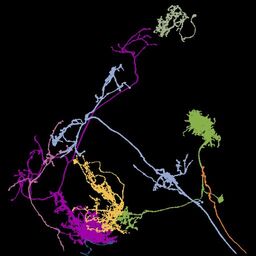}
            \hfil%
            \includegraphics[width=0.16\textwidth]{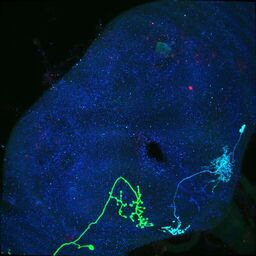}
            \hfil%
            \includegraphics[width=0.16\textwidth]{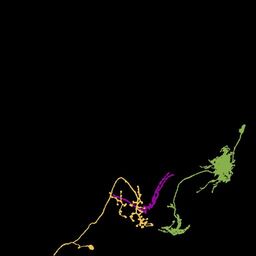}
            \hfil%
        \end{subfigure}\\
        \begin{subfigure}[b]{\textwidth}
            \hfil%
            \includegraphics[width=0.16\textwidth]{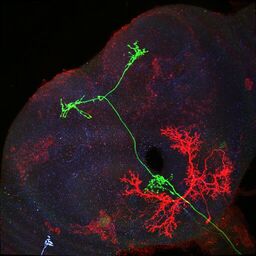}
            \hfil%
            \includegraphics[width=0.16\textwidth]{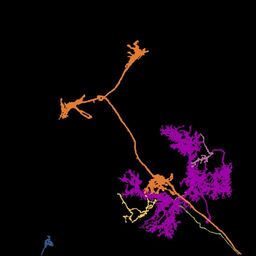}
            \hfil%
            \includegraphics[width=0.16\textwidth]{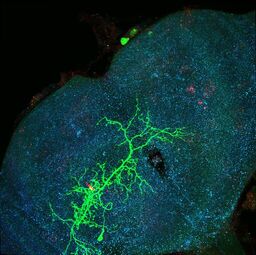}
            \hfil%
            \includegraphics[width=0.16\textwidth]{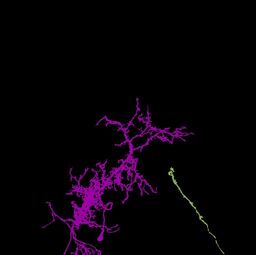}
            \hfil%
            \includegraphics[width=0.16\textwidth]{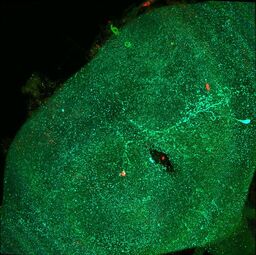}
            \hfil%
            \includegraphics[width=0.16\textwidth]{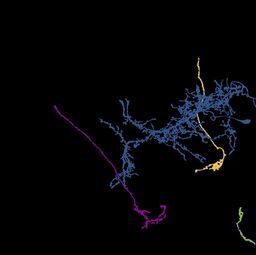}
            \hfil%
        \end{subfigure}\\
        \begin{subfigure}[b]{\textwidth}
            \hfil%
            \includegraphics[width=0.16\textwidth]{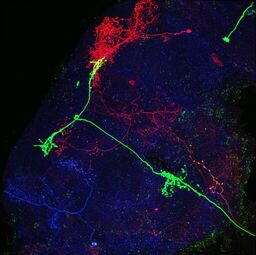}
            \hfil%
            \includegraphics[width=0.16\textwidth]{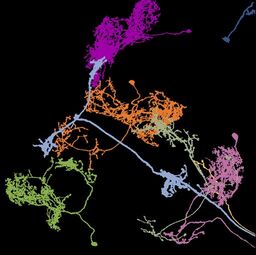}
            \hfil%
            \includegraphics[trim=0 22 0 21, clip, width=0.16\textwidth]{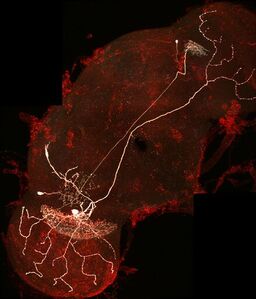}
            \hfil%
            \includegraphics[trim=0 22 0 21, clip, width=0.16\textwidth]{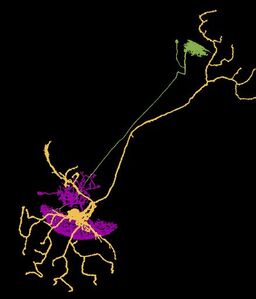}
            \hfil%
            \includegraphics[trim=0 4 0 4, clip, width=0.16\textwidth]{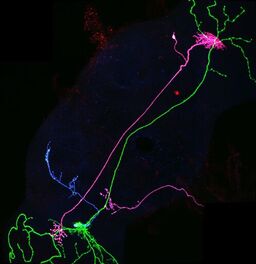}
            \hfil%
            \includegraphics[trim=0 4 0 4, clip, width=0.16\textwidth]{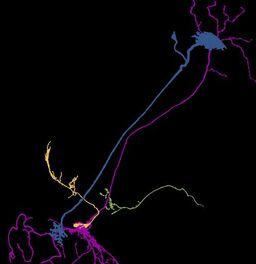}
            \hfil%
        \end{subfigure}\\
        \begin{subfigure}[b]{\textwidth}
            \hfil%
            \includegraphics[width=0.16\textwidth]{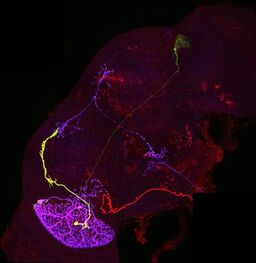}
            \hfil%
            \includegraphics[width=0.16\textwidth]{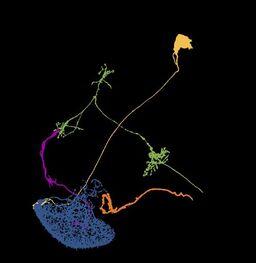}
            \hfil%
            \includegraphics[trim=0 12 0 13, clip, width=0.16\textwidth]{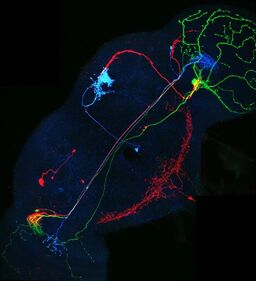}
            \hfil%
            \includegraphics[trim=0 12 0 13, clip, width=0.16\textwidth]{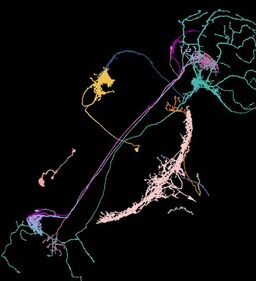}
            \hfil%
            \includegraphics[width=0.16\textwidth]{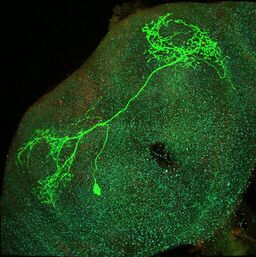}
            \hfil%
            \includegraphics[width=0.16\textwidth]{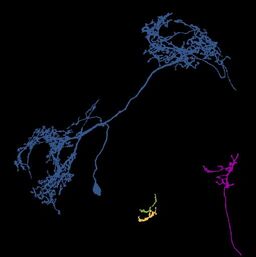}
            \hfil%
        \end{subfigure}\\
		\begin{subfigure}[b]{0.16\textwidth}
			\caption*{}
		\end{subfigure}&
        \begin{subfigure}[b]{0.16\textwidth}
			\caption*{}
		\end{subfigure}&
        \begin{subfigure}[b]{0.16\textwidth}
			\caption*{}
		\end{subfigure}&
		\begin{subfigure}[b]{0.16\textwidth}
			\caption*{}
		\end{subfigure}&
        \begin{subfigure}[b]{0.16\textwidth}
			\caption*{}
		\end{subfigure}&
        \begin{subfigure}[b]{0.16\textwidth}
			\caption*{}
		\end{subfigure}
    \end{tabular}
    \vspace{-5mm}
    \caption{Maximum intensity projections (MIP) of 3d light microscopy samples and ground truth (gt) instance segmentations of all samples in the completely labeled set. MIP and gt are depicted next to each other in alternating order. Images are scaled to same width, some images are center cropped. Figure continued on next page.}
    \label{suppl_fig:samples_completely}
\end{figure*}

\begin{figure*}
    \centering
    \begin{tabular}{m{0.14\textwidth}m{0.14\textwidth}m{0.14\textwidth}m{0.14\textwidth}m{0.14\textwidth}m{0.14\textwidth}}
        \begin{subfigure}[b]{\textwidth}
        \caption*{Training set}
            \hfil%
			\includegraphics[width=0.16\textwidth]{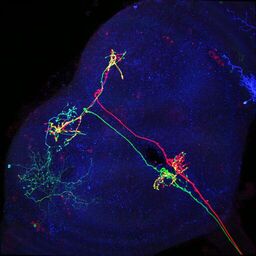}
            \hfil%
			\includegraphics[width=0.16\textwidth]{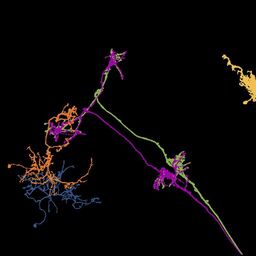}
            \hfil%
            \includegraphics[width=0.16\textwidth]{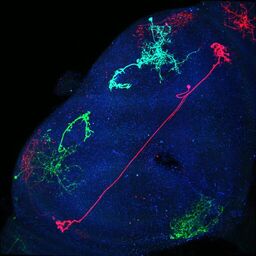}
            \hfil%
			\includegraphics[width=0.16\textwidth]{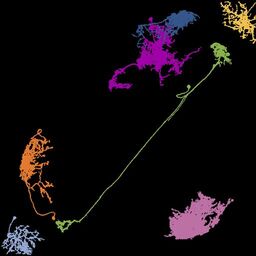}
            \hfil%
            \includegraphics[width=0.16\textwidth]{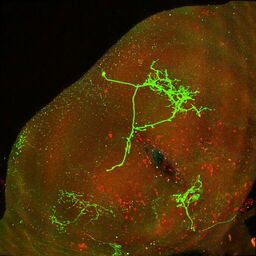}
            \hfil%
			\includegraphics[width=0.16\textwidth]{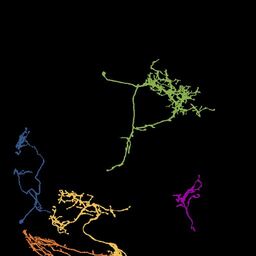}
            \hfil%
        \end{subfigure}\\
        \begin{subfigure}[b]{\textwidth}
            \hfil%
			\includegraphics[width=0.16\textwidth]{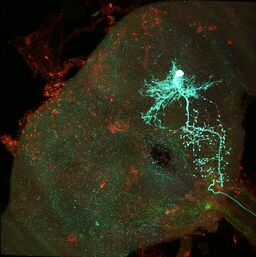}
            \hfil%
			\includegraphics[width=0.16\textwidth]{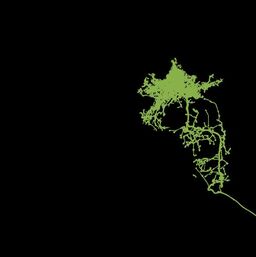}
            \hfil%
            \includegraphics[width=0.16\textwidth]{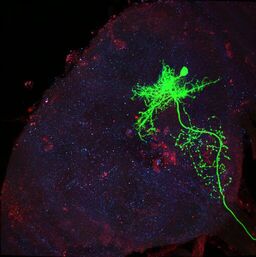}
            \hfil%
			\includegraphics[width=0.16\textwidth]{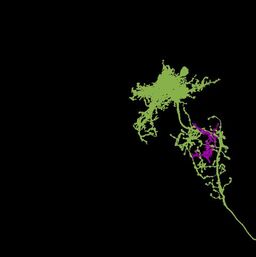}
            \hfil%
            \includegraphics[width=0.16\textwidth]{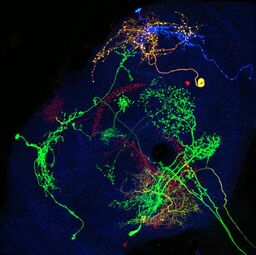}
            \hfil%
			\includegraphics[width=0.16\textwidth]{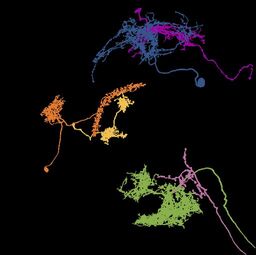}
            \hfil%
          \end{subfigure}\\
        \begin{subfigure}[b]{\textwidth}
            \hfil%
			\includegraphics[width=0.16\textwidth]{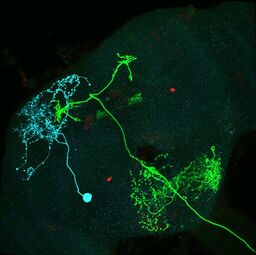}
            \hfil%
			\includegraphics[width=0.16\textwidth]{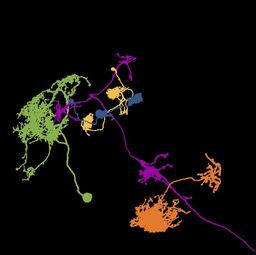}
            \hfil%
            \includegraphics[width=0.16\textwidth]{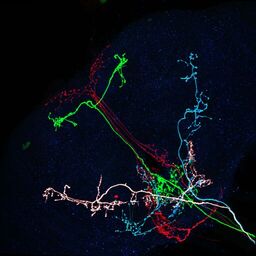}
            \hfil%
			\includegraphics[width=0.16\textwidth]{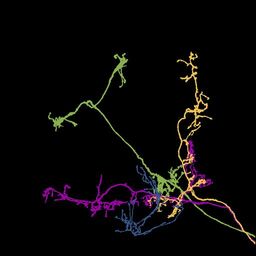}
            \hfil%
            \includegraphics[width=0.16\textwidth]{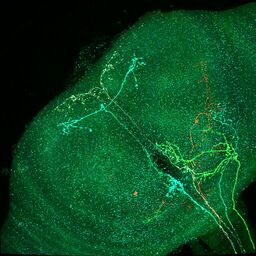}
            \hfil%
			\includegraphics[width=0.16\textwidth]{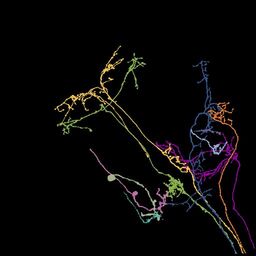}
            \hfil%
        \end{subfigure}\\
        \begin{subfigure}[b]{\textwidth}
            \hfil%
			\includegraphics[width=0.16\textwidth]{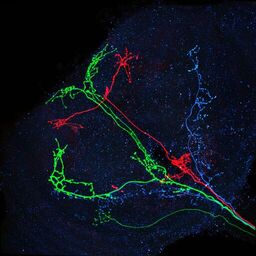}
            \hfil%
			\includegraphics[width=0.16\textwidth]{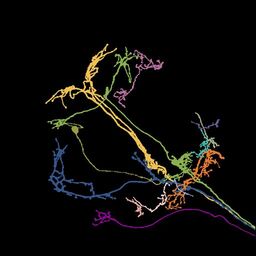}
            \hfil%
            \includegraphics[width=0.16\textwidth]{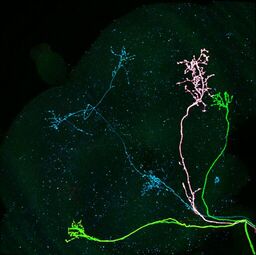}
            \hfil%
			\includegraphics[width=0.16\textwidth]{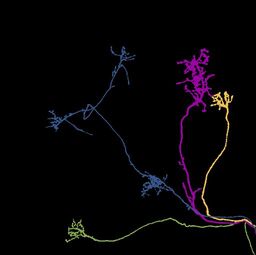}
            \hfil%
            \includegraphics[width=0.16\textwidth]{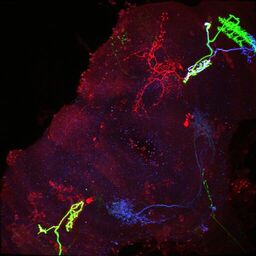}
            \hfil%
			\includegraphics[width=0.16\textwidth]{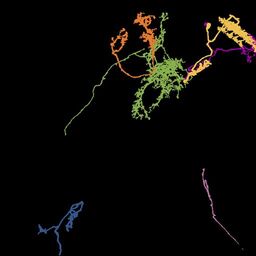}
            \hfil%
          \end{subfigure}\\
        \begin{subfigure}[b]{\textwidth}
            \hfil%
			\includegraphics[width=0.16\textwidth]{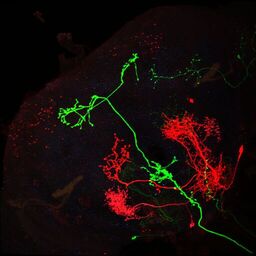}
            \hfil%
			\includegraphics[width=0.16\textwidth]{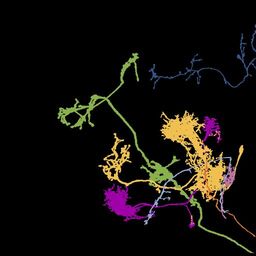}
            \hfil%
            \includegraphics[width=0.16\textwidth]{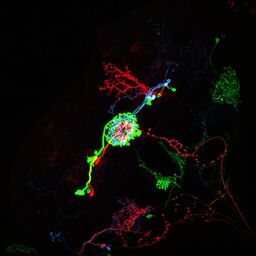}
            \hfil%
			\includegraphics[width=0.16\textwidth]{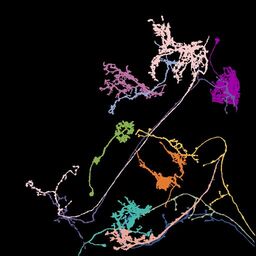}
            \hfil%
            \includegraphics[width=0.16\textwidth]{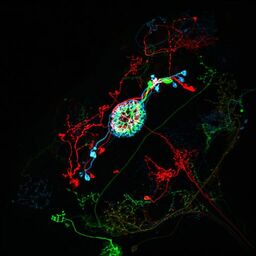}
            \hfil%
			\includegraphics[width=0.16\textwidth]{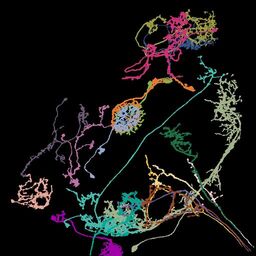}
            \hfil%
        \end{subfigure}\\
        \begin{subfigure}[b]{\textwidth}
            \hfil%
			\includegraphics[width=0.16\textwidth]{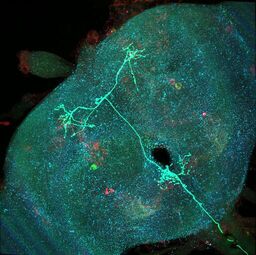}
            \hfil%
			\includegraphics[width=0.16\textwidth]{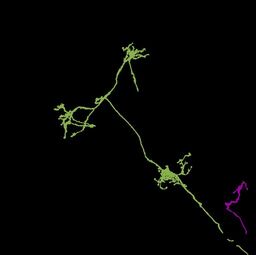}
            \hfil%
            \includegraphics[width=0.16\textwidth]{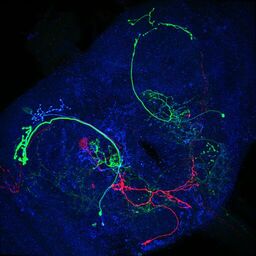}
            \hfil%
			\includegraphics[width=0.16\textwidth]{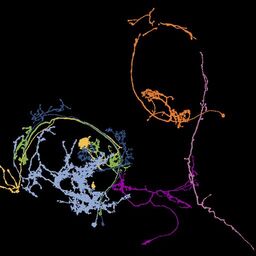}
            \hfil%
            \includegraphics[width=0.16\textwidth]{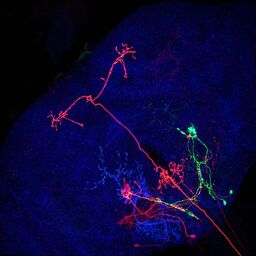}
            \hfil%
			\includegraphics[width=0.16\textwidth]{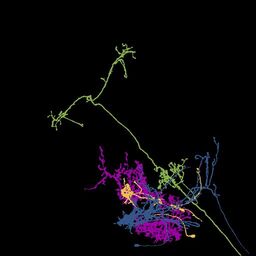}
            \hfil%
          \end{subfigure}\\
		\begin{subfigure}[b]{0.16\textwidth}
			\caption*{}
		\end{subfigure}&
        \begin{subfigure}[b]{0.16\textwidth}
			\caption*{}
		\end{subfigure}&
        \begin{subfigure}[b]{0.16\textwidth}
			\caption*{}
		\end{subfigure}&
		\begin{subfigure}[b]{0.16\textwidth}
			\caption*{}
		\end{subfigure}&
        \begin{subfigure}[b]{0.16\textwidth}
			\caption*{}
		\end{subfigure}&
        \begin{subfigure}[b]{0.16\textwidth}
			\caption*{}
		\end{subfigure}
    \end{tabular}
    \vspace{-5mm}
    \caption{\label{suppl_fig:samples_partly}Maximum intensity projections (MIP) of 3d light microscopy samples and ground truth (gt) instance segmentations of all samples in the partly labeled set. MIP and gt are depicted next to each other in alternating order. Images are scaled to same width, some images are center cropped. Figure continued on next page.}
\end{figure*}

Table~\ref{tab:samples_completely_labeled} provides a list of the included MCFO acquisitions in the completely labeled FISBe dataset including information on the split (train/val/test) in which each sample was used.
Table~\ref{tab:samples_partly_labeled} provides a list of the included MCFO acquisitions in the partly labeled FISBe dataset including information on the split (train/val/test) in which each sample was used.
Fig.~\ref{fig:ortho_view} shows orthographic view for an exemplary sample. It highlights the thin structures and overall sparseness of foreground.
Fig.~\ref{suppl_fig:samples_completely} shows maximum intensity projections together with the gt instance segmentation of all samples in the completely labeled set, separated by train/val/test.
Fig.~\ref{suppl_fig:samples_partly} shows maximum intensity projections together with the gt instance segmentation of all samples in the partly labeled set, separated by train/val/test.

\begin{table*}[bpht]
	\centering
	\caption{\label{tab:samples_completely_labeled}List of MCFO acquisitions in completely labeled FlyLight Instance Segmentation dataset with sample name ($<$\textit{GAL4 line}$>$-$<$\textit{slide code}$>$), number of annotated neurons, density category and split.}
	\begin{tabu} to 1\linewidth{ X[4.2l] X[0.8c] X[0.6c] X[0.8c] | X[4.2l] X[0.8c] X[0.6c] X[0.8c]  }
        \midrule
		  Sample name & Neurons & Cat. & Split & Sample name & Neurons & Cat. & Split\\
        \midrule
R38F04-20181005\_63\_G3       & 2                     & 2                     & train         & VT047848-20171020\_66\_J2     & 5                     & 2                     & train\\
R38F04-20181005\_63\_G5       & 3                     & 2                     & train         & VT047848-20171020\_66\_I5     & 12                    & 2                     & train\\
R38F04-20181005\_63\_H1       & 3                     & 2                     & train         & VT061467-20180911\_62\_E5     & 4                     & 2                     & train\\
R53A10-20181019\_64\_A4       & 1                     & 2                     & train         & R22C03-20180918\_66\_J2       & 2                     & 2                     & val\\
R75E01-20181030\_64\_D1       & 3                     & 2                     & train         & VT012403-20171128\_61\_B2     & 5                     & 2                     & val\\
VT008647-20171222\_63\_D2     & 6                     & 3                     & train         & VT033614-20171124\_64\_H5     & 3                     & 3                     & val\\
VT008647-20171222\_63\_D1     & 7                     & 3                     & train         & VT033614-20171124\_64\_H1     & 4                     & 3                     & val\\
VT008647-20171222\_63\_E1     & 8                     & 3                     & train         & VT041298-20171114\_63\_C3     & 7                     & 2                     & val\\
VT019303-20171013\_65\_B6     & 3                     & 2                     & train         & JRC\_SS04989-20160318\_24\_A2 & 3                     & 2                     & test\\
VT019307-20171013\_65\_F1     & 6                     & 3                     & train         & R14A02-20180905\_65\_A6       & 7                     & 3                     & test\\
VT033051-20171128\_61\_E4     & 2                     & 2                     & train         & R54A09-20181019\_64\_H1       & 1                     & 2                     & test\\
VT033051-20171128\_61\_E2     & 4                     & 2                     & train         & VT011145-20171222\_63\_I1     & 9                     & 3                     & test\\
VT040433-20170919\_63\_D6     & 8                     & 2                     & train         & VT027175-20171031\_62\_H3     & 3                     & 2                     & test\\
VT047848-20171020\_66\_I3     & 3                     & 2                     & train         & VT027175-20171031\_62\_H4     & 6                     & 2                     & test\\
VT047848-20171020\_66\_I2     & 4                     & 2                     & train         & VT050157-20171110\_61\_C1     & 5                     & 2                     & test\\
\bottomrule
	\end{tabu}
\end{table*}

\begin{table*}[bpht]
	\centering
	\caption{\label{tab:samples_partly_labeled}List of MCFO acquisitions in partly labeled FlyLight Instance Segmentation dataset with sample name ($<$\textit{GAL4 line}$>$-$<$\textit{slide code}$>$), number of annotated neurons, density category and split.}
	\begin{tabu} to 1\linewidth{ X[4.2l] X[0.8c] X[0.6c] X[0.8c] | X[4.2l] X[0.8c] X[0.6c] X[0.8c]  }
        \midrule
		  Sample name & Neurons & Cat. & Split & Sample name & Neurons & Cat. & Split\\
        \midrule
	R14B11-20180905\_65\_D2       & 5                   & 2                 & train         & VT050217-20171110\_61\_D6     & 5                   & 2                 & train\\
	R14B11-20180905\_65\_D6       & 9                   & 2                 & train         & VT050217-20171110\_61\_E1     & 6                   & 2                 & train\\
	R24D12-20180921\_65\_J6       & 5                   & 3                 & train         & VT058568-20170926\_64\_E1     & 13                  & 3                 & train\\
	R38F04-20181005\_63\_G2       & 1                   & 2                 & train         & VT060731-20170517\_63\_F1     & 6                   & 2                 & train\\
	R38F04-20181005\_63\_G4       & 2                   & 2                 & train         & VT060731-20170517\_63\_F2     & 7                   & 2                 & train\\
	VT003236-20170602\_62\_G4     & 6                   & 3                 & train         & VT061467-20180911\_62\_E4     & 1                   & 2                 & train\\
	VT003236-20170602\_62\_G5     & 6                   & 3                 & train         & VT062059-20170727\_61\_D4     & 6                   & 2                 & train\\
	VT007080-20170517\_61\_A2     & 4                   & 2                 & train         & JRC\_SS05008-20160318\_24\_B1 & 4                   &                   & val\\
	VT007080-20170517\_61\_A4     & 10                  & 2                 & train         & JRC\_SS05008-20160318\_24\_B2 & 6                   &                   & val\\
	VT007080-20170517\_61\_A5     & 15                  & 2                 & train         & R22C03-20180918\_66\_J1       & 3                   & 2                 & val\\
	VT008135-20171122\_61\_C2     & 4                   & 2                 & train         & R9F03-20181030\_62\_B5        & 3                   & 2                 & val\\
	VT008647-20171222\_63\_D5     & 6                   & 3                 & train         & VT008194-20171222\_63\_A3     & 13                  & 2                 & val\\
	VT008647-20171222\_63\_D6     & 7                   & 3                 & train         & VT008194-20171222\_63\_A5     & 17                  & 2                 & val\\
	VT010264-20171222\_63\_H2     & 12                  & 3                 & train         & VT012403-20171128\_61\_B1     & 6                   & 2                 & val\\
	VT010264-20171222\_63\_H5     & 19                  & 3                 & train         & VT033614-20171124\_64\_H4     & 2                   & 3                 & val\\
	VT011049-20180918\_66\_I1     & 2                   & 1                 & train         & VT039350-20171020\_64\_A1     & 11                  & 3                 & val \\
	VT024641-20170615\_62\_D2     & 7                   & 2                 & train         & VT039350-20171020\_64\_A3     & 8                   & 3                 & val\\
	VT024641-20170615\_62\_D3     & 4                   & 2                 & train         & VT039350-20171020\_64\_A6     & 5                   & 3                 & val\\
	VT024641-20170615\_62\_D5     & 5                   & 2                 & train         & VT059775-20170630\_63\_D5     & 7                   & 2                 & val\\
	VT024641-20170615\_62\_D6     & 10                  & 2                 & train         & R54A09-20181019\_64\_H4       & 4                   & 2                 & test\\
	VT024641-20170615\_62\_E1     & 4                   & 2                 & train         & R54A09-20181019\_64\_H6       & 1                   & 2                 & test\\
	VT025523-20170915\_64\_I1     & 11                  & 2                 & train         & R73H08-20181030\_62\_G5       & 2                   & 2                 & test\\
	VT026776-20171017\_62\_J1     & 13                  & 3                 & train         & VT006202-20170511\_63\_C4     & 8                   & 2                 & test\\
	VT033051-20171128\_61\_E3     & 1                   & 2                 & train         & VT011145-20171222\_63\_I2     & 8                   & 3                 & test\\
	VT033296-20171010\_62\_B4     & 4                   & 2                 & train         & VT021537-20171003\_61\_C3     & 5                   & 3                 & test\\
	VT034391-20171128\_61\_G2     & 2                   & 2                 & train         & VT023747-20171017\_61\_F1     & 10                  & 2                 & test\\
	VT038149-20171103\_62\_F1     & 6                   & 3                 & train         & VT027175-20171031\_62\_H6     & 2                   & 2                 & test\\
	VT039484-20171020\_64\_C1     & 7                   & 3                 & train         & VT028606-20170721\_65\_A2     & 14                  & 3                 & test\\
	VT039484-20171020\_64\_C2     & 12                  & 3                 & train         & VT028606-20170721\_65\_A3     & 12                  & 3                 & test\\
	VT040430-20170919\_63\_C4     & 3                   & 2                 & train         & VT033453-20170721\_65\_D2     & 7                   & 2                 & test\\
	VT040433-20170919\_63\_E1     & 6                   & 2                 & train         & VT033453-20170721\_65\_D4     & 7                   & 2                 & test\\
	VT045568-20171020\_66\_C5     & 4                   & 2                 & train         & VT033453-20170721\_65\_D5     & 5                   & 2                 & test\\
	VT045568-20171020\_66\_D2     & 3                   & 2                 & train         & VT046838-20170922\_62\_A2     & 8                   & 2                 & test\\
	VT047848-20171020\_66\_I1     & 6                   & 2                 & train         & VT050157-20171110\_61\_C5     & 3                   & 2                 & test\\
	VT047848-20171020\_66\_I4     & 8                   & 2                 & train         & VT058571-20170926\_64\_G6     & 5                   & 2                 & test\\
	VT047848-20171020\_66\_J1     & 8                   & 2                 & train         &   & & & \\
\bottomrule
	\end{tabu}
\end{table*}

\begin{figure*}
    \centering
    \begin{tabular}{m{0.14\textwidth}m{0.14\textwidth}m{0.14\textwidth}m{0.14\textwidth}m{0.14\textwidth}m{0.14\textwidth}}
        \begin{subfigure}[b]{\textwidth}
            \caption*{Validation set}
            \hfil%
            \includegraphics[width=0.16\textwidth]{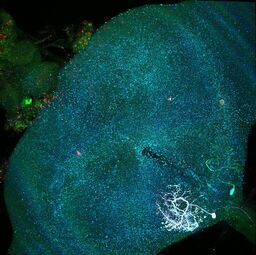}
            \hfil%
            \includegraphics[width=0.16\textwidth]{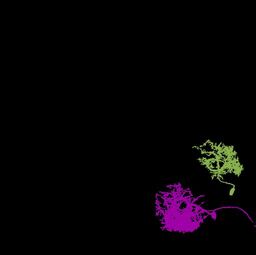}
            \hfil%
            \includegraphics[width=0.16\textwidth]{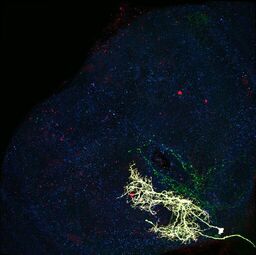}
            \hfil%
            \includegraphics[width=0.16\textwidth]{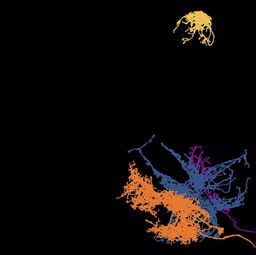}
            \hfil%
            \includegraphics[width=0.16\textwidth]{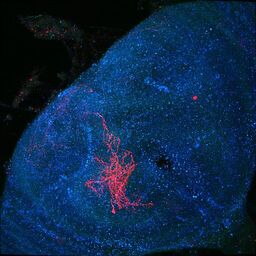}
            \hfil%
            \includegraphics[width=0.16\textwidth]{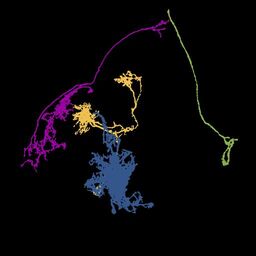}
            \hfil%
        \end{subfigure}\\
      \begin{subfigure}[b]{\textwidth}
          \hfil%
          \includegraphics[width=0.16\textwidth]{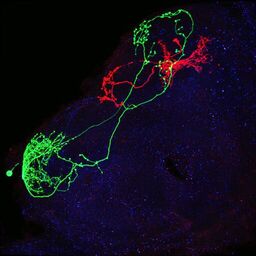}
          \hfil%
          \includegraphics[width=0.16\textwidth]{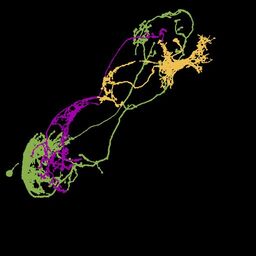}
          \hfil%
          \includegraphics[width=0.16\textwidth]{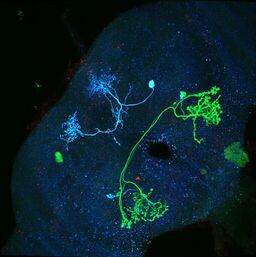}
          \hfil%
          \includegraphics[width=0.16\textwidth]{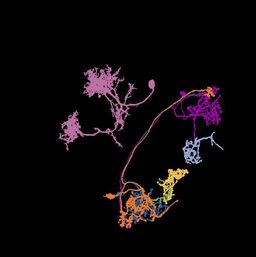}
          \hfil%
          \includegraphics[width=0.16\textwidth]{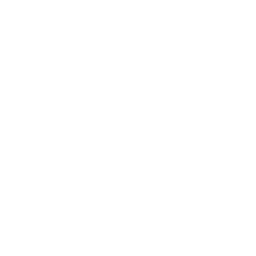}
          \hfil%
          \includegraphics[width=0.16\textwidth]{figures/all_samples/placeholder.jpg}
          \hfil%
        \end{subfigure}\\
        \begin{subfigure}[b]{\textwidth}
        \caption*{Test set}
            \hfil
            \includegraphics[width=0.16\textwidth]{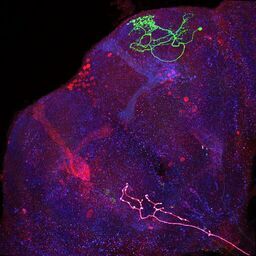}
            \hfil
            \includegraphics[width=0.16\textwidth]{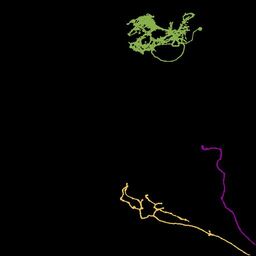}
            \hfil
            \includegraphics[width=0.16\textwidth]{figures/all_samples/completely/test/raw/R14A02-20180905_65_A6.jpg}
            \hfil
            \includegraphics[width=0.16\textwidth]{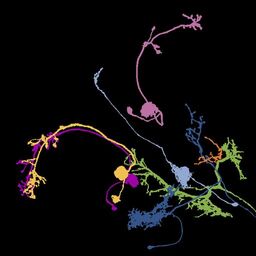}
            \hfil
            \includegraphics[width=0.16\textwidth]{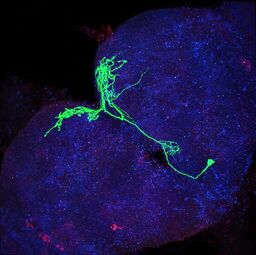}
            \hfil
            \includegraphics[width=0.16\textwidth]{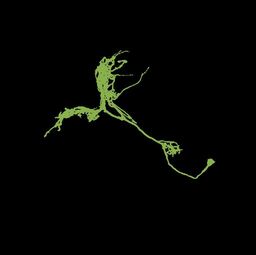}
            \hfil
        \end{subfigure}\\
        \begin{subfigure}[b]{\textwidth}
            \hfil
            \includegraphics[width=0.16\textwidth]{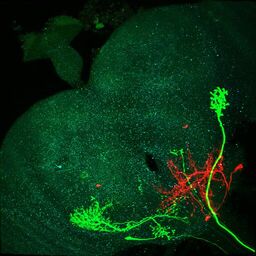}
            \hfil
            \includegraphics[width=0.16\textwidth]{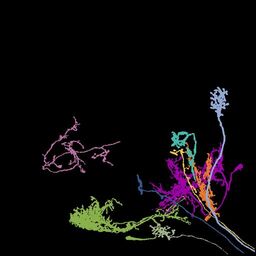}
            \hfil
            \includegraphics[width=0.16\textwidth]{figures/all_samples/completely/test/raw/VT027175-20171031_62_H3.jpg}
            \hfil
            \includegraphics[width=0.16\textwidth]{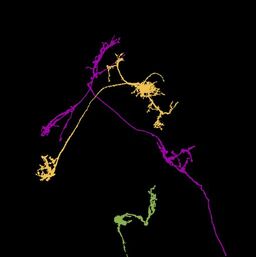}
            \hfil
            \includegraphics[width=0.16\textwidth]{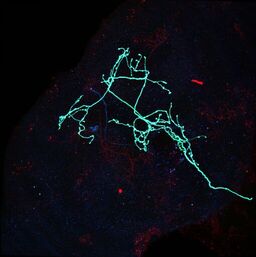}
            \hfil
            \includegraphics[width=0.16\textwidth]{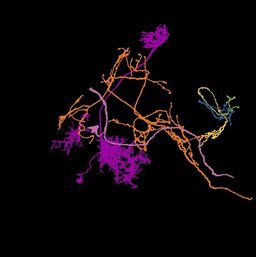}
            \hfil
        \end{subfigure}\\
      \begin{subfigure}[b]{\textwidth}
        \hfil%
        \includegraphics[width=0.16\textwidth]{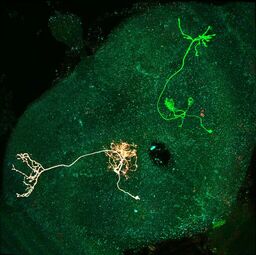}
        \hfil%
        \includegraphics[width=0.16\textwidth]{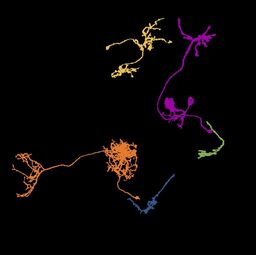}
        \hfil%
        \includegraphics[width=0.16\textwidth]{figures/all_samples/placeholder.jpg}
        \hfil%
        \includegraphics[width=0.16\textwidth]{figures/all_samples/placeholder.jpg}
        \hfil%
        \includegraphics[width=0.16\textwidth]{figures/all_samples/placeholder.jpg}
        \hfil%
        \includegraphics[width=0.16\textwidth]{figures/all_samples/placeholder.jpg}
        \hfil%
        \end{subfigure}\\
		\begin{subfigure}[b]{0.16\textwidth}
			\caption*{}
		\end{subfigure}&
        \begin{subfigure}[b]{0.16\textwidth}
			\caption*{}
		\end{subfigure}&
        \begin{subfigure}[b]{0.16\textwidth}
			\caption*{}
		\end{subfigure}&
		\begin{subfigure}[b]{0.16\textwidth}
			\caption*{}
		\end{subfigure}&
        \begin{subfigure}[b]{0.16\textwidth}
			\caption*{}
		\end{subfigure}&
        \begin{subfigure}[b]{0.16\textwidth}
			\caption*{}
		\end{subfigure}
    \end{tabular}
    \vspace{-5mm}
    \caption{Maximum intensity projections (MIP) of 3d light microscopy samples and ground truth (gt) instance segmentations of all samples in the completely labeled set. MIP and gt are depicted next to each other in alternating order. Images are scaled to same width, some images are center cropped. Figure continued from previous page.}
\end{figure*}

\begin{figure*}
    \centering
    \begin{tabular}{m{0.14\textwidth}m{0.14\textwidth}m{0.14\textwidth}m{0.14\textwidth}m{0.14\textwidth}m{0.14\textwidth}}
      \begin{subfigure}[b]{\textwidth}
        \caption*{Training set (continued from previous page)}
            \hfil%
			\includegraphics[width=0.16\textwidth]{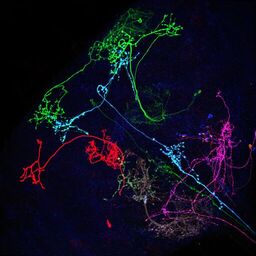}
            \hfil%
			\includegraphics[width=0.16\textwidth]{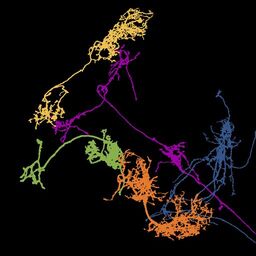}
            \hfil%
            \includegraphics[width=0.16\textwidth]{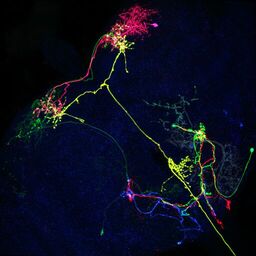}
            \hfil%
			\includegraphics[width=0.16\textwidth]{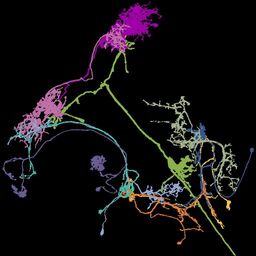}
            \hfil%
            \includegraphics[width=0.16\textwidth]{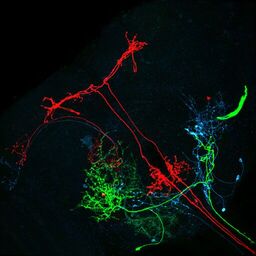}
            \hfil%
			\includegraphics[width=0.16\textwidth]{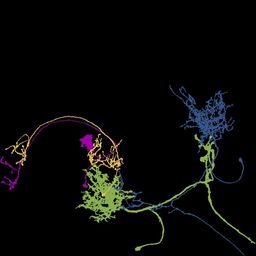}
            \hfil%
        \end{subfigure}\\
        \begin{subfigure}[b]{\textwidth}
            \hfil%
			\includegraphics[width=0.16\textwidth]{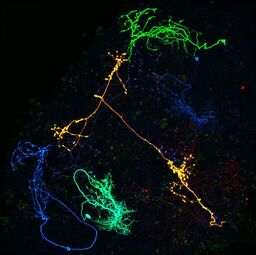}
            \hfil%
			\includegraphics[width=0.16\textwidth]{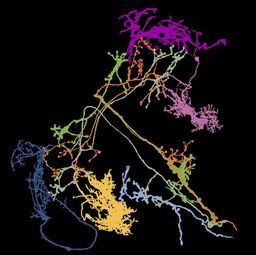}
            \hfil%
            \includegraphics[width=0.16\textwidth]{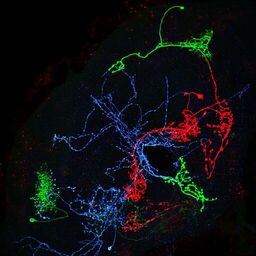}
            \hfil%
			\includegraphics[width=0.16\textwidth]{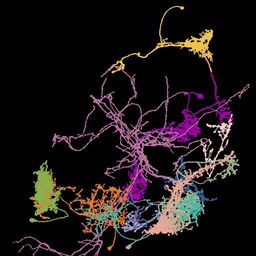}
            \hfil%
            \includegraphics[width=0.16\textwidth]{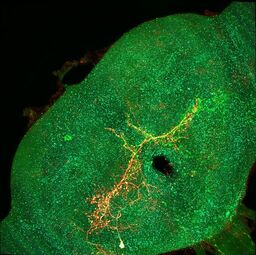}
            \hfil%
			\includegraphics[width=0.16\textwidth]{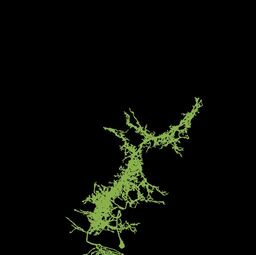}
            \hfil%
        \end{subfigure}\\
        \begin{subfigure}[b]{\textwidth}
            \hfil%
			\includegraphics[width=0.16\textwidth]{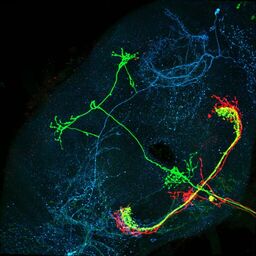}
            \hfil%
			\includegraphics[width=0.16\textwidth]{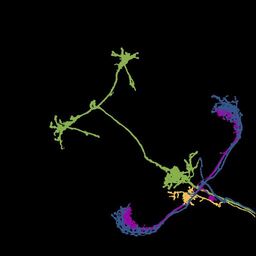}
            \hfil%
            \includegraphics[width=0.16\textwidth]{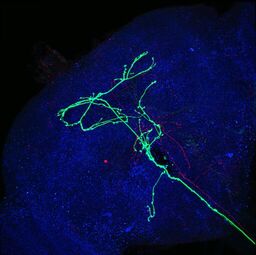}
            \hfil%
			\includegraphics[width=0.16\textwidth]{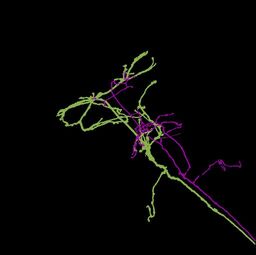}
            \hfil%
            \includegraphics[width=0.16\textwidth]{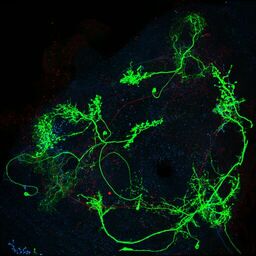}
            \hfil%
			\includegraphics[width=0.16\textwidth]{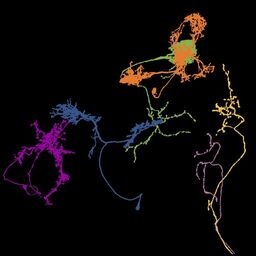}
            \hfil%
        \end{subfigure}\\
        \begin{subfigure}[b]{\textwidth}
            \hfil%
			\includegraphics[width=0.16\textwidth]{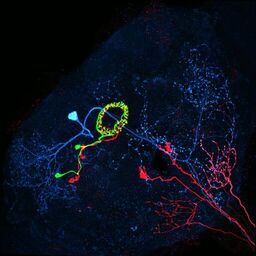}
            \hfil%
			\includegraphics[width=0.16\textwidth]{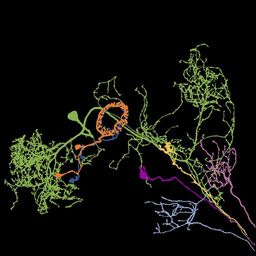}
            \hfil%
            \includegraphics[width=0.16\textwidth]{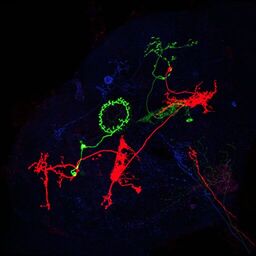}
            \hfil%
			\includegraphics[width=0.16\textwidth]{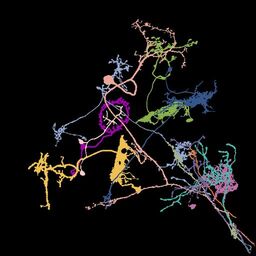}
            \hfil%
            \includegraphics[width=0.16\textwidth]{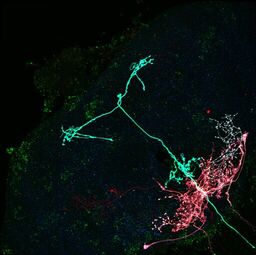}
            \hfil%
			\includegraphics[width=0.16\textwidth]{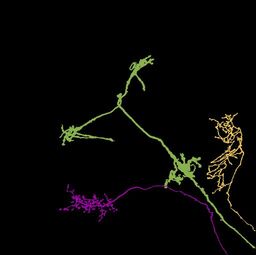}
            \hfil%
        \end{subfigure}\\
        \begin{subfigure}[b]{\textwidth}
            \hfil%
			\includegraphics[width=0.16\textwidth]{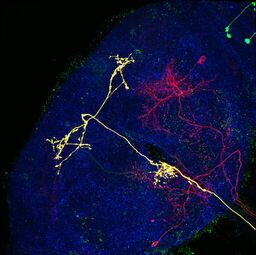}
            \hfil%
			\includegraphics[width=0.16\textwidth]{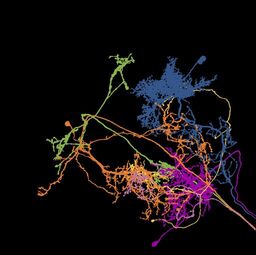}
            \hfil%
            \includegraphics[width=0.16\textwidth]{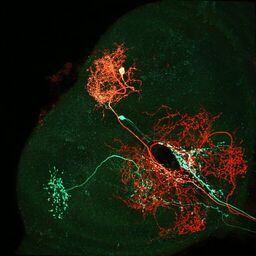}
            \hfil%
			\includegraphics[width=0.16\textwidth]{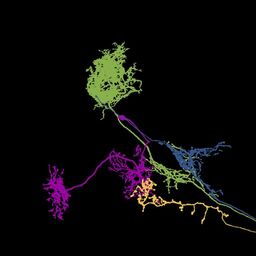}
            \hfil%
            \includegraphics[width=0.16\textwidth]{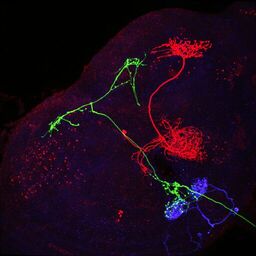}
            \hfil%
			\includegraphics[width=0.16\textwidth]{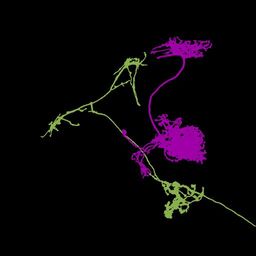}
            \hfil%
          \end{subfigure}\\
        \begin{subfigure}[b]{\textwidth}
            \hfil%
			\includegraphics[trim=6 0 6 0, clip, width=0.16\textwidth]{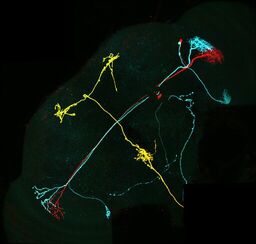}
            \hfil%
			\includegraphics[trim=6 0 6 0, clip, width=0.16\textwidth]{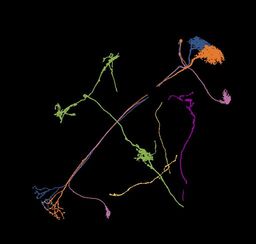}
            \hfil%
            \includegraphics[trim=0 25 0 25, clip, width=0.16\textwidth]{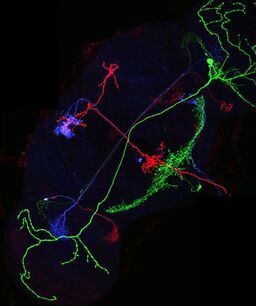}
            \hfil%
            \includegraphics[trim=0 25 0 25, clip, width=0.16\textwidth]{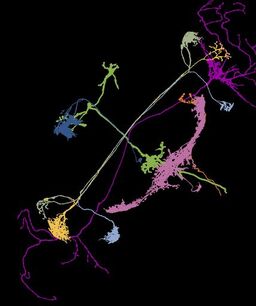}
            \hfil%
            \includegraphics[trim=0 6 0 5, clip, width=0.16\textwidth]{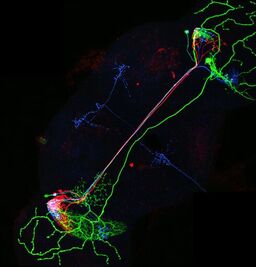}
            \hfil%
			\includegraphics[trim=0 6 0 5, clip, width=0.16\textwidth]{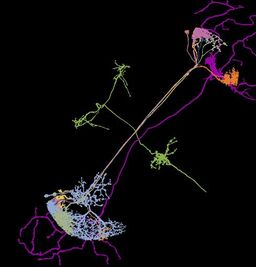}
            \hfil%
        \end{subfigure}\\
        \begin{subfigure}[b]{\textwidth}
            \hfil%
			\includegraphics[width=0.16\textwidth]{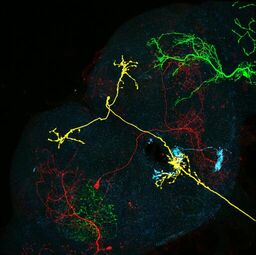}
            \hfil%
			\includegraphics[width=0.16\textwidth]{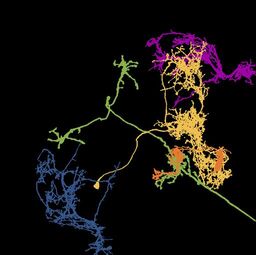}
            \hfil%
            \includegraphics[width=0.16\textwidth]{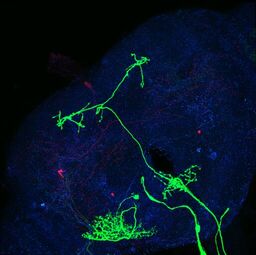}
            \hfil%
			\includegraphics[width=0.16\textwidth]{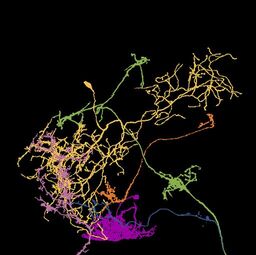}
            \hfil%
            \includegraphics[width=0.16\textwidth]{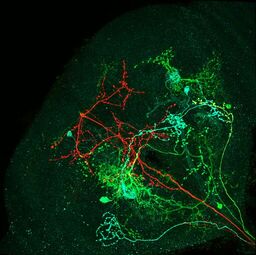}
            \hfil%
			\includegraphics[width=0.16\textwidth]{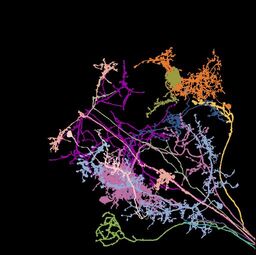}
            \hfil%
        \end{subfigure}\\
		\begin{subfigure}[b]{0.16\textwidth}
			\caption*{}
		\end{subfigure}&
        \begin{subfigure}[b]{0.16\textwidth}
			\caption*{}
		\end{subfigure}&
        \begin{subfigure}[b]{0.16\textwidth}
			\caption*{}
		\end{subfigure}&
		\begin{subfigure}[b]{0.16\textwidth}
			\caption*{}
		\end{subfigure}&
        \begin{subfigure}[b]{0.16\textwidth}
			\caption*{}
		\end{subfigure}&
        \begin{subfigure}[b]{0.16\textwidth}
			\caption*{}
		\end{subfigure}
    \end{tabular}
    \vspace{-5mm}
    \caption{Maximum intensity projections (MIP) of 3d light microscopy samples and ground truth (gt) instance segmentations of all samples in the partly labeled set. MIP and gt are depicted next to each other in alternating order. Images are scaled to same width, some images are center cropped. Figure continued on next page.}
\end{figure*}
\begin{figure*}
    \centering
    \begin{tabular}{m{0.14\textwidth}m{0.14\textwidth}m{0.14\textwidth}m{0.14\textwidth}m{0.14\textwidth}m{0.14\textwidth}}
      \begin{subfigure}[b]{\textwidth}
        \caption*{Training set (continued from previous page)}
            \hfil%
			\includegraphics[width=0.16\textwidth]{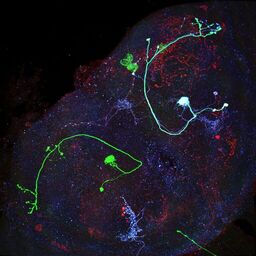}
            \hfil%
			\includegraphics[width=0.16\textwidth]{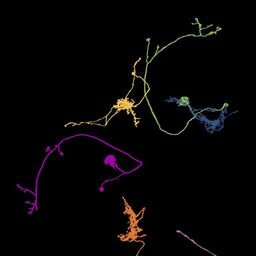}
            \hfil%
            \includegraphics[width=0.16\textwidth]{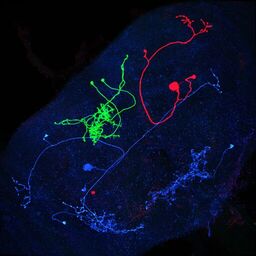}
            \hfil%
			\includegraphics[width=0.16\textwidth]{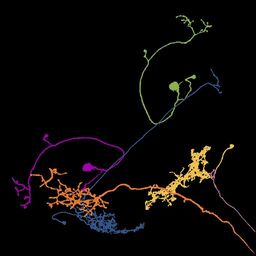}
            \hfil%
            \includegraphics[width=0.16\textwidth]{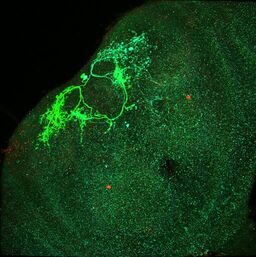}
            \hfil%
			\includegraphics[width=0.16\textwidth]{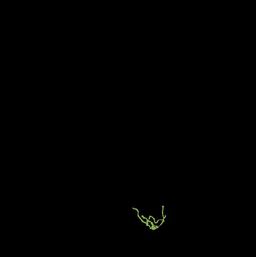}
            \hfil%
        \end{subfigure}\\
      \begin{subfigure}[b]{\textwidth}
        \hfil%
        \includegraphics[width=0.16\textwidth]{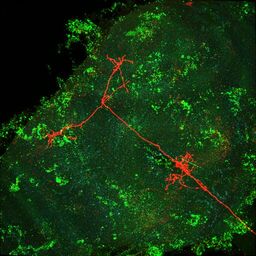}
        \hfil%
        \includegraphics[width=0.16\textwidth]{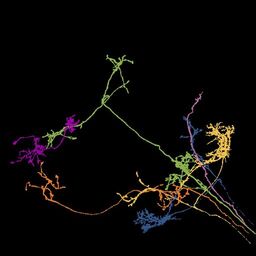}
        \hfil%
        \includegraphics[width=0.16\textwidth]{figures/all_samples/placeholder.jpg}
        \hfil%
        \includegraphics[width=0.16\textwidth]{figures/all_samples/placeholder.jpg}
        \hfil%
        \includegraphics[width=0.16\textwidth]{figures/all_samples/placeholder.jpg}
        \hfil%
        \includegraphics[width=0.16\textwidth]{figures/all_samples/placeholder.jpg}
        \hfil%
        \end{subfigure}\\
        \begin{subfigure}[b]{\textwidth}
        \caption*{Validation set}
            \hfil%
			\includegraphics[width=0.16\textwidth]{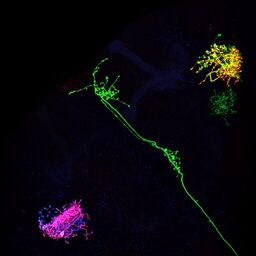}
            \hfil%
			\includegraphics[width=0.16\textwidth]{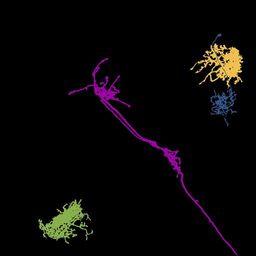}
            \hfil%
            \includegraphics[width=0.16\textwidth]{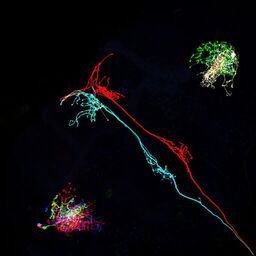}
            \hfil%
			\includegraphics[width=0.16\textwidth]{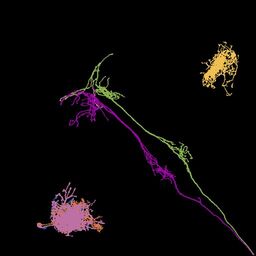}
            \hfil%
            \includegraphics[width=0.16\textwidth]{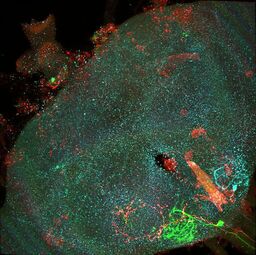}
            \hfil%
			\includegraphics[width=0.16\textwidth]{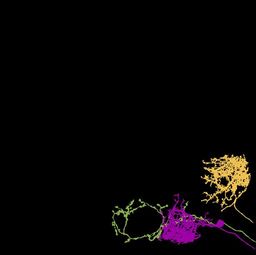}
            \hfil%
        \end{subfigure}\\
        \begin{subfigure}[b]{\textwidth}
            \hfil%
			\includegraphics[width=0.16\textwidth]{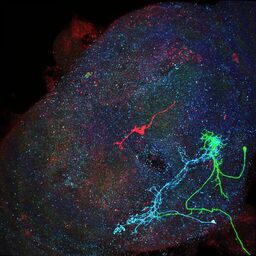}
            \hfil%
			\includegraphics[width=0.16\textwidth]{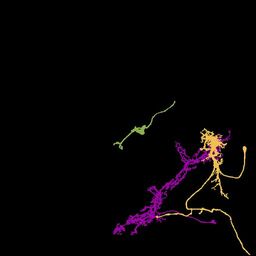}
            \hfil%
            \includegraphics[width=0.16\textwidth]{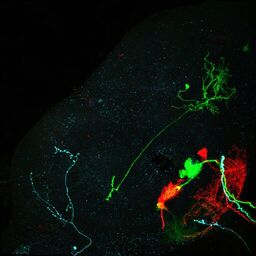}
            \hfil%
			\includegraphics[width=0.16\textwidth]{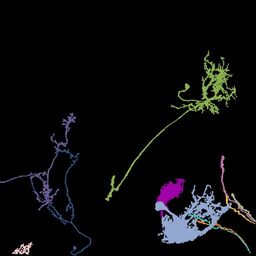}
            \hfil%
            \includegraphics[width=0.16\textwidth]{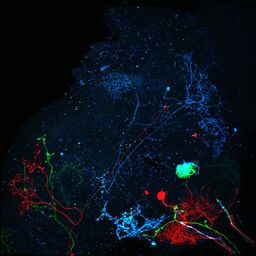}
            \hfil%
			\includegraphics[width=0.16\textwidth]{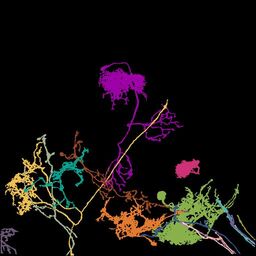}
            \hfil%
        \end{subfigure}\\
        \begin{subfigure}[b]{\textwidth}
            \hfil%
			\includegraphics[width=0.16\textwidth]{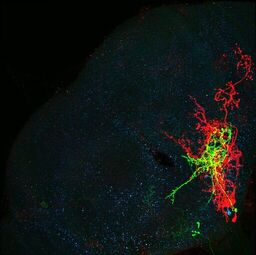}
            \hfil%
			\includegraphics[width=0.16\textwidth]{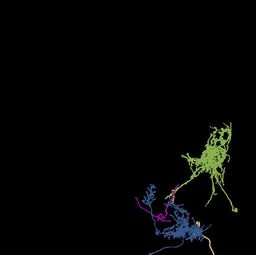}
            \hfil%
            \includegraphics[width=0.16\textwidth]{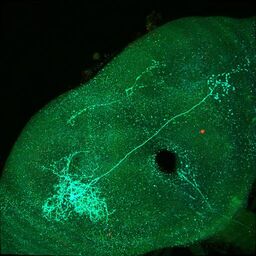}
            \hfil%
			\includegraphics[width=0.16\textwidth]{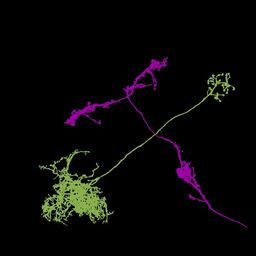}
            \hfil%
            \includegraphics[width=0.16\textwidth]{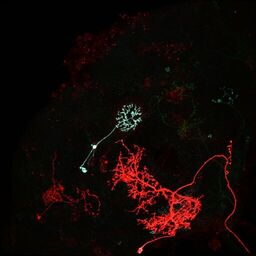}
            \hfil%
			\includegraphics[width=0.16\textwidth]{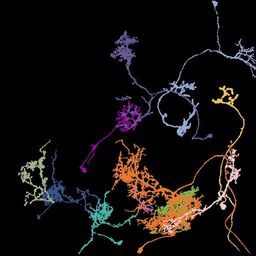}
            \hfil%
        \end{subfigure}\\
        \begin{subfigure}[b]{\textwidth}
            \hfil%
			\includegraphics[width=0.16\textwidth]{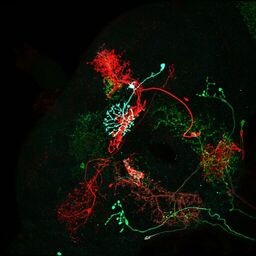}
            \hfil%
			\includegraphics[width=0.16\textwidth]{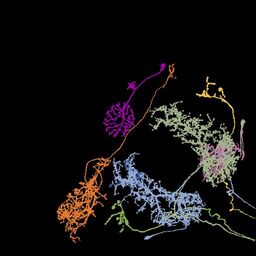}
            \hfil%
            \includegraphics[width=0.16\textwidth]{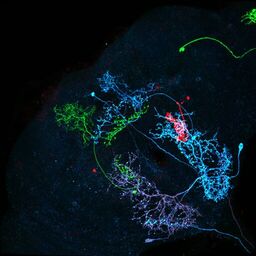}
            \hfil%
			\includegraphics[width=0.16\textwidth]{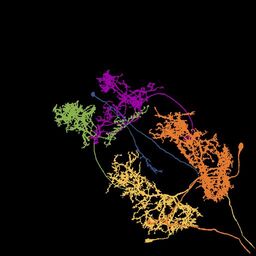}
            \hfil%
            \includegraphics[width=0.16\textwidth]{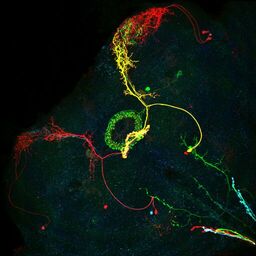}
            \hfil%
			\includegraphics[width=0.16\textwidth]{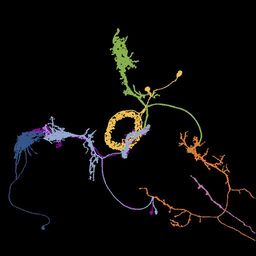}
            \hfil%
        \end{subfigure}\\
		\begin{subfigure}[b]{0.16\textwidth}
			\caption*{}
		\end{subfigure}&
        \begin{subfigure}[b]{0.16\textwidth}
			\caption*{}
		\end{subfigure}&
        \begin{subfigure}[b]{0.16\textwidth}
			\caption*{}
		\end{subfigure}&
		\begin{subfigure}[b]{0.16\textwidth}
			\caption*{}
		\end{subfigure}&
        \begin{subfigure}[b]{0.16\textwidth}
			\caption*{}
		\end{subfigure}&
        \begin{subfigure}[b]{0.16\textwidth}
			\caption*{}
		\end{subfigure}
    \end{tabular}
    \vspace{-5mm}
    \caption{Maximum intensity projections (MIP) of 3d light microscopy samples and ground truth (gt) instance segmentations of all samples in the partly labeled set. MIP and gt are depicted next to each other in alternating order. Images are scaled to same width, some images are center cropped. Figure continued on next page.}
\end{figure*}
\begin{figure*}
    \centering
    \begin{tabular}{m{0.14\textwidth}m{0.14\textwidth}m{0.14\textwidth}m{0.14\textwidth}m{0.14\textwidth}m{0.14\textwidth}}
        \begin{subfigure}[b]{\textwidth}
        \caption*{Test set}
            \hfil%
			\includegraphics[width=0.16\textwidth]{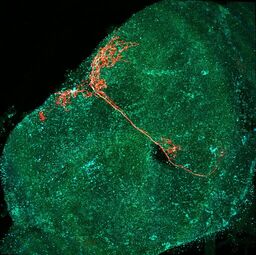}
            \hfil%
			\includegraphics[width=0.16\textwidth]{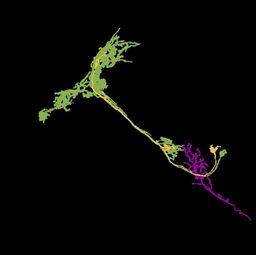}
            \hfil%
            \includegraphics[width=0.16\textwidth]{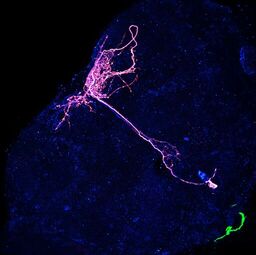}
            \hfil%
			\includegraphics[width=0.16\textwidth]{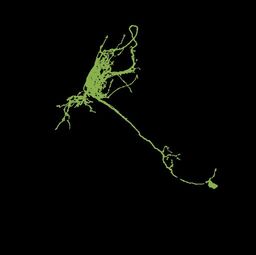}
            \hfil%
            \includegraphics[width=0.16\textwidth]{figures/all_samples/partly/test/raw/R73H08-20181030_62_G5.jpg}
            \hfil%
			\includegraphics[width=0.16\textwidth]{figures/all_samples/partly/test/gt/R73H08-20181030_62_G5.jpg}
            \hfil%
        \end{subfigure}\\
        \begin{subfigure}[b]{\textwidth}
            \hfil%
			\includegraphics[width=0.16\textwidth]{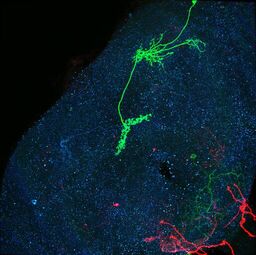}
            \hfil%
			\includegraphics[width=0.16\textwidth]{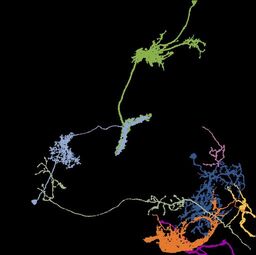}
            \hfil%
            \includegraphics[width=0.16\textwidth]{figures/all_samples/partly/test/raw/VT011145-20171222_63_I2.jpg}
            \hfil%
			\includegraphics[width=0.16\textwidth]{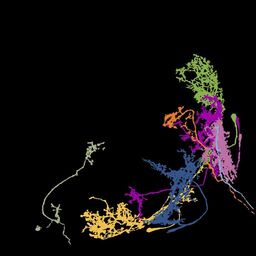}
            \hfil%
            \includegraphics[width=0.16\textwidth]{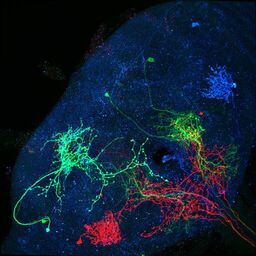}
            \hfil%
			\includegraphics[width=0.16\textwidth]{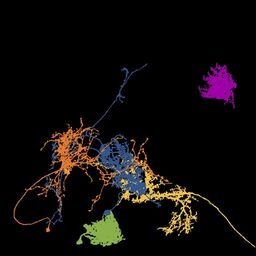}
            \hfil%
        \end{subfigure}\\
        \begin{subfigure}[b]{\textwidth}
            \hfil%
			\includegraphics[width=0.16\textwidth]{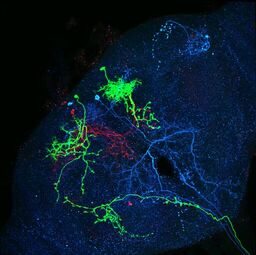}
            \hfil%
			\includegraphics[width=0.16\textwidth]{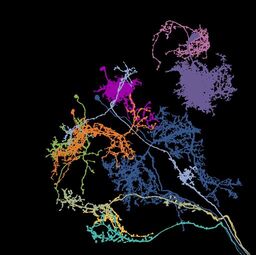}
            \hfil%
            \includegraphics[width=0.16\textwidth]{figures/all_samples/partly/test/raw/VT027175-20171031_62_H6.jpg}
            \hfil%
			\includegraphics[width=0.16\textwidth]{figures/all_samples/partly/test/gt/VT027175-20171031_62_H6.jpg}
            \hfil%
            \includegraphics[width=0.16\textwidth]{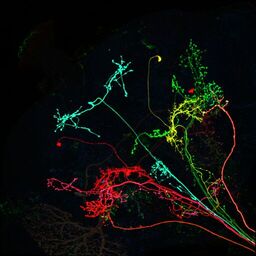}
            \hfil%
			\includegraphics[width=0.16\textwidth]{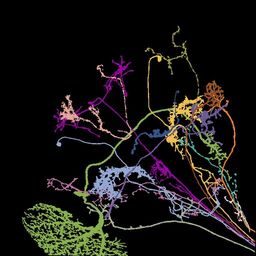}
            \hfil%
        \end{subfigure}\\
        \begin{subfigure}[b]{\textwidth}
            \hfil%
			\includegraphics[width=0.16\textwidth]{figures/all_samples/partly/test/raw/VT028606-20170721_65_A3.jpg}
            \hfil%
			\includegraphics[width=0.16\textwidth]{figures/all_samples/partly/test/gt/VT028606-20170721_65_A3.jpg}
            \hfil%
            \includegraphics[width=0.16\textwidth]{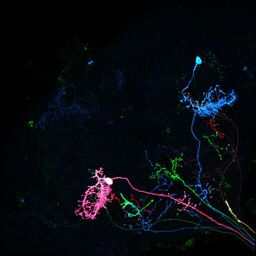}
            \hfil%
			\includegraphics[width=0.16\textwidth]{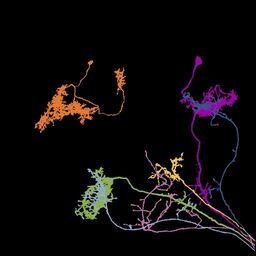}
            \hfil%
            \includegraphics[width=0.16\textwidth]{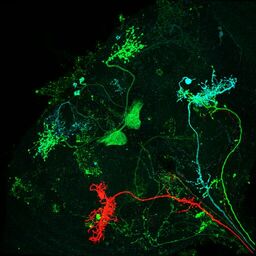}
            \hfil%
			\includegraphics[width=0.16\textwidth]{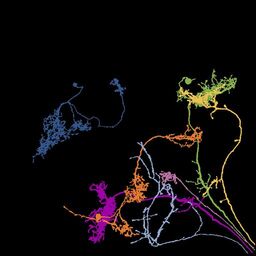}
            \hfil%
        \end{subfigure}\\
        \begin{subfigure}[b]{\textwidth}
            \hfil%
			\includegraphics[width=0.16\textwidth]{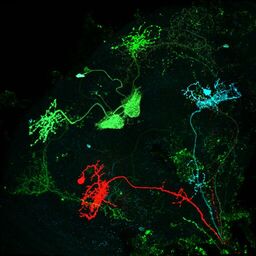}
            \hfil%
			\includegraphics[width=0.16\textwidth]{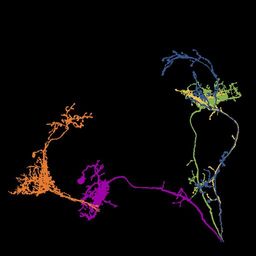}
            \hfil%
            \includegraphics[width=0.16\textwidth]{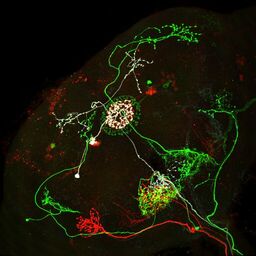}
            \hfil%
			\includegraphics[width=0.16\textwidth]{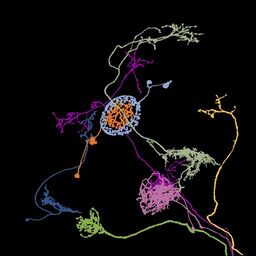}
            \hfil%
            \includegraphics[width=0.16\textwidth]{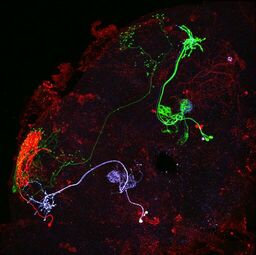}
            \hfil%
			\includegraphics[width=0.16\textwidth]{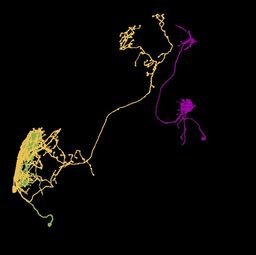}
            \hfil%
        \end{subfigure}\\
      \begin{subfigure}[b]{\textwidth}
        \hfil%
        \includegraphics[width=0.16\textwidth]{figures/all_samples/partly/test/raw/VT058571-20170926_64_G6.jpg}
        \hfil%
        \includegraphics[width=0.16\textwidth]{figures/all_samples/partly/test/gt/VT058571-20170926_64_G6.jpg}
        \hfil%
        \includegraphics[width=0.16\textwidth]{figures/all_samples/placeholder.jpg}
        \hfil%
        \includegraphics[width=0.16\textwidth]{figures/all_samples/placeholder.jpg}
        \hfil%
        \includegraphics[width=0.16\textwidth]{figures/all_samples/placeholder.jpg}
        \hfil%
        \includegraphics[width=0.16\textwidth]{figures/all_samples/placeholder.jpg}
        \hfil%
        \end{subfigure}\\
		\begin{subfigure}[b]{0.16\textwidth}
			\caption*{}
		\end{subfigure}&
        \begin{subfigure}[b]{0.16\textwidth}
			\caption*{}
		\end{subfigure}&
        \begin{subfigure}[b]{0.16\textwidth}
			\caption*{}
		\end{subfigure}&
		\begin{subfigure}[b]{0.16\textwidth}
			\caption*{}
		\end{subfigure}&
        \begin{subfigure}[b]{0.16\textwidth}
			\caption*{}
		\end{subfigure}&
        \begin{subfigure}[b]{0.16\textwidth}
			\caption*{}
		\end{subfigure}
    \end{tabular}
    \vspace{-5mm}
    \caption{Maximum intensity projections (MIP) of 3d light microscopy samples and ground truth (gt) instance segmentations of all samples in the partly labeled set. MIP and gt are depicted next to each other in alternating order. Images are scaled to same width, some images are center cropped. Figure continued from previous page.}
\end{figure*}



\end{document}